\documentclass{article}

 \usepackage[preprint]{neurips_2021}

\usepackage{comment}
\usepackage[utf8]{inputenc} 
\usepackage[T1]{fontenc}    
\usepackage{hyperref}       
\hypersetup{
	colorlinks=true,
	linkcolor=[rgb]{0,0.5,0.5},
	filecolor=cyan,      
	urlcolor=red,
	citecolor=blue,
}
\usepackage{url}            
\usepackage{booktabs}       
\usepackage{amsfonts}       
\usepackage{nicefrac}       
\usepackage{microtype}      
\usepackage{appendix}
\usepackage{microtype}      
\usepackage{xcolor}         
\usepackage{svg}
\usepackage{graphicx}
\usepackage{subfigure}
\usepackage{amsmath}
\usepackage{amsthm}
\usepackage{mathrsfs}
\usepackage{algorithm}
\usepackage{algorithmic}
\usepackage{amssymb}

\DeclareMathOperator*{\argmax}{arg\,max}

\theoremstyle{definition}
\theoremstyle{Theorem}
\theoremstyle{Lemma}
\theoremstyle{Corollary}
\theoremstyle{Proof}
\theoremstyle{Remark}

\title{An Entropy Regularization Free Mechanism for Policy-based Reinforcement Learning}

\author{%
  Changnan Xiao \\
  ByteDance\\
  \texttt{xiaochangnan@bytedance.com} \\
   \And
   Haosen Shi \\
   Nankai University \\
   \texttt{shihaosen98@gmail.com} \\
   \AND
   Jiajun Fan \\
   Nankai University \\
   \texttt{jiajunfanthu@gmail.com} \\
   \And
   Shihong Deng \\
   ByteDance \\
   \texttt{dengshihong@bytedance.com} \\
}
\begin{document}

\maketitle

\begin{abstract}
Policy-based reinforcement learning methods suffer from the policy collapse problem.
We find valued-based reinforcement learning methods with $\epsilon$-greedy mechanism are capable of enjoying three characteristics, Closed-form Diversity, Objective-invariant Exploration and Adaptive Trade-off, which help value-based methods avoid the policy collapse problem.
However, there does not exist a parallel mechanism for policy-based methods that achieves all three characteristics.
In this paper, we propose an entropy regularization free mechanism that is designed for policy-based methods, which achieves Closed-form Diversity, Objective-invariant Exploration and Adaptive Trade-off.
Our experiments show that our mechanism is super sample-efficient for policy-based methods and boosts a policy-based baseline to a new State-Of-The-Art on Arcade Learning Environment.
\end{abstract}

\section{INTRODUCTION}
\label{sec:intro}

Reinforcement Learning (RL) algorithms can be divided into two categories, value-based methods and policy-based methods \citep{pcl}.
Policy-based reinforcement learning methods learn a parameterized policy directly without consulting a value function \citep{sutton}.
These methods suffer from the policy collapse problem, where the entropy of the policy drops to zero but the learned policy is far from achieving the optimal policy 
\citep{a2c,polytope,a3c}. 
It's unsatisfactory to mitigate this problem by increasing the amount of training data due to the fact that, when the learned policy is trapped at a sub-optimal solution, it cannot be improved anymore. 
This weakness prevents policy-based methods from enjoying the benefit of a large training scale.

Value-based reinforcment learning methods generate the behavior policy through a learned state-action value function \citep{sutton}.
The behavior policy is generally controlled by $\epsilon$-greedy.
This mechanism enjoys several characteristics. 
Firstly, $\epsilon$-greedy is closed-form, and it enjoys one dimension of flexibility for adjusting the exploration rate of the behavior policy.
Specifically, the exploration rate can be simply controlled by adjusting the scale of $\epsilon$.
So it has a closed-form family of behavior policies, which is diverse in one dimension of freedom, $\epsilon$.
We call this characteristic that \textit{the behavior policy can be sampled from a family of policies that are defined in a closed-form function of the target policy} as Closed-form Diversity.
Secondly, the objective function of the value-based methods is free from $\epsilon$.
The choice of $\epsilon$ is arbitrary, which would not interfere with 
the convergence of the target policy to the optimal policy.
We call this characteristic that \textit{no matter how the behavior policy explores, the objective of the target policy is always the original objective of the MDP} as Objective-invariant Exploration.
Thirdly, due to the fact that $\epsilon$-greedy is closed-form and would not influence the convergence property of the target policy, it's not difficult to implement some adaptive mechanism for a better trade-off between exploration and exploitation, such as \citep{agent57,adaptiveepsilon,adaptiveepsilon2}.
We call this characteristic that \textit{there exists some mechanisms that the behavior policy can adaptively balance the trade-off between exploration and exploitation} as Adaptive Trade-off.

From the perspective of exploration and exploitation, these three characteristics of $\epsilon$-greedy are critical to balancing exploration and exploitation.
Closed-form Diversity guarantees sufficient exploration, where the trajectories are generated from not only the target policy but a family of behavior policies with different exploration rates.
Objective-invariant Exploration guarantees sufficient exploitation, where the optimal policy and fundamental elements of the Markov Decision Process (MDP) are consistent, regardless from which behavior policy the trajectories are generated.
Adaptive Trade-off guarantees sample efficiency, where the behavior policy with elite trajectories should be paid more attention by the target policy.

As mentioned above, policy-based methods suffer from the policy collapse problem. 
Utilizing entropy regularization in the objective function for encouraging exploring is a crucial component for recent policy-based RL methods \citep{ppo,a3c,impala,sac,sql}.
The entropy regularization coefficient is usually
declining according to some annealing mechanism or fixed at a picked value.
But the exploration rate induced by the entropy regularization coefficient is implicit, which fails to define the exploration rate of the behavior policy as a closed-form function of the target policy.
Meanwhile, since it encourages exploration of the behavior policy by introducing an entropy regularization term into the objective function of the target policy, the objective of the target policy is inconsistent and not Objective-invariant during the training process.
The gap between original MDP and entropy augmented MDP leads that target policy converges to a sub-optimal policy \citep{revisitingsoft}.
Moreover, in practice, it is hard to control the effect of such mechanism adaptive to different environments and targets. 
It generally takes many parallel experiments to find a proper regularization coefficient, which results in a lower sample efficiency behind the reported results.
Although some policy-based methods \citep{softandapp,vmpo} achieve Adaptive Trade-off, they also fail to achieve Closed-Form Diversity and Objective-invariant Exploration.

\begin{table}[h]
\centering
\begin{tabular}{cccccccc}
\toprule
                        & DQN           & R2D2      & Agent57 & A2C  & PPO    & IMPALA & SAC   \\ 
\midrule
Closed-form Diversity   & \checkmark      & \checkmark & \checkmark & $\times$ & $\times$ & $\times$ & $\times$    \\
Objective-invariant Exploration & \checkmark    & \checkmark & $\times$ & $\times$ & $\times$ & $\times$ & $\times$     \\
Adaptive Trade-off      & $\times$        & $\times$   & \checkmark & $\times$  & $\times$ & $\times$ & \checkmark   \\
\bottomrule
\end{tabular}
\caption{Survey of Algorithms. \checkmark represents the algorithm has the characteristic. $\times$ represents not.}
\end{table}

Value-based methods \citep{dqn,r2d2,agent57} is capable to enjoying three characteristics.
DQN and R2D2 lack a meta-controller. 
Agent57 involves intrinsic rewards, so it's not Objective-invariant.

But for policy-based methods, to the best of our knowledge, there is no mechanism for policy-based methods that is capable to enjoying Closed-form Diversity, Objective-invariant Exploration and Adaptive Trade-off.


We propose a new mechanism DiCE\footnote{DiCE means that the behavior policy can be sampled from a family of behavior policies as easy as tossing a dice.}, Directly Control Entropy, which is designed specifically for policy-based methods.
DiCE for policy-based methods is analogy to $\epsilon$-greedy for valued-based methods and it also enjoys Closed-form Diversity, Objective-invariant Exploration and Adaptive Trade-off.
Same as $\epsilon$-greedy, DiCE holds one dimension of freedom to adjust the exploration rate, which is the temperature $\tau$.
The behavior policies are in a Closed-form family.
The target policy is always Objective-invariant, no matter what the behavior policy's temperature is.
Additionally, we exploit these properties of DiCE and provide a simple implementation for the Adaptive Trade-off between exploration and exploitation.
Our experiments show that DiCE boosts policy-based baseline by a large margin and achieves the new State-Of-The-Art (SOTA) under 200M training scale.

Our main contributions are as follows:
\begin{itemize}
\setlength{\topsep}{0pt}
\setlength{\itemsep}{0pt}
\setlength{\partopsep}{0pt}
\setlength{\parsep}{0pt}
\setlength{\parskip}{0pt}
    \item We propose three critical characteristics of value-based methods, Closed-Form Diversity, Objective-invariant Exploration and Adaptive Trade-off.
    \item We propose a new mechanism for policy-based methods that enjoys Closed-Form Diversity, Objective-invariant Exploration and Adaptive Trade-off.
    \item We provide one kind of bandit-based implementation for Adaptive Trade-off.
    \item Our experiments show that the proposed mechanism boosts policy-based methods to a new state-of-the-art.
\end{itemize}

\begin{figure}[ht]
    \subfigure[Mean]{
    \includegraphics[height=3.9cm]{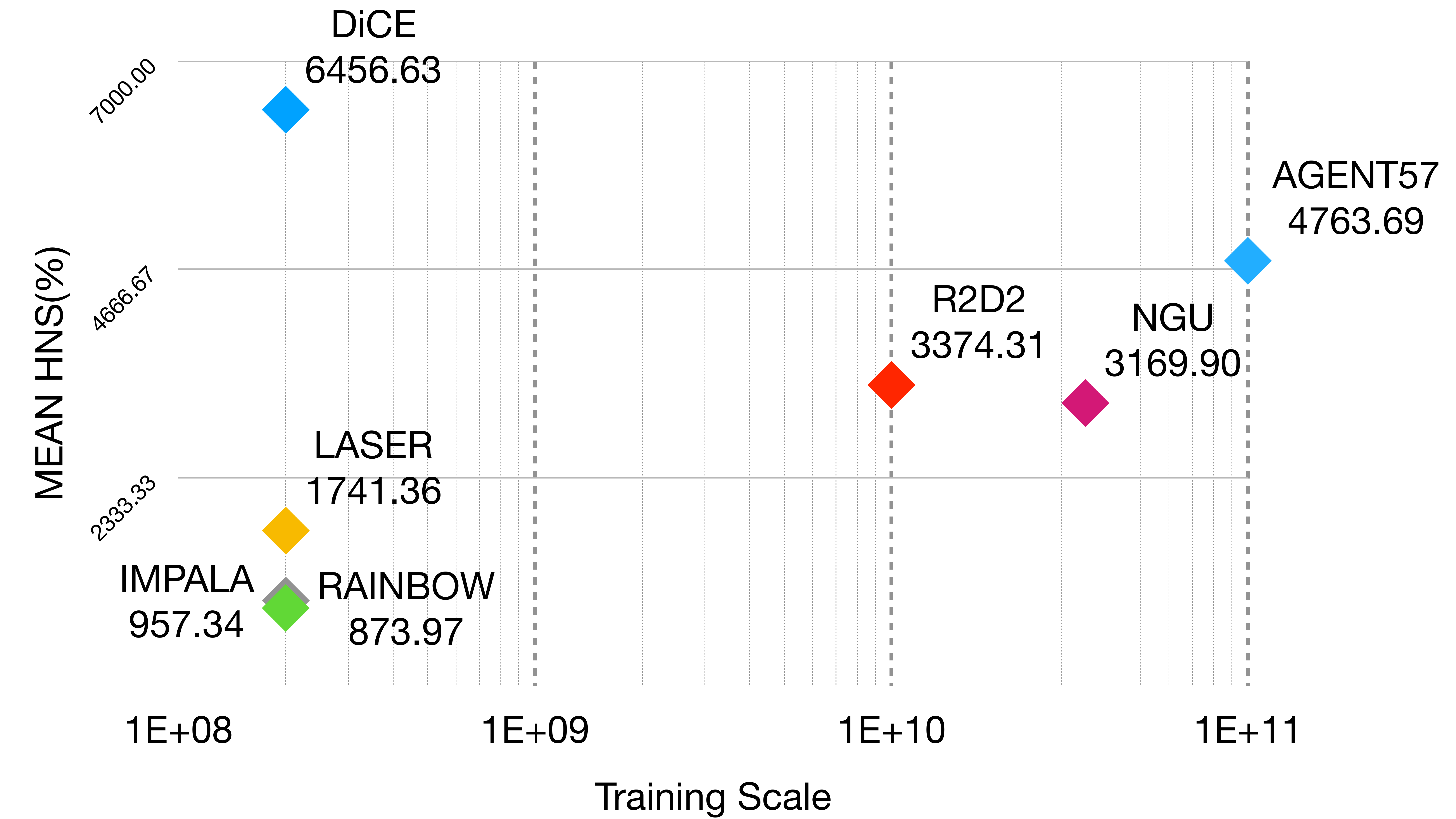}
    }
    \subfigure[Median]{
    \includegraphics[height=3.9cm]{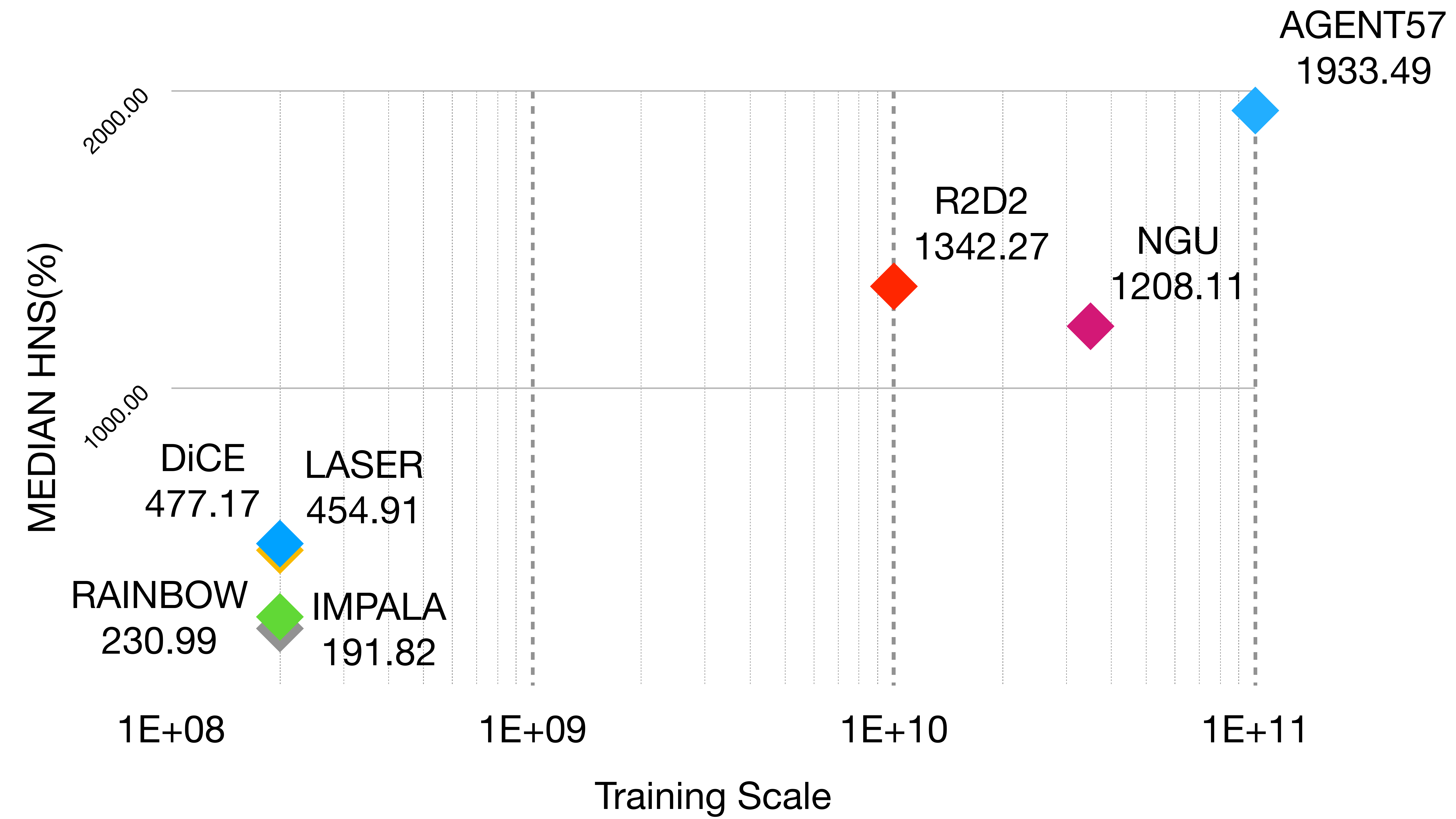}
    }
    \vspace{-0.1cm}
    \caption{Human Normalized Scores (\%) of 57 games.}
\end{figure}

\section{BACKGROUND}
\label{sec:bg}

Consider an MDP, defined by a 5-tuple $(\mathcal{S}, \mathcal{A}, p, r, \gamma)$, where $\mathcal{S}$ is thee state space, $\mathcal{A}$ is the action space, $p$ is the state transition probability function which associates distributions over state to state-action pair, $r: \mathcal{S}\times\mathcal{A} \rightarrow \mathbb{R}$ is the reward function, and $\gamma \in (0,1)$ is the discounted factor. 
Let $\pi$ a distribution over action to each state be the policy. 

The objective of reinforcement learning is to find
\begin{equation}
\argmax_\pi \ \mathcal{J}_\pi = \mathbb{E}_{(s_0,a_0,s_1,\ldots,s_t,a_t,\ldots) \sim \pi} \left[ \sum_{k=0}^\infty \gamma^k r_{k} \right].
\label{eq:objective}
\end{equation}
, where $\mathbb{E}_{(s_0,a_0,s_1,\ldots,s_t,a_t,\ldots) \sim \pi}$ is a brief form of $\mathbb{E}_{a_1 \sim \pi(a_1 | s_0),s_{1}\sim p(s_{1}|a_1,s_0),\ldots}$.

Value-based methods maximize $\mathcal{J}_\pi$ by generalized policy iteration (GPI) \citep{sutton}. 
The policy evaluation is conducted by minimizing $\mathbb{E}_{\pi} [(\mathcal{T} Q^{\pi} - Q^\pi) ^ 2]$, where $\mathcal{T} Q^{\pi}$ is 
$$
\begin{aligned}
    \mathcal{T} Q^{\pi} (s_t, a_t) &= \mathbb{E}_p [r_t + \gamma V^{\pi}(s_t)],\\
    V^{\pi}(s_t)&=\mathbb{E}_\pi[Q^\pi(s_t,a_t)]
\end{aligned}
$$
where $\mathbb{E}_p$ is a shorthand of 
$\mathbb{E}_{s_{t+1}\sim p(s_{t+1}|s_t, a_t)}$ 
and $\mathbb{E}_{\pi}$ is a shorthand of 
$\mathbb{E}_{a_t \sim \pi(a_t | s_t)}$.
The policy improvement is usually achieved by $\epsilon$-greedy. 
A refined structure design of $Q^\pi$ is provided by dueling-Q \citep{dueling_q}. It estimates $Q^\pi$ by the summation of the advantage function and the state value function, $Q^\pi = A^\pi + V^\pi$.

When the large scale training is involved, the off-policy problem is inevitable.
Denote $\mu$ as the behavior policy, $\pi$ as the target policy.
$c_t \overset{def}{=} \min\{\frac{\pi_t}{\mu_t}, \Bar{c}\}$ is the clipped importance sampling. 
For brevity, $c_{[t: t+k]} \overset{def}{=} \prod_{i=0}^{k} c_{t+i}$. 
ReTrace \citep{retrace} estimates $Q(s_t, a_t)$ by clipped per-step importance sampling
\begin{equation}
\label{eq:retrace}
    Q^{\Tilde{\pi}} (s_t, a_t) 
= \mathbb{E}_{\mu} [ Q(s_t, a_t) + \sum_{k \geq 0} \gamma^k 
c_{[t+1:t+k]} \delta^{Q}_{t+k} Q ],
\end{equation}
where $\delta^{Q}_t Q \overset{def}{=} r_t + \gamma Q(s_{t+1}, a_{t+1}) - Q(s_t, a_t)$. 
The ReTrace operator is a contraction mapping
and 
has a fixed point
$Q^{\Tilde{\pi}_{ReTrace}}$ corresponding to some $\Tilde{\pi}_{ReTrace}$.
$Q^\pi$ is learned by minimizing $\mathbb{E}_{\mu} [(Q^{\Tilde{\pi}} - Q^\pi) ^ 2]$, which gives the gradient ascent direction 
\begin{equation}
\mathbb{E}_{\mu} [(Q^{\Tilde{\pi}} - Q^\pi) \nabla Q^\pi].
\label{eq:q_grad}
\end{equation}

\begin{table}[H]
    \centering
    \begin{tabular}{cccccc}
    \toprule
     & & \multicolumn{2}{|c|}{HNS(\%)} & \multicolumn{2}{c}{SABER(\%)} \\ 
    \cmidrule{2-4} \cmidrule{5-6}
     & \multicolumn{1}{c|}{Num. Frames} & Mean & \multicolumn{1}{c|}{Median} & Mean & Median \\ 
    \midrule
    DiCE & 200M & 6456.63 & 477.17 & 50.11 & 13.90 \\ 
    Rainbow & 200M & 873.97 & 230.99 & 28.39 & 4.92 \\ 
    IMPALA & 200M & 957.34 & 191.82 & 29.45 & 4.31 \\ 
    LASER & 200M & 1741.36 & 454.91 & 36.77 & 8.08 \\
    \midrule
    R2D2 & 10B & 3374.31 & 1342.27 & 60.43 & 33.62 \\
    NGU  & 35B & 3169.90 & 1208.11 & 50.47 & 21.19 \\
    Agent57 & 100B & 4763.69 & 1933.49 & 76.26 & 43.62 \\
    \bottomrule
    \end{tabular}
    \caption{Atari Scores. 
    Rainbow's scores are from \citep{rainbow}.
    IMPALA's scores are from \citep{impala}.
    LASER's scores are from \citep{laser}, no sweep at 200M.
    R2D2's scores are from \citep{r2d2}.
    NGU's scores are from \citep{ngu}.
    Agent57's scores are from \citep{agent57}.}
    \label{tab:atari_results}
\end{table}

Policy-based methods maximize $\mathcal{J}_\pi$ by policy gradient.
It's shown \citep{sutton} that 
$\nabla \mathcal{J}_\pi = \mathbb{E}_\pi [G \nabla \log \pi]$, where $G=\sum_{i=t}^\infty \gamma^ir_i$.
When involved with an estimated baseline, it becomes an actor-critic algorithm.
The policy gradient is given by $\nabla \mathcal{J}_\pi = \mathbb{E}_\pi [(G - V^\pi) \nabla \log \pi]$, where $V^\pi$ is optimized by minimizing $\mathbb{E}_\pi [(G - V^\pi)^2]$.

Since \citep{williams1991function}, many on-policy policy-based methods \citep{a3c,ppo} add an entropy regularization to loss function like $\alpha \nabla \mathcal{H}(\pi)$, where $\mathcal{H}(\pi)=-\sum_{a'} \pi(a'|s)\log \pi(a'|s)$ is the entropy of $\pi$.

As one large scale policy-based method, IMPALA \citep{impala} also introduces entropy. 
Besides, IMPALA introduces V-Trace off-policy actor-critic algorithm to correct the discrepancy between the target policy and the behavior policy. 
Let $\rho_t \overset{def}{=} \min\{\frac{\pi_t}{\mu_t}, \Bar{\rho} \}$, 
$c_t \overset{def}{=} \min\{\frac{\pi_t}{\mu_t}, \Bar{c} \}$ to be the clipped importance sampling.
For brevity, let $c_{[t: t+k]} \overset{def}{=} \prod_{i=0}^{k} c_{t+i}$. 
V-Trace estimates $V(s_t)$ by
\begin{equation}
\label{eq:V-Trace}
    V^{\Tilde{\pi}} (s_t) 
        = \mathbb{E}_{\mu} [ 
        V(s_t) + \sum_{k \geq 0} \gamma^k 
     c_{[t:t+k-1]} \rho_{t+k}  \delta^{V}_{t+k} V ],
\end{equation}
where $\delta^{V}_t V \overset{def}{=} r_t + \gamma V(s_{t+1}) - V(s_t)$. 
$V^\pi$ is learned by minimizing $\mathbb{E}_\mu [(V^{\Tilde{\pi}} - V^\pi)^2]$, giving the gradient ascent direction 
\begin{equation}
    \mathbb{E}_\mu [(V^{\Tilde{\pi}} - V^\pi)\nabla V^\pi].
\label{eq:v_grad}
\end{equation}

If $\Bar{c} \leq \Bar{\rho}$, the V-Trace operator is a contraction mapping, and $V$ converges to $V^{\Tilde{\pi}}$ that corresponds to 
$$
        \Tilde{\pi}(a|s) = \frac
        {\min \left\{\Bar{\rho} \mu (a|s), \pi(a|s)\right\}}
        {\sum_{b \in \mathcal{A}}\min \left\{\Bar{\rho} \mu (b|s), \pi(b|s)\right\}}.
$$
Combining V-Trace into the Actor-Critic paradigm, the direction of the policy gradient is given by
\begin{equation}
    \mathbb{E}_\mu \left[\rho_t (r_t + \gamma V^{\Tilde{\pi}}(s_{t+1}) - V(s_t)) \nabla \log \pi \right].
\label{eq:pg_impala}
\end{equation}


Another related work is maximum entropy RL \citep{sql,sac}, which maximizes state value as well as the entropy of the policy. 
Here we call them MaxEnt.
The objective of MaxEnt is to
$$
maximize\ \mathcal{J}_{soft} = \mathbb{E}_{\pi} \left[\sum_{k=0}^{\infty} \gamma^k (r_k + \tau \textbf{H}[\pi]) \right].
$$
$Q^{\pi}_{soft}$ and $V^{\pi}_{soft}$ are defined by the \textit{soft} Bellman equation
\begin{equation}
\label{eq:soft_bellman}
\begin{aligned}
    &\mathcal{T} Q^\pi_{soft} (s_t, a_t) = \mathbb{E}_p [r_t + \gamma V^\pi_{soft} (s_{t+1})],  \\
    where \ \ 
    &V^\pi_{soft} (s_t) = \mathbb{E}_\pi [Q^\pi_{soft}(s_t, a_t) - \tau \log \pi(a_t| s_t)].
\end{aligned}
\end{equation}
$\pi_{soft}$ is defined as a Boltzmann policy
$$
\pi_{soft}(a_t | s_t) \propto \exp ((Q^\pi_{soft} (s_t, a_t) - V^\pi_{soft} (s_t)) / \tau).
$$
Although \citep{sac} achieves Adaptive Trade-off by adjusting $\tau$ adaptively, it still suffers some drawbacks.
\textbf{i)} Since $\tau$ in $\mathcal{J}_{soft}$ encourages exploration by maximizing entropy reward when optimizing the target policy, it couples the exploration rate of the behavior policy with the target policy, which makes it neither Closed-form nor Diversity.
\textbf{ii)} $\mathcal{J}_{soft}$ is regularized by \textbf{H}[$\pi$] compared to $\mathcal{J}_\pi$, which means the target policy is not Objective-invariant during training.

\section{Method}
\label{sec:method}

In this section, we introduce DiCE and also check three characteristics.
At the beginning of \ref{sec:closed-form}, we use $V_\theta$ to estimates $V^\pi$, $A_\theta$ to estimates $A^\pi$, $Q_\theta = V_\theta + A_\theta$ to estimate $Q^\pi$ and $\pi_\theta$ to approximate $\pi$, where $\theta$ represents all parameters to be optimized.
But the target functions of $V_\theta$, $A_\theta$ and $Q_\theta$ will be changed later, which is detailed in \ref{sec:closed-form}.
In \ref{sec:objective-invariant} and \ref{sec:ent_control}, $V_\theta$, $A_\theta$ and $Q_\theta$ estimate the new target functions.

\subsection{Closed-form Diversity}
\label{sec:closed-form}

Assume $v_\theta \in \mathbb{R}^{|\mathcal{A}|}$ is a parameterized vector. For brevity, we write $v_\theta$ as $v$.
Inspired by MaxEnt, we define a Boltzmann policy as
$$
\pi_\tau (v) \propto \exp (v / \tau).
$$
It's evident that, 
$$
\begin{aligned}
    &\textbf{H}[\pi_\tau] \to 0,&\ &\pi_\tau \to argmax,&\ &as\ \tau \to 0+; \\
    &\textbf{H}[\pi_\tau] \to \log |\mathcal{A}|,&\ &\pi_\tau \to uniform,&\ &as\ \tau \to +\infty. \\
\end{aligned}
$$
Since $\textbf{H}[\pi_\tau]$ is a continuous function of $\tau$, for $\forall\ 0 < e < \log |\mathcal{A}|$, $\exists\  \tau_0$ s.t. $\textbf{H} [\pi_{\tau_0}] = e$.
So we can simply control $\textbf{H}[\pi_\tau]$ by choosing a proper $\tau$.

If we define a family of behavior policies as 
$$
\mathcal{B}_v = \{\pi_\tau(v) |\, \tau \in (0, +\infty)\},
$$
it's obvious that $\mathcal{B}_v$ is a family of closed-form functions of $v$ and $\mathcal{B}_v$ has one dimension of freedom $\tau$.
So $\mathcal{B}_v$ achieves Closed-form Diversity for $v$.

Since the behavior policy is sampled from a family $\mathcal{B}_v$, it's nature to ask what the definitions of $V^\pi$, $Q^\pi$ and $A^\pi$ are, i.e. what functions $V_\theta$, $Q_\theta$ and $A_\theta$ are estimating.
As $(V^{\pi_\tau}, Q^{\pi_\tau})$ satisfies the following Bellman equation,
$$
\begin{aligned}
    \mathcal{T}Q^{\pi_\tau}_t &= \mathbb{E}_{p} [ r_t + \gamma V^{\pi_\tau}_{t+1} ], \\
    where \ V^{\pi_\tau}_t &= \mathbb{E}_{a_t
    \sim \pi_\tau} [Q^{\pi_\tau}_t (\cdot, a_t)], \\
\end{aligned}
$$
we observe that $\tau$ only influences $(V^{\pi_\tau}, Q^{\pi_\tau})$ when taking the expectation $V^{\pi_\tau}_t = \mathbb{E}_{\pi_\tau} [Q^{\pi_\tau}_t]$. 

Assume $\tau$ is sampled from a distribution $\Omega$, the influence is totally eliminated as
$$
\begin{aligned}
    \mathcal{T}Q^{\pi_\tau}_t &= \mathbb{E}_p [ r_t + \gamma V^{\pi_\tau}_{t+1} ], \\
    where \ V^{\pi_\tau}_t &= \mathbb{E}_{\tau \sim \Omega} [\mathbb{E}_{a_t \sim \pi_\tau} [Q^{\pi_\tau}_t(\cdot, a_t)]]. \\
\end{aligned}
$$
Now $Q^{\pi_\tau}_t$ only depends on the transition probability $p$ and the \textit{average} policy $\pi_\Omega \overset{def}{=} \int_{\tau \sim \Omega} \pi_\tau$, which is free from $\tau$.
Hence, we rewrite the above Bellman equation as
$$
\begin{aligned}
    \mathcal{T}Q^{\Omega}_t &= \mathbb{E}_p [ r_t + \gamma V^{\Omega}_{t+1} ], \\ 
    where \ V^{\Omega}_t &= \mathbb{E}_{a_t \sim \pi_\Omega} [Q^{\Omega}_t (\cdot, a_t)], \\
\end{aligned}
$$
where $Q^{\Omega}_t \overset{def}{=} \int_{\tau \sim \Omega} Q^{\pi_\tau}_t$, 
$V^{\Omega}_t \overset{def}{=} \int_{\tau \sim \Omega} V^{\pi_\tau}_t$.
Similarly, we have $A^{\Omega}_t \overset{def}{=} \int_{\tau \sim \Omega} A^{\pi_\tau}_t$.

Now we use $V_\theta$ to estimates $V^\Omega$, $A_\theta$ to estimates $A^\Omega$, $Q_\theta = V_\theta + A_\theta$ to estimate $Q^\Omega$.

\subsection{Objective-invariant Exploration}
\label{sec:objective-invariant}

To check if $\mathcal{B}_v$ is Objective-invariant, by a typical derivative of cross entropy, we have $\nabla \log \pi_\tau = (\textbf{1} - \pi_\tau) \frac{\nabla v}{\tau}$.
Plugging into \eqref{eq:pg_impala}, we have the following policy gradient
$$
    \nabla \mathcal{J}_{\pi_\tau} = \mathbb{E}_{\mu_\tau} \left[\rho_t (r_t + \gamma V^{\Tilde{\pi}_\tau}(s_{t+1}) - V^{\pi_\tau}(s_t)) (\textbf{1} - \pi_\tau) \frac{\nabla v}{\tau} \right],
$$
where ${\mu_\tau}$ is the behavior policy corresponding to ${\pi_\tau}$.


It's obvious that $maximize\ \mathcal{J}_{\pi_\tau}$ is equivalent to $maximize\ \tau \mathcal{J}_{\pi_\tau}$.
Taking an additional $\tau$, we free the scale of policy gradient from $\tau$, 
$$
    \nabla \tau \mathcal{J}_{\pi_\tau} = \mathbb{E}_{\mu_\tau} \left[\rho_t (r_t + \gamma V^{\Tilde{\pi}_\tau}(s_{t+1}) - V^{\pi_\tau}(s_t)) (\textbf{1} - \pi_\tau) \nabla v \right].
$$
Taking expectation w.r.t. $\tau$, we have
\begin{equation}
    \nabla \mathbb{E}_{\tau \sim \Omega} [\tau \mathcal{J}_{\pi_\tau}]
    =\mathbb{E}_{\tau \sim \Omega} \left[ \mathbb{E}_{\mu_\tau} \left[\rho_t (r_t + \gamma V^{\Tilde{\pi}_\tau}(s_{t+1}) - V^{\pi_\tau}(s_t)) (\textbf{1} - \pi_\tau) \nabla v \right] \right].
\label{eq:pg_dice}
\end{equation}
When the importance sampling clip $\Bar{\rho}$ and $\Bar{c}$ is $+\infty$, we know V-Trace is an unbiased estimation \citep{impala}. 
Regardless the function approximation error of $V_\theta$, we have
$$
    \nabla \mathbb{E}_{\tau \sim \Omega} [\tau \mathcal{J}_{\pi_\tau}]
    = \mathbb{E}_{\tau \sim \Omega} \left[ \mathbb{E}_{\pi_\tau} \left[A^{\pi_\tau} (\textbf{1} - \pi_\tau) \nabla v \right] \right].
$$
Therefore, if $\Omega$ is a fixed distribution of $\tau$, $\mathcal{B}_v$ is Objective-invariant as $\mathbb{E}_{\tau \sim \Omega} [\tau \mathcal{J}_{\pi_\tau}]$ is free from $\tau$.
If $\Omega$ can be optimized, let $\tau_0 = \argmax_{\tau} \{\tau \mathcal{J}_{\pi_\tau}\}$, then $\mathbb{E}_{\tau \sim \Omega} [\tau \mathcal{J}_{\pi_\tau}] = \tau_0 \mathcal{J}_{\pi_{\tau_0}}$ is equivalent to the objective defined in \eqref{eq:objective}.
So it's also $\tau$-free.

In practice, we choose $v = A_\theta$, this is because \citep{eq_pg_q} proves the equivalence between the evaluation of the state-action value function and the policy gradient with the evaluation of the state value function.
But it's an open question if there exists a better choice of $v$.

We use the Learner-Actor architecture.
The Actor part of DiCE is shown in \textbf{Algorithm} \ref{alg:dice}, and the full version can be found in appendix \ref{app:full_alg}.

\begin{figure}[ht]
  \centering
  \begin{minipage}{.7\linewidth}
    \begin{algorithm}[H]
      \caption{DiCE Algorithm (Actor).}  
          \begin{algorithmic}
            \STATE Initialize Parameter Server (PS) and Data Collector (DC).
            \STATE // ACTOR
            \STATE Initialize $d_{pull}$, $M$, $\Omega$.
            \STATE Initialize $\theta$, i.e. initialize $A_\theta$, $V_\theta$,$Q_\theta = V_\theta + A_\theta$ and $\pi_\theta$.
            \STATE Initialize $\{\mathcal{B}_m\}_{m=1,...,M}$ and sample $\tau \sim \Omega$.
            \STATE Initialize $count = 0$, $G = 0$.
            \STATE Initialize environment and $s = s_0$.
            \WHILE{$True$}
                \STATE Calculate $\pi_\tau(\cdot | s) \propto \exp (A_\theta / \tau)$.
                \STATE Sample $a \sim \pi_\tau(\cdot | s)$.
                \STATE $s, r, done \sim p(\cdot | s, a)$.
                \STATE $G \leftarrow G + r$.
                \IF{$done$}
                    \STATE Update $\Omega$ by $(\tau, G)$.
                    \STATE Send data to DC.
                    \STATE Sample $\tau \sim \Omega$.
                    \STATE $G \leftarrow 0$.
                    \STATE Reset the environment and $s = s_0$.
                \ENDIF
                \IF{$count \mod d_{pull}$ = 0}
                    \STATE Pull $\theta$ from PS and update $\theta$.
                \ENDIF
                \STATE $count \leftarrow count + 1$.
            \ENDWHILE
          \end{algorithmic}
        \label{alg:dice}
    \end{algorithm}
  \end{minipage}
\end{figure}

\subsection{Adaptive Trade-off}
\label{sec:ent_control}

In Algorithm \ref{alg:dice}, both \textit{"Update $\Omega$ by $(\tau, G)$"} and \textit{"Sample $\tau \sim \Omega$"} are achieved by Adaptive Trade-off.




In practice, we use a Bandits Vote Algorithm (BVA) to adaptively update $\Omega$ to maximize the expected cumulative return. 
Different from other meta-controllers, BVA uses an ensemble of bandits.
During the training process, each bandit is updated individually.
During the evaluation process, each bandit provides one evaluation for each $\tau$ and the final evaluation of each $\tau$ is the average evaluation of all bandits.
During the sample process, each bandit provides several candidates and one candidate is sampled uniformly from all candidates.
\textbf{The details of BVA is shown in Appendix \ref{app:bva}.}

\section{Experiments}
\label{sec:experiments}

We firstly introduce our basic setup.
Then we report our results on ALE, namely, 57 atari games. 
In our ablation study, we report the results of the policy-based baseline of DiCE, which shows that DiCE boosts our policy-based baseline by a large margin.
We also make additional ablation study to check the effect of the proposed characteristics.

\subsection{Basic Setup}
\label{sec:basic_setup}

All the experiment is accomplished using 92 CPU cores and a single Tesla-V100-SXM2-32GB GPU.
To deal with  partially observable MDP (POMDP), we use a recurrent encoder by LSTM \citep{lstm} with 256 units.
We use \textit{burn-in} \citep{r2d2} to deal with representational drift.
We store the recurrent state during inference and make it the start point of the \textit{burn-in} phase.
We train each sample twice.
We do not use any intrinsic reward in any experiment.
To be general, we will not end the episode if life is lost.
 

We use BVA to sample $\tau$s, evaluate $\tau$s and find the best region of $\tau$s.
We use $\tau$s sampled from the best region to evaluate the expected return.
\textbf{Hyperparameters are listed in Appendix \ref{app:hyperparameters}.}

\subsection{Analysis and Summary of Results}
\label{sec:atari_results}

Table \ref{tab:atari_results} summarizes the mean and the median Human Normalized Scores (HNS) of 57 games, as well as Standardized Atari BEnchmark for RL (SABER)\citep{saber}.\footnote{$HNS = \frac{G - G_{random}}{G_{Human} - G_{random}}$. $SABER = \min\{200\%, \frac{G - G_{random}}{G_{HumanWorldRecord} - G_{random}}\}$.} 
\textbf{Scores of 57 games and learning curves are listed in Appendix \ref{app:atari_results}.}
The videos will be released in the future.

In general, DiCE achieves SOTA on mean HNS compared to other algorithms, including 10B+ algorithms.
DiCE also achieves the highest median HNS among 200M algorithms.

The results can roughly be classified into three kinds:
\textbf{(i)} results of some games achieve historical highest score, such as Atlantis, Gopher, Jamesbond.
For fairness, we also report mean and median SABER, which is normalized by Human World Records and capped by 200\%.
The mean SABER is very competitive compared to other 10B+ algorithms.
The mean SABER and median SABER is much higher than other 200M algorithms, which shows the overall performance is much better than other 200M algorithms.
\textbf{(ii)} results of some games are increasing and the learning processes have not converged, such as Alien, BeamRider, ChopperCommand;
\textbf{(iii)} results of some games encounter the bottleneck due to the hard exploration problem, such as IceHockey, PrivateEye, Surround.

One crucial thing for solving \textbf{(iii)} is how to acquire better samples.
Even DiCE achieves controlling the entropy of the behavior policy in a Closed-form, DiCE only maintains one dimension of freedom.
DiCE tries to collect diverse samples by enlarging the behavior policy to a family of policies $\{\pi_{\tau}\}_{\tau \in [K, +\infty]}$.
The family can achieve a large range of entropy.
However, it's obvious that $\pi_{\tau}$ is an order-preserving mapping of $A$,
which means that the total order relations of actions' probability distributions are identical for $\forall\  \pi_{\tau} \in \{\pi_{\tau}\}_{\tau \in [K, +\infty]}$.
Value-based methods such as R2D2 collect samples by a family of policies $\{\pi_\epsilon\}_{\epsilon \in [0, 1]}$, where $\pi_\epsilon$ represents $\epsilon$-greedy.
Except for the max $Q$-value, $\pi_\epsilon$ does not preserve the order.
$\{\pi_\epsilon\}_{\epsilon \in [0, 1]}$ is more likely to explore the state where $\{\pi_{\tau}\}_{\tau \in [K, +\infty]}$ less prefers.
The data distribution of samples is influenced by the inductive bias induced by the family of behavior policies.
It's critical and open to define a proper family of closed-form functions for more elite samples.
Instead, NGU and Agent57 achieve active exploration of the environment by the intrinsic reward.
Such exploration enlarges MDP, so UVFA \citep{uvfa} is needed.

DiCE focuses on Objective-invariant Exploration.
There are some hints to improve DiCE.
The first is to enlarge the function space of the behavior policy by a linear combination. 
Since $\pi_\epsilon$ is identical to $\epsilon \cdot \pi_{\tau \rightarrow +\infty} + (1 - \epsilon) \cdot \pi_{\tau \rightarrow 0+}$,
it's reasonable to consider a combination representation for a larger family of behavior policies, such as $\{\sum_i \alpha_i \cdot \pi_{\tau_i} | \sum_i \alpha_i = 1\}$.
The second is to adopt the idea of LASER by shared experience replay,
which gives diverse samples from different families of policies $\{\{\pi_{n, \tau}\}_{\tau \in [K, +\infty]}\}_{n=1,2,...}$.
The third is to make a better policy evaluation.
Now the value function $V_\theta$ is estimating the \textit{average} policy $\pi_\Omega$, which is $V^\Omega$.
A better way is to use a UVFA or a closed-form expression, if exists, to estimate $\{V^{\pi_{\tau}}\}_{\tau \in [K, +\infty]}$.

\subsection{Ablation Study}
\label{sec:ablation}

We firstly provide the scores of the baseline on 57 atari games.
All hyperparameters of the baseline are the same as DiCE.
Except for that we remove \textit{"Update $\Omega$ by $(\tau, G)$"} and change \textit{"Sample $\tau \sim \Omega$"} to \textit{"$\tau = 1$"} in Algorithm \ref{alg:dice}, the baseline is trained and evaluated identically to DiCE.
From another perspective, we apply DiCE directly on the baseline without any change.

\begin{table}[H]
    \centering
    \begin{tabular}{cccccc}
    \toprule
     & & \multicolumn{2}{|c|}{HNS(\%)} & \multicolumn{2}{c}{SABER(\%)} \\ 
    \cmidrule{2-4} \cmidrule{5-6}
     & \multicolumn{1}{c|}{Num. Frames} & Mean & \multicolumn{1}{c|}{Median} & Mean & Median \\ 
    \midrule
    DiCE & 200M & 6456.63 & 477.17 & 50.11 & 13.90 \\ 
    baseline & 200M & 1929.95 & 195.54 & 35.91 & 8.73 \\ 
    \bottomrule
    \end{tabular}
    \caption{Comparison to baseline.}
    \label{tab:baseline}
\end{table}

The summary is shown in Table \ref{tab:baseline}.
Full comparison on 57 games are listed in Appendix \ref{app:baseline}.
The progress is impressive.
DiCE promotes mean HNS by $235\%$, median HNS by $145\%$, mean SABER by $40\%$ and median SABER by $59\%$.
Recalling Table \ref{tab:atari_results}, although our baseline meets SOTA performance, there is no evident performance difference between our baseline and other 200M algorithms.
But DiCE achieves the best performance on all criteria among 200M algorithm and is approaching 10B+ algorithms on some criteria.
This phenomenon proves that DiCE is very effective and boosts our policy-based baseline to a new State-Of-The-Art.

\begin{table}[H]
    \centering
    \begin{tabular}{cc}
    \toprule
    \textbf{Name} & \textbf{Characteristics} \\
    \midrule
    DiCE & Closed-form Diversity, Objective-invariant Exploration, Adaptive Trade-off \\
    \midrule
    \textit{baseline} & N/A \\
    \midrule
    \textit{random} & Closed-form Diversity, Objective-invariant Exploration \\
    \midrule
    \textit{entropy} &  Closed-form Diversity, Adaptive Trade-off \\
    \bottomrule
    \end{tabular}
    \caption{Ablation Settings.}
    \label{tab:ablation_parameters}
\end{table}

We make additional ablation study to find how much each characteristic contributes to DiCE.
To check Objective-invariant, we add an entropy regularization with coefficient = 1 to the loss function, named as \textit{entropy}.
To check Adaptive Trade-off, we remove \textit{"Update $\Omega$ by $(\tau, G)$"} in Algorithm \ref{alg:dice}, which means the distribution of $\tau$ is always the fixed initial distribution, named as \textit{random}.
It's impossible to check Closed-form Diversity solely.
If Closed-form Diversity does not hold, there is no need to consider Objective-invariant Exploration and Adaptive Trade-off.

We choose Breakout and ChopperCommand to do ablation study, because the performances of DiCE and the baseline on these two games are comparable.

\begin{figure*}[h]
\centering
\subfigure[breakout]{
\includesvg[width=0.33\columnwidth,inkscapelatex=false]{nips2/ablation/Breakout_nips2_ab.svg}
}
\subfigure[chopper\_command]{
\includesvg[width=0.33\columnwidth,inkscapelatex=false]{nips2/ablation/Chooper_Command_nips2_ab.svg}
}
\subfigure{
    \begin{minipage}[t]{4cm}
    \centering
    \raisebox{0.4\height}{\includegraphics[width=0.5\columnwidth]{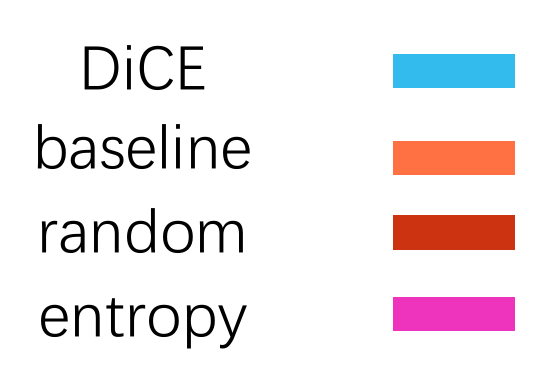}}
    \end{minipage}
}
\caption{Evaluation Return Curves.}
\label{fig:ablation}
\end{figure*}

The evaluation curve are shown in Figure \ref{fig:ablation}.
It's obvious that \textit{random} performs the worst, which evidently proves that Adaptive Trade-off is critical in DiCE.
On the early stage of Breakout, DiCE and \textit{entropy} show a higher sample efficiency that \textit{baseline} and \textit{random}.
Notice that DiCE and \textit{entropy} have Adaptive Trade-off, while \textit{baseline} and \textit{random} do not.
We see Adaptive Trade-Off helps boosting the sample efficiency.
There is no significant difference between DiCE and \textit{entropy}, as DiCE performs better on ChopperCommand but worse on Breakout.
So we cannot make a conclusion whether Objective-invariant is better or not.
As for Closed-form Diversity, the full comparison between DiCE and \textit{baseline} has already shown that DiCE is effective.
Since Closed-form Diversity is the necessary condition for DiCE, there is no doubt that Closed-form Diversity is important.
In general, we conclude that Closed-form Diversity and Adaptive Trade-off are critical.
As for Objective-invariant Exploration, it's uncertain whether Objective-invariant is better, but Exploration is one of the most important thing in RL.

Although DiCE enlarges the behavior policy to a family of behavior policies, it's still a question to find what a good exploration is.
DisCor \citep{discor} has claimed that \textit{the choice of the sampling distribution is of crucial importance for the stability and efficiency of ADP algorithms.}
With a given replay buffer, DisCor re-weights each sample to mitigate this phenomenon.
But if working on sample acquiring rather than sample re-weighting,
it's an interesting question to find a \textit{proper} family of behavior policies such that the target policy learned with collected samples can achieve higher sample efficiency.
Meanwhile, the superiority of this \textit{proper} family of the behavior policies should be guaranteed.
We leave this question for future study.

\section{Conclusion} 
\label{sec:conclusion}

This paper proposes three characteristics, Closed-form Diversity, Objective-invariant Exploration and Adaptive Trade-off.
We propose a mechanism for policy-based methods that enjoys three characteristics.
The mechanism is sample efficient and boosts policy-based baseline to State-Of-The-Art with 200M training scale.
The overall performance is higher than other 200M algorithms and is competitive compared to 10B+ algorithms.
We analyse the ablation cases and discuss the potential improvement in future work.



\bibliographystyle{rusnat}
\bibliography{bibs}

\begin{thebibliography}{28}
\providecommand{\natexlab}[1]{#1}
\providecommand{\EM}{\em}
\providecommand{\RNtxt}{\relax}
\RNtxt{}

\bibitem[Badia et~al.(2020{\natexlab{a}})A.~P. Badia, B.~Piot, S.~Kapturowski,
  P.~Sprechmann, A.~Vitvitskyi, D.~Guo, C.~Blundell]{agent57}
{\EM Badia Adri{\`a}~Puigdom{\`e}nech, Piot Bilal, Kapturowski Steven,
  Sprechmann Pablo, Vitvitskyi Alex, Guo Daniel, Blundell Charles}.
\newblock Agent57: Outperforming the atari human benchmark
  \allowbreak\newblock// arXiv preprint arXiv:2003.13350. 2020{\natexlab{a}}.

\bibitem[Badia et~al.(2020{\natexlab{b}})A.~P. Badia, P.~Sprechmann,
  A.~Vitvitskyi, D.~Guo, B.~Piot, S.~Kapturowski, O.~Tieleman, M.~Arjovsky,
  A.~Pritzel, A.~Bolt, et~al.]{ngu}
{\EM Badia Adri{\`a}~Puigdom{\`e}nech, Sprechmann Pablo, Vitvitskyi Alex, Guo
  Daniel, Piot Bilal, Kapturowski Steven, Tieleman Olivier, Arjovsky
  Mart{\'\i}n, Pritzel Alexander, Bolt Andew, others }.
\newblock Never Give Up: Learning Directed Exploration Strategies
  \allowbreak\newblock// arXiv preprint arXiv:2002.06038. 2020{\natexlab{b}}.

\bibitem[Dadashi et~al.(2019)R.~Dadashi, A.~A. Taiga, N.~Le~Roux,
  D.~Schuurmans, M.~G. Bellemare]{polytope}
{\EM Dadashi Robert, Taiga Adrien~Ali, Le~Roux Nicolas, Schuurmans Dale,
  Bellemare Marc~G}.
\newblock The value function polytope in reinforcement learning
  \allowbreak\newblock// International Conference on Machine Learning. 2019.
  1486--1495.

\bibitem[Espeholt et~al.(2018)L.~Espeholt, H.~Soyer, R.~Munos, K.~Simonyan,
  V.~Mnih, T.~Ward, Y.~Doron, V.~Firoiu, T.~Harley, I.~Dunning, et~al.]{impala}
{\EM Espeholt Lasse, Soyer Hubert, Munos Remi, Simonyan Karen, Mnih Volodymir,
  Ward Tom, Doron Yotam, Firoiu Vlad, Harley Tim, Dunning Iain, others }.
\newblock Impala: Scalable distributed deep-rl with importance weighted
  actor-learner architectures \allowbreak\newblock// arXiv preprint
  arXiv:1802.01561. 2018.

\bibitem[Haarnoja et~al.(2017)T.~Haarnoja, H.~Tang, P.~Abbeel, S.~Levine]{sql}
{\EM Haarnoja Tuomas, Tang Haoran, Abbeel Pieter, Levine Sergey}.
\newblock Reinforcement learning with deep energy-based policies
  \allowbreak\newblock// arXiv preprint arXiv:1702.08165. 2017.

\bibitem[Haarnoja et~al.(2018{\natexlab{a}})T.~Haarnoja, A.~Zhou, P.~Abbeel,
  S.~Levine]{sac}
{\EM Haarnoja Tuomas, Zhou Aurick, Abbeel Pieter, Levine Sergey}.
\newblock Soft actor-critic: Off-policy maximum entropy deep reinforcement
  learning with a stochastic actor \allowbreak\newblock// arXiv preprint
  arXiv:1801.01290. 2018{\natexlab{a}}.

\bibitem[Haarnoja et~al.(2018{\natexlab{b}})T.~Haarnoja, A.~Zhou,
  K.~Hartikainen, G.~Tucker, S.~Ha, J.~Tan, V.~Kumar, H.~Zhu, A.~Gupta,
  P.~Abbeel, et~al.]{softandapp}
{\EM Haarnoja Tuomas, Zhou Aurick, Hartikainen Kristian, Tucker George,
  Ha~Sehoon, Tan Jie, Kumar Vikash, Zhu Henry, Gupta Abhishek, Abbeel Pieter,
  others }.
\newblock Soft actor-critic algorithms and applications \allowbreak\newblock//
  arXiv preprint arXiv:1812.05905. 2018{\natexlab{b}}.

\bibitem[Hessel et~al.(2017)M.~Hessel, J.~Modayil, H.~Van~Hasselt, T.~Schaul,
  G.~Ostrovski, W.~Dabney, D.~Horgan, B.~Piot, M.~Azar, D.~Silver]{rainbow}
{\EM Hessel Matteo, Modayil Joseph, Van~Hasselt Hado, Schaul Tom, Ostrovski
  Georg, Dabney Will, Horgan Dan, Piot Bilal, Azar Mohammad, Silver David}.
\newblock Rainbow: Combining improvements in deep reinforcement learning
  \allowbreak\newblock// arXiv preprint arXiv:1710.02298. 2017.

\bibitem[Kapturowski et~al.(2018)S.~Kapturowski, G.~Ostrovski, J.~Quan,
  R.~Munos, W.~Dabney]{r2d2}
{\EM Kapturowski Steven, Ostrovski Georg, Quan John, Munos Remi, Dabney Will}.
\newblock Recurrent experience replay in distributed reinforcement learning
  \allowbreak\newblock// International conference on learning representations.
  2018.

\bibitem[Kumar et~al.(2020)A.~Kumar, A.~Gupta, S.~Levine]{discor}
{\EM Kumar Aviral, Gupta Abhishek, Levine Sergey}.
\newblock Discor: Corrective feedback in reinforcement learning via
  distribution correction \allowbreak\newblock// arXiv preprint
  arXiv:2003.07305. 2020.

\bibitem[Mnih et~al.(2016{\natexlab{a}})V.~Mnih, A.~P. Badia, M.~Mirza,
  A.~Graves, T.~Lillicrap, T.~Harley, D.~Silver, K.~Kavukcuoglu]{a2c}
{\EM Mnih Volodymyr, Badia Adria~Puigdomenech, Mirza Mehdi, Graves Alex,
  Lillicrap Timothy, Harley Tim, Silver David, Kavukcuoglu Koray}.
\newblock Asynchronous methods for deep reinforcement learning
  \allowbreak\newblock// International conference on machine learning.
  2016{\natexlab{a}}.  1928--1937.

\bibitem[Mnih et~al.(2016{\natexlab{b}})V.~Mnih, A.~P. Badia, M.~Mirza,
  A.~Graves, T.~P. Lillicrap, T.~Harley, D.~Silver, K.~Kavukcuoglu]{a3c}
{\EM Mnih Volodymyr, Badia Adrià~Puigdomènech, Mirza Mehdi, Graves Alex,
  Lillicrap Timothy~P., Harley Tim, Silver David, Kavukcuoglu Koray}.
\newblock Asynchronous Methods for Deep Reinforcement Learning.
  2016{\natexlab{b}}.

\bibitem[Mnih et~al.(2015)V.~Mnih, K.~Kavukcuoglu, D.~Silver, A.~A. Rusu,
  J.~Veness, M.~G. Bellemare, A.~Graves, M.~Riedmiller, A.~K. Fidjeland,
  G.~Ostrovski, et~al.]{dqn}
{\EM Mnih Volodymyr, Kavukcuoglu Koray, Silver David, Rusu Andrei~A, Veness
  Joel, Bellemare Marc~G, Graves Alex, Riedmiller Martin, Fidjeland Andreas~K,
  Ostrovski Georg, others }.
\newblock Human-level control through deep reinforcement learning
  \allowbreak\newblock// nature. 2015. 518, 7540. 529--533.

\bibitem[Munos et~al.(2016)R.~Munos, T.~Stepleton, A.~Harutyunyan,
  M.~Bellemare]{retrace}
{\EM Munos Remi, Stepleton Tom, Harutyunyan Anna, Bellemare Marc}.
\newblock Safe and Efficient Off-Policy Reinforcement Learning
  \allowbreak\newblock// Advances in Neural Information Processing Systems 29.
  2016.  1054--1062.

\bibitem[Nachum et~al.(2017)O.~Nachum, M.~Norouzi, K.~Xu, D.~Schuurmans]{pcl}
{\EM Nachum Ofir, Norouzi Mohammad, Xu~Kelvin, Schuurmans Dale}.
\newblock Bridging the gap between value and policy based reinforcement
  learning \allowbreak\newblock// Advances in Neural Information Processing
  Systems. 2017.  2775--2785.

\bibitem[Santos~Mignon dos, Rocha da(2017)A.~dos Santos~Mignon, R.~L. d.~A.
  da~Rocha]{adaptiveepsilon2}
{\EM Santos~Mignon Alexandre dos, Rocha Ricardo Luis de~Azevedo da}.
\newblock An Adaptive Implementation of $\varepsilon$-Greedy in Reinforcement
  Learning \allowbreak\newblock// Procedia Computer Science. 2017. 109.
  1146--1151.

\bibitem[Schaul et~al.(2015)T.~Schaul, D.~Horgan, K.~Gregor, D.~Silver]{uvfa}
{\EM Schaul Tom, Horgan Daniel, Gregor Karol, Silver David}.
\newblock Universal Value Function Approximators \allowbreak\newblock// ICML.
  2015.  1312--1320.

\bibitem[Schmidhuber(1997)S.~H.~J. Schmidhuber]{lstm}
{\EM Schmidhuber Sepp Hochreiter;~Jürgen}.
\newblock Long short-term memory \allowbreak\newblock// Neural Computation.
  1997.

\bibitem[Schmitt et~al.(2020)S.~Schmitt, M.~Hessel, K.~Simonyan]{laser}
{\EM Schmitt Simon, Hessel Matteo, Simonyan Karen}.
\newblock Off-policy actor-critic with shared experience replay
  \allowbreak\newblock// International Conference on Machine Learning. 2020.
  8545--8554.

\bibitem[Schulman et~al.(2017{\natexlab{a}})J.~Schulman, X.~Chen,
  P.~Abbeel]{eq_pg_q}
{\EM Schulman John, Chen Xi, Abbeel Pieter}.
\newblock Equivalence between policy gradients and soft q-learning
  \allowbreak\newblock// arXiv preprint arXiv:1704.06440. 2017{\natexlab{a}}.

\bibitem[Schulman et~al.(2017{\natexlab{b}})J.~Schulman, F.~Wolski,
  P.~Dhariwal, A.~Radford, O.~Klimov]{ppo}
{\EM Schulman John, Wolski Filip, Dhariwal Prafulla, Radford Alec, Klimov
  Oleg}.
\newblock Proximal policy optimization algorithms \allowbreak\newblock// arXiv
  preprint arXiv:1707.06347. 2017{\natexlab{b}}.

\bibitem[Song et~al.(2020)H.~F. Song, A.~Abdolmaleki, J.~T. Springenberg,
  A.~Clark, H.~Soyer, J.~W. Rae, S.~Noury, A.~Ahuja, S.~Liu, D.~Tirumala,
  N.~Heess, D.~Belov, M.~Riedmiller, M.~M. Botvinick]{vmpo}
{\EM Song H.~Francis, Abdolmaleki Abbas, Springenberg Jost~Tobias, Clark Aidan,
  Soyer Hubert, Rae Jack~W., Noury Seb, Ahuja Arun, Liu Siqi, Tirumala Dhruva,
  Heess Nicolas, Belov Dan, Riedmiller Martin, Botvinick Matthew~M.}
\newblock V-MPO: On-Policy Maximum a Posteriori Policy Optimization for
  Discrete and Continuous Control \allowbreak\newblock// International
  Conference on Learning Representations. 2020.

\bibitem[Song et~al.(2019)Z.~Song, R.~Parr, L.~Carin]{revisitingsoft}
{\EM Song Zhao, Parr Ron, Carin Lawrence}.
\newblock Revisiting the softmax bellman operator: New benefits and new
  perspective \allowbreak\newblock// International Conference on Machine
  Learning. 2019.  5916--5925.

\bibitem[Sutton, Barto(2018)R.~S. Sutton, A.~G. Barto]{sutton}
{\EM Sutton Richard~S, Barto Andrew~G}.
\newblock Reinforcement learning: An introduction. 2018.

\bibitem[Tokic(2010)M.~Tokic]{adaptiveepsilon}
{\EM Tokic Michel}.
\newblock Adaptive $\varepsilon$-greedy exploration in reinforcement learning
  based on value differences \allowbreak\newblock// Annual Conference on
  Artificial Intelligence. 2010.  203--210.

\bibitem[Toromanoff et~al.(2019)M.~Toromanoff, E.~Wirbel, F.~Moutarde]{saber}
{\EM Toromanoff Marin, Wirbel Emilie, Moutarde Fabien}.
\newblock Is deep reinforcement learning really superhuman on atari? leveling
  the playing field \allowbreak\newblock// arXiv preprint arXiv:1908.04683.
  2019.

\bibitem[Wang et~al.(2016)Z.~Wang, T.~Schaul, M.~Hessel, H.~Hasselt,
  M.~Lanctot, N.~Freitas]{dueling_q}
{\EM Wang Ziyu, Schaul Tom, Hessel Matteo, Hasselt Hado, Lanctot Marc, Freitas
  Nando}.
\newblock Dueling network architectures for deep reinforcement learning
  \allowbreak\newblock// International conference on machine learning. 2016.
  1995--2003.

\bibitem[Williams, Peng(1991)R.~J. Williams, J.~Peng]{williams1991function}
{\EM Williams Ronald~J, Peng Jing}.
\newblock Function optimization using connectionist reinforcement learning
  algorithms \allowbreak\newblock// Connection Science. 1991. 3, 3. 241--268.

\end{thebibliography}

\clearpage

\begin{appendices}

\section{Bandit Vote Algorithm}
\label{app:bva}

The algorithm is shown in \textbf{Algorithm} \ref{alg:bva}.

\begin{figure}[ht]
  \centering
  \begin{minipage}{.7\linewidth}
    \begin{algorithm}[H]
      \caption{Bandits Vote Algorithm.}  
          \begin{algorithmic}
            \FOR{$m=1,...,M$}
                \STATE Sample $mode \sim \{argmax, random\}$, sample $lr$, sample $width$.
                \STATE Initialize $B_m = Bandit(mode, l, r, acc, width, lr, \textbf{w}, \textbf{N}, d)$.
            \ENDFOR
            \WHILE{$True$}  
                \FOR{$m=1,...,M$}
                    \STATE Evaluate $\mathcal{B}_m$ by \eqref{eq:bandit_eval}.
                    \STATE Sample candidates $c_{m, 1}, ..., c_{m, d}$ by $mode$ and \eqref{eq:bandit_score} from $\mathcal{B}_m$.
                \ENDFOR
                \STATE Sample $x$ from $\{c_{m, i}\}_{m=1,...,M; i=1,...,d}$.
                \STATE Execute $x$ and receive $g$.
                \FOR{$m=1,...,M$}
                    \STATE Update $\mathcal{B}_m$ with $(x, g)$ by \eqref{eq:bandit_update}.
                \ENDFOR
            \ENDWHILE
          \end{algorithmic}  
        \label{alg:bva}
    \end{algorithm}
  \end{minipage}
\end{figure}

Let's firstly define a bandit as $\mathcal{B} = Bandit(mode, l, r, acc, width, lr, \textbf{w}, \textbf{N}, d)$.
\begin{itemize}
    \item $mode$ is the mode of sampling, with two choices, $argmax$ and $random$.
    \item $l$ is the left boundary of $\mathcal{B}$, and each $x$ is clipped to $x = \max \{x, l\}$.
    \item $r$ is the right boundary of $\mathcal{B}$, and each $x$ is clipped to $x = \min \{x, r\}$.
    \item $acc$ is the accuracy, where each $x$ is located in the $\lfloor (\min\{\max\{x, l\}, r\} - l) / acc \rfloor$th tile.
    \item $width$ is the tile coding width, where the value of the $i$th tile is estimated by the average of $\{\textbf{w}_{i-width},...,\textbf{w}_{i+width}\}$.
    \item $lr$ is the learning rate.
    \item $\textbf{w}$ is a vector in $\mathbb{R}^{\lfloor (r-l) / acc \rfloor}$, which represents the weight of each tile.
    \item $\textbf{N}$ is a vector in $\mathbb{R}^{\lfloor (r-l) / acc \rfloor}$, which counts the number of sampling of each tile.
    \item $d$ is an integer, which represents how many candidates is provided by $\mathcal{B}$ when sampling.
\end{itemize}

During the evaluation process, we evaluate the value of the $i$th tile by
\begin{equation}
\label{eq:bandit_eval}
V_i = \frac{1}{2 * width + 1}\sum_{k=i-width}^{i+width} \textbf{w}_k.\footnotemark
\end{equation}
\footnotetext{For the boundary case, we change the summation indexes and the denominator accordingly.}

During the training process, for each sample $(x, g)$, where $g$ is the target value. Since $x$ locates in the $i$th tile, we update $\mathcal{B}$ by
\begin{equation}
\label{eq:bandit_update}
\left\{
\begin{aligned}
&i = \lfloor (\min\{\max\{x, l\}, r\} - l) / acc \rfloor, \\
&\textbf{w}_j 
\leftarrow \textbf{w}_j + lr * \left(g - V_i\right), j = i - width, ..., i + width, \\
& \textbf{N}_i \leftarrow \textbf{N}_i + 1.
\end{aligned}
\right.
\end{equation}

During the sampling process, we firstly evaluate $\mathcal{B}$ by \eqref{eq:bandit_eval} and get $(V_1, ..., V_{\lfloor (r-l) / acc \rfloor})$.
We calculate the score of $i$th tile by
\begin{equation}
\label{eq:bandit_score}
score_i = \frac{V_i - \mu(\{V_j\}_{j=1,...,\lfloor(r-l)/acc\rfloor})}{\sigma(\{V_j\}_{j=1,...,\lfloor(r-l)/acc\rfloor})} + c \cdot \sqrt{\frac{\log (1 + \sum_j \textbf{N}_j)}{1 + \textbf{N}_i}}.
\end{equation}
For different $mode$s, we sample the candidates by the following mechanism,
\begin{itemize}
    \item if $mode$ = $argmax$, find tiles with top-$d$ $score$s, then sample $d$ candidates from these tiles, one uniformly from a tile;
    \item if $mode$ = $random$, sample $d$ tiles with $score$s as the logits without replacement, then sample $d$ candidates from these tiles, one uniformly from a tile;
\end{itemize}

In practice, we define a set of bandits $\{\mathcal{B}_m\}_{m=1,...,M}$.
At each step, we sample $d$ candidates $\{c_{m, i}\}_{i=1,...,d}$ from each $B_m$, so we have a set of $m \times d$ candidates $\{c_{m, i}\}_{m=1,...,M; i=1,...,d}$.
Then we sample uniformly from these $m \times d$ candidates to get $x$. 
At last, we transform the selected $x$ to $\tau$ by $\tau = \frac{1}{\exp (x) - 1}$.
When we receive $(\tau, g)$, we transform $\tau$ to $x$ by $x = \log (1 + 1 / \tau)$. 
Then we update each $B_m$ by \eqref{eq:bandit_update}.

\clearpage
\section{Full Algorithm}
\label{app:full_alg}
\begin{figure}[ht]
  \centering
  \begin{minipage}{0.7\linewidth}
    \begin{algorithm}[H]
      \caption{DiCE Algorithm full version.}  
          \begin{algorithmic}
            \STATE Initialize Parameter Server (PS) and Data Collector (DC).
            \STATE
            \STATE // LEARNER
            \STATE Initialize $d_{push}$.
            \STATE Initialize $\theta$, i.e. initialize $A_\theta$, $V_\theta$, $Q_\theta = V_\theta + A_\theta$.
            \STATE Initialize $\pi_\tau \propto \exp (A_\theta / \tau)$.
            \STATE Initialize $count = 0$.
            \WHILE{$True$}
                \STATE Load data from DC.
                \STATE Estimate $Q^\pi$ and $V^\pi$ by proper off-policy algorithms.
                \STATE \ \ \ \ (For instance, ReTrace \eqref{eq:retrace} for $Q^\pi$ and V-Trace \eqref{eq:V-Trace} for $V^\pi$.)
                \STATE Update $\theta$ by policy gradient \eqref{eq:pg_dice} and policy evaluation \eqref{eq:q_grad}, \eqref{eq:v_grad}.
                \IF{$count$ mod $d_{push}$ = 0}
                    \STATE Push $\theta$ to PS.
                \ENDIF
                \STATE $count \leftarrow count + 1$.
            \ENDWHILE
            \STATE
            \STATE // ACTOR
            \STATE Initialize $d_{pull}$, $M$, $\Omega$.
            \STATE Initialize $\theta$, i.e. initialize $A_\theta$, $V_\theta$,$Q_\theta = V_\theta + A_\theta$ and $\pi_\theta$.
            \STATE Initialize $\{\mathcal{B}_m\}_{m=1,...,M}$ and sample $\tau \sim \Omega$.
            \STATE Initialize $count = 0$, $G = 0$.
            \STATE Initialize environment and $s = s_0$.
            \WHILE{$True$}
                \STATE Calculate $\pi_\tau(\cdot | s) \propto \exp (A_\theta / \tau)$.
                \STATE Sample $a \sim \pi_\tau(\cdot | s)$.
                \STATE $s, r, done \sim p(\cdot | s, a)$.
                \STATE $G \leftarrow G + r$.
                \IF{$done$}
                    \STATE Update $\Omega$ by $(\tau, G)$.
                    \STATE Send data to DC.
                    \STATE Sample $\tau \sim \Omega$.
                    \STATE $G \leftarrow 0$.
                    \STATE Reset the environment and $s = s_0$.
                \ENDIF
                \IF{$count \mod d_{pull}$ = 0}
                    \STATE Pull $\theta$ from PS and update $\theta$.
                \ENDIF
                \STATE $count \leftarrow count + 1$.
            \ENDWHILE
          \end{algorithmic}
        \label{alg:full_dice}
    \end{algorithm}
    \footnotetext{In practice, to save memory usage, we pre-send data back as long as the trajectory length surpasses recurrent sequence length.}
  \end{minipage}
\end{figure}
\clearpage

\section{Hyperparameters}
\label{app:hyperparameters}
\begin{table}[H]
\begin{center}
\begin{tabular}{l@{\hspace{.43cm}}l@{\hspace{.22cm}}}
\toprule
\textbf{Parameter} & \textbf{Value}  \\
\midrule
Image Size & (84, 84) \\
Grayscale & Yes \\
Num. Action Repeats & 4 \\
Num. Frame Stacks & 4 \\
Action Space & Full \\
End of Episode When Life Lost & No \\
Num. States & 200M \\
Sample Reuse & 2 \\
Num. Environments & 160 \\
Reward Shape & $\log (abs (r) + 1.0) \cdot (2 \cdot 1_{\{r \geq 0\}} - 1_{\{r < 0\}})$ \\
Reward Clip & No \\
Intrinsic Reward & No \\
Random No-ops & 30 \\
Burn-in & 40 \\
Seq-length & 80 \\
Burn-in Stored Recurrent State & Yes \\
Bootstrap & Yes \\
Batch size & 64 \\
Discount ($\gamma$) & 0.997 \\
$V$-loss Scaling ($\xi$) & 1.0 \\
$Q$-loss Scaling ($\alpha$) & 10.0 \\
$\pi$-loss Scaling ($\beta$) & 10.0 \\
Entropy Regularization & No \\
Importance Sampling Clip $\Bar{c}$ & 1.05 \\
Importance Sampling Clip $\Bar{\rho}$ & 1.05 \\
Backbone & IMPALA,deep \\
LSTM Units & 256 \\
Optimizer & Adam Weight Decay \\
Weight Decay Rate & 0.01 \\
Weight Decay Schedule & Anneal linearly to 0 \\
Learning Rate & 5e-4 \\
Warmup Steps & 4000 \\
Learning Rate Schedule & Anneal linearly to 0 \\
AdamW $\beta_1$ & 0.9 \\
AdamW $\beta_2$ & 0.98 \\
AdamW $\epsilon$ & 1e-6 \\
AdamW Clip Norm & 50.0 \\
Learner Push Model Every $n$ Steps & 25 \\
Actor Pull Model Every $n$ Steps & 64 \\
Num. Bandits & 7 \\
Bandit Learning Rate & Uniform([0.05, 0.1, 0.2]) \\
Bandit Tiling Width & Uniform([1, 2, 3]) \\
Num. Bandit Candidates & 7 \\
Bandit Value Normalization & Yes \\
Bandit UCB Scaling & 1.0 \\
Bandit Search Range for $1 / \tau$ & [0.0, 50.0] \\
\bottomrule
\end{tabular}
\caption{Hyperparameters for Atari Experiments.}
\end{center}
\label{tab:fixed_model_hyperparameters_atari}
\end{table}
\clearpage

\section{Comparison to Baseline}
\label{app:baseline}

\scriptsize
\begin{center}
\begin{tabular}{ccccccc}
\toprule
Games & RND & HUMAN & BASELINE & HNS(\%) & DiCE & HNS(\%) \\
\midrule
Scale  &     &       & 200M   &       &  200M   &        \\
\midrule
 alien  & 227.8 & 7127.8 & \textbf{13720} & \textbf{195.54}  & 10641 & 150.92 \\
 amidar & 5.8   & \textbf{1719.5}  & 560   & 32.34 & 653.9  & 37.82 \\
 assault & 222.4 & 742  & 16228  & 3080.37  & \textbf{36251} & \textbf{6933.91}  \\
 asterix & 210   & 8503.3 & 213580 & 2572.80 & \textbf{851210} & \textbf{10261.30} \\
 asteroids & 719 & 47388.7 & 18621   & 38.36 & \textbf{759170} & \textbf{1625.15} \\
 atlantis & 12850 & 29028.1 & 3211600 & 19772.10 & \textbf{3670700} & \textbf{22609.89} \\
 bank heist & 14.2 & 753.1  & 895.3   & 119.24 & \textbf{1381}   & \textbf{184.98} \\
 battle zone & 236 & 37187.5 & 70137   & 189.17 & \textbf{130410} & \textbf{352.28} \\
 beam rider & 363.9 & 16926.5 & 34920   & 208.64 & \textbf{104030}  & \textbf{625.90} \\
 berzerk & 123.7 & \textbf{2630.4}  & 1648   & 60.81  & 1222   & 43.81 \\
 bowling & 23.1 & 160.7   & 162.4     & 101.24 & \textbf{176.4}     & \textbf{111.41} \\
 boxing  & 0.1  & 12.1    & 98.3   & 818.33 & \textbf{99.9}   & \textbf{831.67} \\
 breakout & 1.7 & 30.5    & 624.3  & 2161.81 & \textbf{696}    & \textbf{2410.76} \\
 centipede & 2090.9 & 12017  & \textbf{102600} & \textbf{1012.57} & 38938  & 371.21 \\
 chopper command & 811 & 7387.8 & \textbf{616690} & \textbf{9364.42} & 41495 & 618.60 \\
 crazy climber & 10780.5 & 36829.4 & \textbf{161250} & \textbf{600.70} & 157250 & 584.73 \\
 defender & 2874.5 & 18688.9 & 421600 & 2647.75 & \textbf{837750} & \textbf{5279.21} \\
 demon attack & 152.1 & 1971 & 291590 & 16022.76 & \textbf{549450} & \textbf{30199.46} \\
 double dunk & -18.6 & -16.4 & 20.25 & 1765.91 & \textbf{23}   & \textbf{1890.91} \\
 enduro      & 0   & 860.5 & 10019 & 1164.32 & \textbf{14317}  & \textbf{1663.80} \\
 fishing derby & -91.7 & -38.8 & \textbf{53.24} & \textbf{273.99} & 48.8 & 265.60 \\
 freeway       & 0     & 29.6  & 3.46   & 11.69 & \textbf{33.7}   & \textbf{113.85} \\
 frostbite    & 65.2  & 4334.7 & 1583 & 35.55 & \textbf{8102} & \textbf{188.24} \\
 gopher  & 257.6 & 2412.5 & 188680 & 8743.90 & \textbf{454150} & \textbf{21063.27} \\
 gravitar & 173 & 3351.4  & 4311  & 130.19  & \textbf{6150}  & \textbf{188.05} \\
 hero   & 1027 & \textbf{30826.4} & 24236 & 77.88 & 17655 & 55.80 \\
 ice hockey & -11.2 & 0.9 & \textbf{1.56}  & \textbf{105.45} & -8.1  & 25.62 \\
 jamesbond  & 29    & 302.8 & 12468 & 4543.10 & \textbf{567020} & \textbf{207082.18} \\
 kangaroo   & 52    & 3035 & 5399 & 179.25 & \textbf{14286} & \textbf{477.17} \\
 krull     & 1598   & 2665.5 & \textbf{64347} & \textbf{5878.13} & 11104 & 890.49 \\
 kung fu master & 258.5 & 22736.3 & 124630.1 & 553.31 & \textbf{1270800} & \textbf{5652.43} \\
 montezuma revenge & 0  & \textbf{4753.3}  & 2488.4  & 52.35 & 2528  & 53.18 \\
 ms pacman  & 307.3 & 6951.6    & \textbf{7579}  & \textbf{109.44} & 4296  & 60.03 \\
 name this game & 2292.3 & 8049 & \textbf{32098} &\textbf{517.76}  & 30037 & 481.95 \\
 phoenix & 761.5 & 7242.6 & 498590 & 7681.23 & \textbf{597580} & \textbf{9208.60} \\
 pitfall & -229.4 & \textbf{6463.7} & -17.8 & 3.16 & -21.8 & 3.10 \\
 pong    & -20.7  & 14.6  & 20.39  & 116.40 & \textbf{21}    & \textbf{118.13} \\
 private eye & 24.9 & \textbf{69571.3} & 134.1  & 0.16 & 15095  & 21.67 \\
 qbert  & 163.9 & 13455.0 & \textbf{21043} & \textbf{157.09} & 19091 & 142.40 \\
 riverraid & 1338.5 & \textbf{17118.0} & 11182 & 62.38 & 17081 & 99.77 \\
 road runner & 11.5 & 7845  & \textbf{251360} & \textbf{3208.64} & 57102 & 728.80 \\
 robotank   & 2.2   & 11.9 & 10.44  & 84.95 & \textbf{69.7}  & \textbf{695.88} \\
 seaquest  & 68.4 & \textbf{42054.7} & 11862  & 28.09 & 2728  & 6.33 \\
 skiing & -17098  & \textbf{-4336.9} & -12730 & 34.23 & -9327 & 60.90 \\
 solaris & 1236.3 & \textbf{12326.7} & 2319 & 9.76 & 3653  & 21.79 \\
 space invaders & 148 & 1668.7 & 3031 & 189.58 & \textbf{105810} & \textbf{6948.25} \\
 star gunner & 664 & 10250 & 337150 & 3510.18 & \textbf{358650} & \textbf{3734.47} \\
 surround    & -10 & \textbf{6.5}   & -10  & 0.00 & -9.8  & 1.21 \\
 tennis  & -23.8   & -8.3 & -21.05 & 17.74 & \textbf{23.7}  & \textbf{306.45} \\
 time pilot & 3568 & 5229.2 & 84341 & 4862.62 & \textbf{150930} & \textbf{8871.35} \\
 tutankham  & 11.4 & 167.6  & \textbf{381} & \textbf{236.62} & 380.3 & 236.17 \\
 up n down  & 533.4 & 11693.2 & 416020 & 3723.06 & \textbf{907170} & \textbf{8124.13} \\
 venture    & 0     & 1187.5  & 0  & 0.00 & \textbf{1969}  & \textbf{165.81} \\
 video pinball & 0 & 17667.9  & 297920 & 1686.22 & \textbf{673840} & \textbf{3813.92} \\
 wizard of wor & 563.5 & 4756.5 & \textbf{26008} & \textbf{606.83} & 21325  & 495.15 \\
 yars revenge & 3092.9 & 54576.9 & 76903.5 & 143.37 & \textbf{84684}  & \textbf{158.48} \\
 zaxxon       & 32.5   & 9173.3 & 46070.8  & 503.66 & \textbf{62133}  & \textbf{679.38} \\
\hline
MEAN HNS(\%) &     0.00 & 100.00   & & 1929.95 &      & \textbf{6456.63} \\
\hline
MEDIAN HNS(\%) & 0.00   & 100.00   & & 195.54 &      & \textbf{477.17} \\
\bottomrule
\end{tabular}
\end{center}
\normalsize
\clearpage

\scriptsize
\begin{center}
\begin{tabular}{ccccccccccc}
\toprule
Games & RND & HWR & BASELINE & SABER(\%) & DiCE & SABER(\%) \\
\midrule
Scale  &     &       & 200M   &     &  200M   &  \\
\midrule
 alien              & 227.8     & \textbf{251916}    & 13729  & 5.36                                &10641             &4.14    \\
 amidar             & 5.8       & \textbf{104159}    & 560    & 0.53         &653.9             &0.62            \\
 assault            & 222.4     & 8647               & 16228  & 189.99      &\textbf{36251}             &\textbf{200.00}   \\
 asterix            & 210       & \textbf{1000000}   & 213580 & 21.34         &851210            &85.12  \\
 asteroids          & 719       & \textbf{10506650}  & 18621  & 0.17         &759170            &7.22   \\
 atlantis           & 12850     & \textbf{10604840}  & 3211600 & 30.20         &3670700               &34.53   \\
 bank heist         & 14.2      & \textbf{82058}     & 895.3 & 1.07         &1381              &1.67  \\
 battle zone        & 236       &\textbf{801000}    & 70137 & 8.73                  &130410            &16.26   \\
 beam rider         & 363.9     & \textbf{999999}    & 34920 & 3.46         &104030            &10.37    \\
 berzerk            & 123.7     & \textbf{1057940}   & 1648 & 0.14         &1222              &0.10        \\
 bowling            & 23.1      & \textbf{300}       & 162.4 & 50.31         &176.4             &55.36   \\
 boxing             & 0.1       & \textbf{100}       & 98.3 & 98.30    &99.9              &99.90    \\
 breakout           & 1.7       & \textbf{864}       & 624.3 & 72.20         &696               &80.52   \\
 centipede          & 2090.9    & \textbf{1301709}   & 102600 & 7.73         &38938             &2.84 \\
 chopper command    & 811       & \textbf{999999}    & 616690 & 61.64         &41495             &4.07 \\
 crazy climber      & 10780.5   & \textbf{219900}    & 161250 & 71.95                  &157250            &70.04        \\
 defender           & 2874.5    & \textbf{6010500}   & 421600 & 6.97         &837750            &13.90        \\
 demon attack       & 152.1     & \textbf{1556345}   & 291590 & 18.73         &549450            &35.30        \\
 double dunk        & -18.6     & 21                 & 20.25 & 98.11         &\textbf{23}            &\textbf{105.05}\\
 enduro             & 0         & 9500               & 10019 & 105.46         &\textbf{14317}             &\textbf{150.71}\\
 fishing derby      & -91.7     & \textbf{71}        & 53.24 & 89.08         &48.8              &86.36 \\
 freeway            & 0         & \textbf{38}        & 3.46 & 9.11         &33.7              &88.68  \\
 frostbite          & 65.2      & \textbf{454830}    & 1583 & 0.33         &8102              &1.77         \\          
 gopher             & 257.6     & 355040             & 188680 & 53.11         &\textbf{454150}            &\textbf{127.94}\\
 gravitar           & 173       & \textbf{162850}    & 4311 & 2.54         &6150              &3.67         \\
 hero               & 1027      & \textbf{1000000}   & 24236 & 2.32         &17655             &1.66\\
 ice hockey         & -11.2     & \textbf{36}        & 1.56 & 27.03                  &-8.1              &6.57\\
 jamesbond          & 29        & 45550              & 12468 & 27.33         &\textbf{567020}            &\textbf{200.00} \\
 kangaroo           & 52        & \textbf{1424600}   & 5399 & 0.38         &14286             &1.00        \\
 krull              & 1598      & \textbf{104100}    & 64347 & 61.22                  &11104             &9.27                \\
 kung fu master     & 258.5     & 1000000   & 124630.1 & 12.44         &\textbf{1270800}               &\textbf{127.09}         \\
 montezuma revenge  &0          & \textbf{1219200}   & 2488.4 & 0.20         &2528              &0.21        \\
 ms pacman          & 307.3     & \textbf{290090}    & 7579 & 2.51         &4296              &1.38      \\
 name this game     & 2292.3    & 25220              & \textbf{32098} & \textbf{130.00}         & 30037             & 121.01  \\
 phoenix            & 761.5     & \textbf{4014440}   & 498590 & 12.40         &597580            &14.87            \\
 pitfall            & -229.4    & \textbf{114000}    & -17.8 & 0.19       &-21.8             &0.18      \\
 pong               & -20.7     & \textbf{21}        & 20.39 & 98.54   &\textbf{21}            &\textbf{100.00}    \\
 private eye        & 24.9      & \textbf{101800}    & 134.1 & 0.11         &15095             &14.81            \\
 qbert              & 163.9     & \textbf{2400000}   & 21043 & 0.87         &19091             &0.79       \\
 riverraid          & 1338.5    & \textbf{1000000}   & 11182 & 0.99         &17081             &1.58      \\
 road runner        & 11.5      & \textbf{2038100}   & 251360 & 12.33         &57102             &2.80             \\
 robotank           & 2.2       & \textbf{76}        & 10.44 & 11.17                  &69.7              &91.46 \\
 seaquest           & 68.4      & \textbf{999999}    & 11862 & 1.18                  &2728              &0.27 \\
 skiing             & -17098    & \textbf{-3272}     & -12730 & 31.59         &-9327             &56.21        \\
 solaris            & 1236.3    & \textbf{111420}    & 2319 & 0.98         &3653              &2.19         \\
 space invaders     & 148       & \textbf{621535 }   & 3031 & 0.46         &105810            &17.00              \\
 star gunner        & 664       & 77400              & 337150 & 200.00         &\textbf{358650}            &\textbf{200.00}   \\
 surround           & -10       & \textbf{9.6}                & -10 & 0.00         &-9.8              &1.02\\
 tennis             & -23.8     & 21                 & -21.05 & 6.14         &\textbf{23.7}              &\textbf{106.03}\\
 time pilot         & 3568      & 65300              & 84341 & 130.84         &\textbf{150930}            &\textbf{200.00}   \\
 tutankham          & 11.4      & \textbf{5384}      & 381 & 6.88         &380.3             &6.87             \\
 up n down          & 533.4     & 82840              & 416020 & 200.00         &\textbf{907170}            &\textbf{200.00} \\
 venture            & 0         & \textbf{38900}     & 0 & 0.00         &1969              &5.06                   \\
 video pinball      & 0         & \textbf{89218328}  & 297920 & 0.33         &673840            &0.76                   \\\
 wizard of wor      & 563.5     & \textbf{395300}    & 26008 & 6.45         &21325             &5.26                  \\
 yars revenge       & 3092.9    & \textbf{15000105}  & 76903.5 & 0.49         &84684             &0.54                   \\
 zaxxon             & 32.5      & \textbf{83700}     & 46070.8 & 55.03                  &62133             &74.22   \\
\hline
MEAN SABER(\%) &     0.00 & \textbf{100.00}   &         & 35.91 &      &50.11\\
\hline
MEDIAN SABER(\%) & 0.00   & \textbf{100.00}   &         & 8.73 &      &13.90  \\
\bottomrule
\end{tabular}
\end{center}
\clearpage

\section{Atari Results}
\label{app:atari_results}

\subsection{Atari Games Table of Scores Based on Human Average Records}
Random scores and average human's scores are from \citep{agent57}.
Rainbow's scores are from \citep{rainbow}.
IMPALA's scores are from \citep{impala}.
LASER's scores are from \citep{laser}, no sweep at 200M.
As there are many versions of R2D2 and NGU, we use original papers'.
R2D2's scores are from \citep{r2d2}.
NGU's scores are from \citep{ngu}.
Agent57's scores are from \citep{agent57}.

\tiny
\begin{center}
\setlength{\tabcolsep}{2.0pt}
\begin{tabular}{ccccccccccc}
\toprule
Games & RND & HUMAN & RAINBOW & HNS(\%) & IMPALA & HNS(\%) & LASER & HNS(\%) & DiCE & HNS(\%) \\
\midrule
Scale  &     &       & 200M   &       &  200M    &        & 200M   &
       &  200M   &  \\
\midrule
 alien  & 227.8 & 7127.8 & 9491.7 & 134.26 & 15962.1  & 228.03 & \textbf{35565.9} & \textbf{512.15} & 10641 & 150.92 \\
 amidar & 5.8   & 1719.5 & \textbf{5131.2} & \textbf{299.08} & 1554.79  & 90.39  & 1829.2  & 106.4  & 653.9  & 37.82 \\
 assault & 222.4 & 742   & 14198.5 & 2689.78 & 19148.47 & 3642.43  & 21560.4 & 4106.62 & \textbf{36251} & \textbf{6933.91}  \\
 asterix & 210   & 8503.3 & 428200 & 5160.67 & 300732   & 3623.67  & 240090  & 2892.46 & \textbf{851210} & \textbf{10261.30} \\
 asteroids & 719 & 47388.7 & 2712.8 & 4.27   & 108590.05 & 231.14  & 213025  &  454.91 & \textbf{759170} & \textbf{1625.15}\\
 atlantis & 12850 & 29028.1 & 826660 & 5030.32 & 849967.5 & 5174.39 & 841200 & 5120.19 & \textbf{3670700} & \textbf{22609.89} \\
 bank heist & 14.2 & 753.1  & 1358   & 181.86  & 1223.15  & 163.61  & 569.4  & 75.14   & \textbf{1381}   & \textbf{184.98} \\
 battle zone & 236 & 37187.5 & 62010 & 167.18  & 20885    & 55.88  & 64953.3 & 175.14  & \textbf{130410} & \textbf{352.28} \\
 beam rider & 363.9 & 16926.5 & 16850.2 & 99.54 & 32463.47 & 193.81 & 90881.6 & 546.52 & \textbf{104030}  & \textbf{625.90} \\
 berzerk & 123.7 & 2630.4  & 2545.6   & 96.62  & 1852.7   & 68.98  & \textbf{25579.5} & \textbf{1015.51} & 1222   & 43.81 \\
 bowling & 23.1 & 160.7   & 30   & 5.01        & 59.92    & 26.76  & 48.3    & 18.31   & \textbf{176.4}     & \textbf{111.41} \\
 boxing  & 0.1  & 12.1    & 99.6 & 829.17      & 99.96    & 832.17 & \textbf{100}   & \textbf{832.5}     & 99.9   & 831.67 \\
 breakout & 1.7 & 30.5    & 417.5 & 1443.75    & \textbf{787.34}   & \textbf{2727.92} & 747.9 & 2590.97  & 696    & 2410.76 \\
 centipede & 2090.9 & 12017 & 8167.3 & 61.22   & 11049.75 & 90.26   & \textbf{292792} & \textbf{2928.65} & 38938  & 371.21 \\
 chopper command & 811 & 7387.8 & 16654 & 240.89 & 28255  & 417.29  & \textbf{761699} & \textbf{11569.27} & 41495 & 618.60 \\
 crazy climber & 10780.5 & 36829.4 & \textbf{168788.5} & \textbf{630.80} & 136950 & 503.69 & 167820  & 626.93 & 157250 & 584.73 \\
 defender & 2874.5 & 18688.9 & 55105 & 330.27 & 185203 & 1152.93 & 336953  & 2112.50   & \textbf{837750} & \textbf{5279.21} \\
 demon attack & 152.1 & 1971 & 111185 & 6104.40 & 132826.98 & 7294.24 & 133530 & 7332.89 & \textbf{549450} & \textbf{30199.46} \\
 double dunk & -18.6 & -16.4 & -0.3   & 831.82  & -0.33     & 830.45  & 14     & 1481.82 & \textbf{23}   & \textbf{1890.91} \\
 enduro      & 0   & 860.5 & 2125.9 & 247.05  & 0       & 0.00     & 0    & 0.00       & \textbf{14317}  & \textbf{1663.80} \\
 fishing derby & -91.7 & -38.8 & 31.3 & 232.51  & 44.85   & 258.13    & 45.2   & 258.79  & \textbf{48.8} & \textbf{265.60} \\
 freeway       & 0     & 29.6  & \textbf{34} & \textbf{114.86}  & 0     & 0.00       & 0    & 0.00       & 33.7   & 113.85 \\
 frostbite     & 65.2  & 4334.7 & \textbf{9590.5} & \textbf{223.10} & 317.75 & 5.92     & 5083.5 & 117.54  & 8102 & 188.24 \\
 gopher  & 257.6 & 2412.5 & 70354.6 & 3252.91    & 66782.3 & 3087.14 & 114820.7 & 5316.40 & \textbf{454150} & \textbf{21063.27} \\
 gravitar & 173 & 3351.4  & 1419.3  & 39.21   & 359.5      & 5.87    & 1106.2   & 29.36   & \textbf{6150}  & \textbf{188.05} \\
 hero   & 1027 & 30826.4 & \textbf{55887.4} & \textbf{184.10}   & 33730.55  & 109.75   & 31628.7 & 102.69   & 17655 & 55.80 \\
 ice hockey & -11.2 & 0.9 & 1.1    & 101.65   & 3.48      & 121.32   & \textbf{17.4}    & \textbf{236.36}   & -8.1  & 25.62 \\
 jamesbond  & 29    & 302.8 & 19809 & 72.24   & 601.5     & 209.09   & 37999.8 & 13868.08 & \textbf{567020} & \textbf{207082.18} \\
 kangaroo   & 52    & 3035 & \textbf{14637.5} & \textbf{488.05} & 1632    & 52.97    & 14308   & 477.91     & 14286 & 477.17 \\
 krull     & 1598   & 2665.5 & 8741.5  & 669.18 & 8147.4  & 613.53   & 9387.5  &  729.70  & \textbf{11104} & \textbf{890.49} \\
 kung fu master & 258.5 & 22736.3 & 52181 & 230.99 & 43375.5 & 191.82 & 607443 & 2701.26  & \textbf{1270800} & \textbf{5652.43} \\
 montezuma revenge & 0  & \textbf{4753.3}  & 384   & 8.08   & 0       & 0.00   & 0.3    & 0.01     & 2528  & 53.18 \\
 ms pacman  & 307.3 & 6951.6   & 5380.4  & 76.35   & \textbf{7342.32} & \textbf{105.88} & 6565.5 & 94.19    & 4296  & 60.03 \\
 name this game & 2292.3 & 8049 & 13136 & 188.37   & 21537.2 & 334.30 & 26219.5 & 415.64  & \textbf{30037} & \textbf{481.95} \\
 phoenix & 761.5 & 7242.6  & 108529 & 1662.80   & 210996.45  & 3243.82 & 519304 & 8000.84 & \textbf{597580} & \textbf{9208.60} \\
 pitfall & -229.4 & \textbf{6463.7} & 0      & 3.43      & -1.66      & 3.40    & -0.6   & 3.42    & -21.8 & 3.10 \\
 pong    & -20.7  & 14.6   & 20.9   & 117.85    & 20.98      & 118.07  & \textbf{21}     &  \textbf{118.13} & \textbf{21}    & \textbf{118.13} \\
 private eye & 24.9 & \textbf{69571.3} & 4234 & 6.05     & 98.5       & 0.11    & 96.3   & 0.10    & 15095  & 21.67 \\
 qbert  & 163.9 & 13455.0 & 33817.5  & 253.20   & \textbf{351200.12}  & \textbf{2641.14} & 21449.6 & 160.15 & 19091 & 142.40 \\
 riverraid & 1338.5 & 17118.0 & 22920.8 & 136.77 & 29608.05  & 179.15  & \textbf{40362.7} & \textbf{247.31} & 17081 & 99.77 \\
 road runner & 11.5 & 7845    & \textbf{62041}   & \textbf{791.85} & 57121     & 729.04  & 45289   & 578.00 & 57102 & 728.80 \\
 robotank   & 2.2   & 11.9  & 61.4   & 610.31    & 12.96     & 110.93  & 62.1    & 617.53 & \textbf{69.7}  & \textbf{695.88} \\
 seaquest  & 68.4 & \textbf{42054.7} & 15898.9 & 37.70    & 1753.2    & 4.01    & 2890.3  & 6.72   & 2728  & 6.33 \\
 skiing & -17098  & \textbf{-4336.9} & -12957.8 & 32.44  & -10180.38 & 54.21   & -29968.4 & -100.86 & -9327 & 60.90 \\
 solaris & 1236.3 & \textbf{12326.7} & 3560.3  & 20.96  & 2365      & 10.18   & 2273.5   & 9.35    & 3653  & 21.79 \\
 space invaders & 148 & 1668.7 & 18789 & 1225.82 & 43595.78 & 2857.09 & 51037.4 & 3346.45 & \textbf{105810} & \textbf{6948.25} \\
 star gunner & 664 & 10250 & 127029    & 1318.22 & 200625   & 2085.97 & 321528  & 3347.21 & \textbf{358650} & \textbf{3734.47} \\
 surround    & -10 & 6.5   & \textbf{9.7}       & \textbf{119.39}  & 7.56     & 106.42  & 8.4     & 111.52  & -9.8  & 1.21 \\
 tennis  & -23.8   & -8.3 & 0        & 153.55    & 0.55     & 157.10  & 12.2    & 232.26  & \textbf{23.7}  & \textbf{306.45} \\
 time pilot & 3568 & 5229.2 & 12926 & 563.36     & 48481.5  & 2703.84 & 105316  & 6125.34 & \textbf{150930} & \textbf{8871.35} \\
 tutankham  & 11.4 & 167.6  & 241   & 146.99     & 292.11   & 179.71  & 278.9   & 171.25  & \textbf{380.3} & \textbf{236.17} \\
 up n down  & 533.4 & 11693.2 & 125755 & 1122.08 & 332546.75 & 2975.08 & 345727 & 3093.19 & \textbf{907170} & \textbf{8124.13} \\
 venture    & 0     & 1187.5  & 5.5    & 0.46    & 0         & 0.00    & 0      & 0.00    & \textbf{1969}  & \textbf{165.81} \\
 video pinball & 0 & 17667.9  & 533936.5 & 3022.07 & 572898.27 & 3242.59 & 511835 & 2896.98 & \textbf{673840} & \textbf{3813.92} \\
 wizard of wor & 563.5 & 4756.5 & 17862.5 & 412.57 & 9157.5    & 204.96  & \textbf{29059.3} & \textbf{679.60} & 21325  & 495.15 \\
 yars revenge & 3092.9 & 54576.9 & 102557 & 193.19 & 84231.14  & 157.60 & \textbf{166292.3} & \textbf{316.99} & 84684  & 158.48 \\
 zaxxon       & 32.5   & 9173.3 & 22209.5 & 242.62 & 32935.5   & 359.96 & 41118    & 449.47 & \textbf{62133}  & \textbf{679.38} \\
\hline
MEAN HNS(\%) &     0.00 & 100.00   &         & 873.97 &         & 957.34  &        & 1741.36 &      & \textbf{6456.63} \\
\hline
MEDIAN HNS(\%) & 0.00   & 100.00   &         & 230.99 &         & 191.82  &        & 454.91  &      & \textbf{477.17} \\
\bottomrule
\end{tabular}
\end{center}
\normalsize
\clearpage

\tiny
\begin{center}
\setlength{\tabcolsep}{2.0pt}
\begin{tabular}{ccccccccc}
\toprule
 Games & R2D2 & HNS(\%) & NGU & HNS(\%) & AGENT57 & HNS(\%) & DiCE & HNS(\%) \\
\midrule
Scale  & 10B   &        & 35B &         & 100B     &        & 200M & \\
\midrule
 alien  & 109038.4 & 1576.97 & 248100 & 3592.35 & \textbf{297638.17} & \textbf{4310.30} & 10641 & 150.92 \\
 amidar & 27751.24 & 1619.04 & 17800  & 1038.35 & \textbf{29660.08}  & \textbf{1730.42} & 653.9  & 37.82 \\
 assault & \textbf{90526.44} & \textbf{17379.53} & 34800 & 6654.66 & 67212.67 & 12892.66 & 36251 & 6933.91  \\
 asterix & \textbf{999080}   & \textbf{12044.30} & 950700 & 11460.94 & 991384.42 & 11951.51 & 851210 & 10261.30 \\
 asteroids & 265861.2 & 568.12 & 230500 & 492.36   & 150854.61 & 321.70   & \textbf{759170} & \textbf{1625.15}\\
 atlantis & 1576068   & 9662.56 & 1653600 & 10141.80 & 1528841.76 & 9370.64 & \textbf{3670700} & \textbf{22609.89} \\
 bank heist & \textbf{46285.6} & \textbf{6262.20} & 17400   & 2352.93  & 23071.5    & 3120.49 & 1381   & 184.98 \\
 battle zone & 513360 & 1388.64 & 691700  & 1871.27  & \textbf{934134.88}  & \textbf{2527.36} & 130410 & 352.28 \\
 beam rider & 128236.08 & 772.05 & 63600  & 381.80   & \textbf{300509.8}   & \textbf{1812.19} & 104030  & 625.90 \\
 berzerk & 34134.8      & 1356.81 & 36200 & 1439.19  & \textbf{61507.83}   & \textbf{2448.80} & 1222   & 43.81 \\
 bowling & 196.36       & 125.92  & 211.9 & 137.21   & \textbf{251.18}     & \textbf{165.76}  & 176.4     & 111.41 \\
 boxing  & 99.16        & 825.50  & 99.7  & 830.00   & \textbf{100}        & \textbf{832.50}  & 99.9   & 831.67 \\
 breakout & \textbf{795.36}      & \textbf{2755.76} & 559.2 & 1935.76  & 790.4      & 2738.54 & 696    & 2410.76 \\
 centipede & 532921.84  & 5347.83 & \textbf{577800} & \textbf{5799.95} & 412847.86  & 4138.15 & 38938  & 371.21 \\
 chopper command & 960648 & 14594.29 & \textbf{999900} & \textbf{15191.11} & \textbf{999900} & \textbf{15191.11} & 41495 & 618.60 \\
 crazy climber & 312768   & 1205.59  & 313400 & 1208.11  & \textbf{565909.85} & \textbf{2216.18} & 157250 & 584.73 \\
 defender & 562106        & 3536.22  & 664100 & 4181.16  & 677642.78 & 4266.80 & \textbf{837750} & \textbf{5279.21} \\
 demon attack & 143664.6  & 7890.07  & 143500 & 7881.02  & 143161.44 & 7862.41 & \textbf{549450} & \textbf{30199.46} \\
 double dunk & 23.12      & 1896.36  & -14.1  & 204.55   & \textbf{23.93}     & \textbf{1933.18} & 23   & 1890.91 \\
 enduro      & 2376.68    & 276.20   & 2000   & 232.42   & 2367.71   & 275.16  & \textbf{14317}  & \textbf{1663.80} \\
 fishing derby & 81.96    & 328.28   & 32     & 233.84   & \textbf{86.97}     & \textbf{337.75}  & 48.8 & 265.60 \\
 freeway       & \textbf{34}       & \textbf{114.86}   & 28.5   & 96.28    & 32.59     & 110.10  & 33.7   & 113.85 \\
 frostbite    & 11238.4  & 261.70   & 206400 & 4832.76  & \textbf{541280.88} & \textbf{12676.32} & 8102 & 188.24 \\
 gopher  & 122196        & 5658.66  & 113400 & 5250.47  & 117777.08 & 5453.59  & \textbf{454150} & \textbf{21063.27} \\
 gravitar & 6750         & 206.93   & 14200  & 441/32   & \textbf{19213.96}  & \textbf{599.07}   & 6150  & 188.05 \\
 hero   & 37030.4        & 120.82   & 69400  & 229.44   & \textbf{114736.26} & \textbf{381.58}    & 17655 & 55.80 \\
 ice hockey & \textbf{71.56}      & \textbf{683.97}   & -4.1   & 58.68    & 63.64     & 618.51   & -8.1  & 25.62 \\
 jamesbond  & 23266      & 8486.85  & 26600  & 9704.53  & 135784.96 & 49582.16 & \textbf{567020} & \textbf{207082.18} \\
 kangaroo   & 14112      & 471.34   & \textbf{35100}  & \textbf{1174.92}  & 24034.16  & 803.96   & 14286 & 477.17 \\
 krull     & 145284.8    & 13460.12 & 127400 & 11784.73 & \textbf{251997.31} & \textbf{23456.61} & 11104 & 890.49 \\
 kung fu master & 200176 & 889.40   & 212100 & 942.45   & 206845.82 & 919.07   & \textbf{1270800} & \textbf{5652.43} \\
 montezuma revenge & 2504 & 52.68   & \textbf{10400}  & \textbf{218.80}   & 9352.01   & 196.75   & 2528  & 53.18 \\
 ms pacman  & 29928.2     & 445.81  & 40800  & 609.44   & \textbf{63994.44}  & \textbf{958.52}   & 4296  & 60.03 \\
 name this game & 45214.8 & 745.61  & 23900  & 375.35   & \textbf{54386.77}  & \textbf{904.94}   & 30037 & 481.95 \\
 phoenix & 811621.6       & 125.11  & \textbf{959100} & \textbf{14786.66} & 908264.15 & 14002.29 & 597580 & 9208.60 \\
 pitfall & 0              & 3.43    & 7800   & 119.97   & \textbf{18756.01}  & \textbf{283.66}   & -21.8 & 3.10 \\
 pong    & \textbf{21}             & \textbf{118.13}  & 19.6   & 114.16   & 20.67     & 117.20   & \textbf{21}    & \textbf{118.13} \\
 private eye & 300        & 0.40    & \textbf{100000} & \textbf{143.75}   & 79716.46  & 114.59   & 15095  & 21.67 \\
 qbert  & 161000          & 1210.10 & 451900 & 3398.79  & \textbf{580328.14} & \textbf{4365.06}  & 19091 & 142.40 \\
 riverraid & 34076.4      & 207.47  & 36700  & 224.10   & \textbf{63318.67}  & \textbf{392.79}   & 17081 & 99.77 \\
 road runner & \textbf{498660}     & \textbf{6365.59} & 128600 & 1641.52  & 243025.8  & 3102.24  & 57102 & 728.80 \\
 robotank   & \textbf{132.4}       & \textbf{1342.27} & 9.1    & 71.13    & 127.32    & 1289.90  & 69.7  & 695.88 \\
 seaquest  & 999991.84    & 2381.55 & \textbf{1000000} & \textbf{2381.57} & 999997.63 & 2381.56  & 2728  & 6.33 \\
 skiing & -29970.32       & -100.87 & -22977.9 & -46.08 & \textbf{-4202.6}   & \textbf{101.05}   & -9327 & 60.90 \\
 solaris & 4198.4         & 26.71   & 4700     & 31.23  & \textbf{44199.93}  & \textbf{387.39}   & 3653  & 21.79 \\
 space invaders & 55889   & 3665.48 & 43400    & 2844.22 & 48680.86 & 3191.48  & \textbf{105810} & \textbf{6948.25}\\
 star gunner & 521728     & 5435.68 & 414600   & 4318.13 & \textbf{839573.53} & \textbf{8751.40} & 358650 & 3734.47\\
 surround    & \textbf{9.96}       & \textbf{120.97}  & -9.6     & 2.42    & 9.5       & 118.18  & -9.8  & 1.21 \\
 tennis  & \textbf{24}             & \textbf{308.39}  & 10.2     & 219.35  & 23.84     & 307.35  & 23.7  & 306.45 \\
 time pilot & 348932    & 20791.28 & 344700  & 20536.51 & \textbf{405425.31} & \textbf{24192.24} & 150930 & 8871.35 \\
 tutankham  & 393.64    & 244.71   & 191.1   & 115.04   & \textbf{2354.91}   & \textbf{1500.33}  & 380.3 & 236.17 \\
 up n down  & 542918.8  & 4860.17  & 620100  & 5551.77  & 623805.73 & 5584.98  & \textbf{907170} & \textbf{8124.13} \\
 venture    & 1992      & 167.75   & 1700    & 143.16   & \textbf{2623.71}   & \textbf{220.94}   & 1969  & 165.81 \\
 video pinball & 483569.72 & 2737.00 & 965300 & 5463.58 & \textbf{992340.74} & \textbf{5616.63}  & 673840 & 3813.92 \\
 wizard of wor & 133264 & 3164.81  & 106200  & 2519.35  & \textbf{157306.41} & \textbf{3738.20}  & 21325  & 495.15 \\
 yars revenge & 918854.32 & 1778.73 & 986000 & 1909.15  & \textbf{998532.37} & \textbf{1933.49}  & 84684  & 158.48 \\
 zaxxon & 181372        & 1983.85  & 111100  & 1215.07  & \textbf{249808.9}  & \textbf{2732.54}  & 62133  & 679.38 \\
\hline
MEAN HNS(\%) &               & 3374.31  &         &  3169.90 &           & 4763.69  &     & \textbf{6456.63} \\
\hline
MEDIAN HNS(\%) &            & 1342.27   &         & 1208.11  &           & \textbf{1933.49}  &     & 477.17 \\
\bottomrule
\end{tabular}
\end{center}
\clearpage

\subsection{Atari Games Table of Scores Based on SABER}

\normalsize
Human World Records (HWR) are from \citep{saber}.

\begin{table}[!hb]
\tiny
\begin{center}
\setlength{\tabcolsep}{1.0pt}
\begin{tabular}{ccccccccccc}
\toprule
Games & RND & HWR & RAINBOW & SABER(\%) & IMPALA & SABER(\%) & LASER & SABER(\%) & DiCE & SABER(\%) \\
\midrule
Scale  &     &       & 200M   &       &  200M    &        & 200M   & &  200M   &  \\
\midrule
 alien              & 227.8     & \textbf{251916}    & 9491.7   &3.68    & 15962.1    & 6.25       & 976.51  & 14.04                                &10641             &4.14    \\
 amidar             & 5.8       & \textbf{104159}    & 5131.2   &4.92    & 1554.79    & 1.49       & 1829.2  & 1.75                                 &653.9             &0.62            \\
 assault            & 222.4     & 8647               & 14198.5  &165.90  & 19148.47   & 200.00     & 21560.4 & 200.00                               &\textbf{36251}             &\textbf{200.00}   \\
 asterix            & 210       & \textbf{1000000}   & 428200   &42.81   & 300732     & 30.06      & 240090  & 23.99                                &851210            &85.12  \\
 asteroids          & 719       & \textbf{10506650}  & 2712.8   &0.02    & 108590.05  & 1.03       & 213025  & 2.02                                 &759170            &7.22   \\
 atlantis           & 12850     & \textbf{10604840}  & 826660   &7.68    & 849967.5   & 7.90       & 841200  & 7.82                                 &3670700               &34.53   \\
 bank heist         & 14.2      & \textbf{82058}     & 1358     &1.64    & 1223.15    & 1.47       & 569.4   & 0.68                                 &1381              &1.67  \\
 battle zone        & 236       &\textbf{801000}    & 62010    &7.71    & 20885      & 2.58       & 64953.3 & 8.08                                           &130410            &16.26   \\
 beam rider         & 363.9     & \textbf{999999}    & 16850.2  &1.65    & 32463.47   & 3.21       & 90881.6 & 9.06                                 &104030            &10.37    \\
 berzerk            & 123.7     & \textbf{1057940}            & 2545.6   &0.23    & 1852.7     & 0.16       & 25579.5 & 2.41                        &1222              &0.10        \\
 bowling            & 23.1      & \textbf{300}       & 30       &2.49    & 59.92      & 13.30      & 48.3    & 9.10                                 &176.4             &55.36   \\
 boxing             & 0.1       & \textbf{100}                & 99.6     &99.60   & 99.96      & 99.96      & \textbf{100}     & \textbf{100.00}    &99.9              &99.90    \\
 breakout           & 1.7       & \textbf{864}                & 417.5    &48.22   & 787.34     & 91.11      & 747.9   & 86.54                       &696               &80.52   \\
 centipede          & 2090.9    & \textbf{1301709}   & 8167.3   &0.47    & 11049.75   & 0.69       & 292792  & 22.37                                &38938             &2.84 \\
 chopper command    & 811       & \textbf{999999}             & 16654    &1.59    & 28255      & 2.75       & 761699  & 76.15                       &41495             &4.07 \\
 crazy climber      & 10780.5   & \textbf{219900}    & 168788.5 &75.56   & 136950     & 60.33      & 167820  & 75.10                                         &157250            &70.04        \\
 defender           & 2874.5    & \textbf{6010500}   & 55105    &0.87    & 185203     & 3.03       & 336953  & 5.56                                 &837750            &13.90        \\
 demon attack       & 152.1     & \textbf{1556345}   & 111185   &7.13    & 132826.98  & 8.53       & 133530  & 8.57                                 &549450            &35.30        \\
 double dunk        & -18.6     & 21                 & -0.3     &46.21   & -0.33      & 46.14      & 14      & 82.32                                &\textbf{23}            &\textbf{105.05}\\
 enduro             & 0         & 9500               & 2125.9   &22.38   & 0          & 0.00       & 0       & 0.00                                 &\textbf{14317}             &\textbf{150.71}\\
 fishing derby      & -91.7     & \textbf{71}        & 31.3     &75.60   & 44.85      & 83.93      & 45.2    & 84.14                                &48.8              &86.36 \\
 freeway            & 0         & \textbf{38}        & 34       &89.47   & 0          & 0.00       & 0       & 0.00                                 &33.7              &88.68  \\
 frostbite          & 65.2      & \textbf{454830}    & 9590.5   &2.09    & 317.75     & 0.06       & 5083.5  & 1.10                                 &8102              &1.77         \\          
 gopher             & 257.6     & 355040             & 70354.6  &19.76   & 66782.3    & 18.75      & 114820.7& 32.29                                &\textbf{454150}            &\textbf{127.94}\\
 gravitar           & 173       & \textbf{162850}    & 1419.3   &0.77    & 359.5      & 0.11       & 1106.2  & 0.57                                 &6150              &3.67         \\
 hero               & 1027      & \textbf{1000000}            & 55887.4  &5.49    & 33730.55   & 3.27       & 31628.7 & 3.06                        &17655             &1.66\\
 ice hockey         & -11.2     & \textbf{36}                 & 1.1      &26.06   & 3.48       & 31.10      & 17.4    & 60.59                                &-8.1              &6.57\\
 jamesbond          & 29        & 45550              & 19809    &43.45   & 601.5      & 1.26       & 37999.8 & 83.41                                &\textbf{567020}            &\textbf{200.00} \\
 kangaroo           & 52        & \textbf{1424600}            & 14637.5  &1.02    & 1632       & 0.11       & 14308   & 1.00                        &14286             &1.00        \\
 krull              & 1598      & \textbf{104100}    & 8741.5   &6.97    & 8147.4     & 6.39       & 9387.5  & 7.60                                          &11104             &9.27                \\
 kung fu master     & 258.5     & 1000000   & 52181    &5.19    & 43375.5    & 4.31       & 607443  & 60.73                                         &\textbf{1270800}               &\textbf{127.09}         \\
 montezuma revenge  &0          & \textbf{1219200}   & 384      &0.03    & 0          & 0.00       & 0.3     & 0.00                                 &2528              &0.21        \\
 ms pacman          & 307.3     & \textbf{290090}    & 5380.4   &1.75    & 7342.32    & 2.43       & 6565.5  & 2.16                                 &4296              &1.38      \\
 name this game     & 2292.3    & 25220              & 13136    &47.30   & 21537.2    & 83.94      & 26219.5 & 104.36                               &\textbf{30037}             &\textbf{121.01}  \\
 phoenix            & 761.5     & \textbf{4014440}   & 108529   &2.69    & 210996.45  & 5.24       & 519304  & 12.92                                &597580            &14.87            \\
 pitfall            & -229.4    & \textbf{114000}    & 0        &0.20    & -1.66      & 0.20       & -0.6    & 0.20               &-21.8             &0.18      \\
 pong               & -20.7     & \textbf{21}                 & 20.9     &99.76   & 20.98      & 99.95      & \textbf{21}      & \textbf{100.00}    &\textbf{21}            &\textbf{100.00}    \\
 private eye        & 24.9      & \textbf{101800}    & 4234     &4.14    & 98.5       & 0.07       & 96.3    & 0.07                                 &15095             &14.81            \\
 qbert              & 163.9     & \textbf{2400000}   & 33817.5  &1.40    & 351200.12  & 14.63      & 21449.6 & 0.89                                 &19091             &0.79       \\
 riverraid          & 1338.5    & \textbf{1000000}   & 22920.8  &2.16    & 29608.05   & 2.83       & 40362.7 & 3.91                                 &17081             &1.58      \\
 road runner        & 11.5      & \textbf{2038100}   & 62041    &3.04    & 57121      & 2.80       & 45289   & 2.22                                 &57102             &2.80             \\
 robotank           & 2.2       & \textbf{76}                 & 61.4     &80.22   & 12.96      & 14.58      & 62.1    & 81.17                                &69.7              &91.46 \\
 seaquest           & 68.4      & \textbf{999999}             & 15898.9  &1.58    & 1753.2     & 0.17       & 2890.3  & 0.28                                 &2728              &0.27 \\
 skiing             & -17098    & \textbf{-3272}     & -12957.8 &29.95   & -10180.38  & 50.03      & -29968.4& -93.09                               &-9327             &56.21        \\
 solaris            & 1236.3    & \textbf{111420}    & 3560.3   &2.11    & 2365       & 1.02       & 2273.5  & 0.94                                 &3653              &2.19         \\
 space invaders     & 148       & \textbf{621535 }   & 18789    &3.00    & 43595.78   & 6.99       & 51037.4 & 8.19                                 &105810            &17.00              \\
 star gunner        & 664       & 77400              & 127029   &164.67  & 200625     & 200.00     & 321528  & 200.00                               &\textbf{358650}            &\textbf{200.00}   \\
 surround           & -10       & 9.6                & \textbf{9.7}      &\textbf{100.51}  & 7.56       & 89.59      & 8.4     & 93.88              &-9.8              &1.02\\
 tennis             & -23.8     & 21                 & 0        &53.13   & 0.55       & 54.35      & 12.2    & 80.36                                &\textbf{23.7}              &\textbf{106.03}\\
 time pilot         & 3568      & 65300              & 12926    &15.16   & 48481.5    & 72.76      & 105316  & 164.82                               &\textbf{150930}            &\textbf{200.00}   \\
 tutankham          & 11.4      & \textbf{5384}      & 241      &4.27    & 292.11     & 5.22       & 278.9   & 4.98                                 &380.3             &6.87             \\
 up n down          & 533.4     & 82840              & 125755   &152.14  & 332546.75  & 200.00     & 345727  & 200.00                               &\textbf{907170}            &\textbf{200.00} \\
 venture            & 0         & \textbf{38900}     & 5.5      &0.01    & 0          & 0.00       & 0       & 0.00                                 &1969              &5.06                   \\
 video pinball      & 0         & \textbf{89218328}  & 533936.5 &0.60    & 572898.27  & 0.64       & 511835  & 0.57                                 &673840            &0.76                   \\\
 wizard of wor      & 563.5     & \textbf{395300}    & 17862.5  &4.38    & 9157.5     & 2.18       & 29059.3 & 7.22                                 &21325             &5.26                  \\
 yars revenge       & 3092.9    & \textbf{15000105}  & 102557   &0.66    & 84231.14   & 0.54       & 166292.3& 1.09                                 &84684             &0.54                   \\
 zaxxon             & 32.5      & \textbf{83700}              & 22209.5  &26.51   & 32935.5    & 39.33      & 41118   & 49.11                                &62133             &74.22   \\
\hline
MEAN SABER(\%) &     0.00 & \textbf{100.00}   &         & 28.39 &         & 29.45  &        & 36.78 &      &50.11\\
\hline
MEDIAN SABER(\%) & 0.00   & \textbf{100.00}   &         & 4.92 &         & 4.31  &        & 8.08  &      &13.90  \\
\bottomrule
\end{tabular}
\end{center}
\end{table}
\clearpage

\begin{table}[!hb]
\tiny
\begin{center}
\setlength{\tabcolsep}{1.0pt}
\begin{tabular}{ccccccccc}
\toprule
 Games & R2D2 & SABER(\%) & NGU & SABER(\%) & AGENT57 & SABER(\%) & DiCE & SABER(\%) \\
\midrule
Scale  & 10B   &        & 35B &         & 100B     &        & 200M & \\
\midrule
 alien              & 109038.4          & 43.23             & 248100          & 98.48          & \textbf{297638.17}   &\textbf{118.17}              &10641             &4.14              \\
 amidar             & 27751.24          & 26.64             & 17800           & 17.08          & \textbf{29660.08}    &\textbf{28.47}               &653.9             &0.62       \\
assault            & \textbf{90526.44} & \textbf{200.00}   & 34800           & 200.00         & 67212.67             &200.00              &36251              &200.00            \\
 asterix            & 999080   & 99.91    & 950700          & 95.07          & \textbf{991384.42}            &\textbf{99.14}              &851210             &85.12         \\
 asteroids          & 265861.2          & 2.52              & 230500          & 2.19           & 150854.61                      &1.43               &\textbf{759170}            &\textbf{7.22}     \\
 atlantis           & 1576068           & 14.76             & 1653600         & 15.49          & 1528841.76                     &14.31              &\textbf{3670700}               &\textbf{34.53}           \\
 bank heist         & \textbf{46285.6}  & \textbf{56.40}    & 17400           & 21.19          & 23071.5                        &28.10              &1381              &1.67      \\
    battle zone        & 513360            & 64.08             & 691700          & 86.35          & \textbf{934134.88}   &\textbf{116.63}           &130410    &16.26            \\
 beam rider         & 128236.08         & 12.79             & 63600           & 6.33           & \textbf{300509.8}    &\textbf{30.03}                                 &104030            &10.37           \\
 berzerk            & 34134.8           & 3.22              & 36200           & 3.41           & \textbf{61507.83}    &\textbf{5.80}                &1222              &0.10    \\
 bowling            & 196.36            & 62.57             & 211.9           & 68.18          & \textbf{251.18}      &\textbf{82.37}               &176.4             &55.36     \\
 boxing             & 99.16             & 99.16             & 99.7            & 99.70          & \textbf{100}         &\textbf{100.00}              &99.9        &99.90       \\
 breakout           & \textbf{795.36}            & \textbf{92.04}             & 559.2           & 64.65          & 790.4                &91.46                        &696         &80.52              \\
 centipede          & 532921.84         & 40.85             & \textbf{577800} & \textbf{44.30}          & 412847.86            &31.61                        &38938       &2.84       \\
 chopper command    &960648             & 96.06             &\textbf{999900}           & \textbf{99.99}          &\textbf{999900}                &\textbf{99.99}                        &41495       &4.07      \\
 crazy climber      & 312768            & 144.41            & 313400          & 144.71         &\textbf{565909.85}    &\textbf{200.00}              &157250      &70.04     \\
 defender           & 562106            & 9.31              & 664100          & 11.01          & 677642.78            &11.23                        &\textbf{837750}      &\textbf{13.90}      \\
 demon attack       & 143664.6          & 9.22              & 143500          & 9.21           & 143161.44            &9.19                         &\textbf{549450}      &\textbf{35.30}       \\
 double dunk        & 23.12             & 105.35            & -14.1           & 11.36          & \textbf{23.93}                &\textbf{107.40}                       &23      &105.05        \\
 enduro             & 2376.68           & 25.02             & 2000            & 21.05          & 2367.71              &24.92                        &\textbf{14317}       &\textbf{150.71}          \\
fishing derby      & 81.96             & 106.74            & 32              & 76.03          & \textbf{86.97}       &\textbf{109.82}               &48.8        &86.36    \\
 freeway            & \textbf{34}       & \textbf{89.47}    & 28.5            & 75.00          & 32.59                &85.76                        &33.7        &88.68                \\
 frostbite          & 11238.4           & 2.46              & 206400          & 45.37          &\textbf{541280.88}    &\textbf{119.01}              &8102        &1.77        \\
 gopher             & 122196            & 34.37             & 113400          & 31.89          & 117777.08            &33.12                        &\textbf{454150}      &\textbf{127.94}            \\
 gravitar           & 6750              & 4.04              & 14200           & 8.62           &\textbf{19213.96}     &\textbf{11.70}               &6150        &3.67     \\
 hero               & 37030.4           & 3.60              & 69400           & 6.84           &\textbf{114736.26}    &\textbf{11.38}               &17655       &1.66     \\
 ice hockey         & \textbf{71.56}    & \textbf{175.34}   &-4.1             & 15.04          & 63.64                &158.56                       &-8.1        &6.57              \\
 jamesbond          & 23266             & 51.05             & 26600           & 58.37          & 135784.96            &200.00                       &\textbf{567020}      &\textbf{200.00}      \\
 kangaroo           & 14112             & 0.99              & \textbf{35100}  & \textbf{2.46}           &24034.16              &1.68                         &14286       &1.00        \\
 krull              & 145284.8          & 140.18            & 127400          & 122.73         & \textbf{251997.31}   &\textbf{200.00}                                &11104       &9.27        \\
 kung fu master     & 200176            & 20.00             & 212100          & 21.19          & 206845.82            &20.66                        &\textbf{1270800}         &\textbf{127.09}          \\
 montezuma revenge  & 2504              & 0.21              & \textbf{10400}  & \textbf{0.85}           &9352.01               &0.77                         &2528        &0.21         \\
 ms pacman          & 29928.2           & 10.22             & 40800           & 13.97          & \textbf{63994.44}    &\textbf{21.98}               &4296        &1.38    \\
 name this game     & 45214.8           & 187.21            & 23900           & 94.24          &\textbf{54386.77}     &\textbf{200.00}              &30037      &121.01    \\
 phoenix            & 811621.6          & 20.20             & \textbf{959100} & \textbf{23.88}          &908264.15             &22.61                                 &597580     &14.87          \\
 pitfall            & 0        & 0.20              & 7800            & 7.03           &\textbf{18756.01}     &\textbf{16.62}      &-21.8      &0.18     \\
 pong               & \textbf{21}       & \textbf{100.00}   & 19.6            & 96.64          & 20.67                &99.21                        &\textbf{21}     &\textbf{100.00}   \\
 private eye        & 300               & 0.27              & \textbf{100000} & \textbf{98.23}          & 79716.46             &78.30                        &15095      &14.81         \\
 qbert              & 161000            & 6.70              & 451900          & 18.82          &\textbf{580328.14}    &\textbf{24.18}               &19091      &0.79    \\
 riverraid          & 34076.4           & 3.28              & 36700           & 3.54           & \textbf{63318.67}    &\textbf{6.21}                &17081      &1.58    \\
 road runner        & \textbf{498660}            & \textbf{24.47}             & 128600          & 6.31           & 243025.8             &11.92                        &57102      &2.80      \\
 robotank           & \textbf{132.4}    & \textbf{176.42}   & 9.1             & 9.35           &127.32                &169.54                       &69.7       &91.46          \\
 seaquest           & 999991.84         & 100.00            & \textbf{1000000}& \textbf{100.00}         &999997.63             &100.00                       &2728       &0.27      \\
 skiing             & -29970.32         & -93.10            & -22977.9        & -42.53         & \textbf{-4202.6}     &\textbf{93.27}               &-9327      &56.21     \\
 solaris            & 4198.4            & 2.69              & 4700            & 3.14           & \textbf{44199.93}    &\textbf{38.99}               &3653       &2.19     \\
 space invaders     & 55889             & 8.97              & 43400           & 6.96           & 48680.86             &7.81                         &\textbf{105810}     &\textbf{17.00}     \\
 star gunner        & 521728            & 200.00            & 414600          & 200.00         &\textbf{839573.53}    &\textbf{200.00}              &358650     &200.00    \\
 surround           & \textbf{9.96}     & \textbf{101.84}   & -9.6            & 2.04           & 9.5                  &99.49                        &-9.8       &1.02                \\
 tennis             & \textbf{24}       & \textbf{106.70}   & 10.2            & 75.89          & 23.84                &106.34                       &23.7       &106.03         \\
 time pilot         & 348932            & 200.00            & 344700          & 200.00         &\textbf{405425.31}    &\textbf{200.00}                                &150930     &200.00    \\
 tutankham          & 393.64            & 7.11              & 191.1           & 3.34           & \textbf{2354.91}     &\textbf{43.62}               &380.3      &6.87    \\
up n down          & 542918.8          & 200.00            & 620100          & 200.00         & 623805.73            &200.00                     &\textbf{907170}         &\textbf{200.00}   \\
 venture            & 1992              & 5.12              & 1700            & 4.37           &\textbf{2623.71}      &\textbf{6.74}             &1969       &5.06  \\
 video pinball      & 483569.72         & 0.54              & 965300          & 1.08           &\textbf{992340.74}    &\textbf{1.11}             &673840         &0.76  \\
 wizard of wor      & 133264            & 33.62             & 106200          & 26.76          &\textbf{157306.41}    &\textbf{39.71}            &21325      &5.26  \\
 yars revenge       & 918854.32         & 6.11              & 986000          & 6.55           &\textbf{998532.37}    &\textbf{6.64}             &84684      &0.54  \\
 zaxxon             & 181372            & 200.00            & 111100          & 132.75         &\textbf{249808.9}     &\textbf{200.00}           &62133      &74.22  \\
\hline
MEAN SABER(\%)        &                   & 60.43             &                 &  50.47         &                      & \textbf{76.26}  &                  &50.11 \\
\hline
MEDIAN SABER(\%)      &                   & 33.62             &                 & 21.19          &                      & \textbf{43.62}  &                  &13.90  \\
\bottomrule
\end{tabular}
\end{center}
\end{table}
\clearpage

\renewcommand{\thesubfigure}{\arabic{subfigure}.}
\begin{figure} 
    \subfigure[alien]{
    \includesvg[width=0.22\columnwidth,inkscapelatex=false]{nips2/alien.svg}
    }
    \subfigure[amidar]{
    \includesvg[width=0.22\columnwidth,inkscapelatex=false]{nips2/amidar.svg}
    }
    \subfigure[assault]{
    \includesvg[width=0.22\columnwidth,inkscapelatex=false]{nips2/assault.svg}
    }
    \subfigure[asterix]{
    \includesvg[width=0.22\columnwidth,inkscapelatex=false]{nips2/asterix.svg}
    }
\end{figure}

\vspace{-0.0cm}
\begin{figure}
    \subfigure[asteroids]{
    \includesvg[width=0.22\columnwidth,inkscapelatex=false]{nips2/asteroids.svg}
    }
    \subfigure[atlantis]{
    \includesvg[width=0.22\columnwidth,inkscapelatex=false]{nips2/atlantis.svg}
    }
    \subfigure[bank\_heist]{
    \includesvg[width=0.22\columnwidth,inkscapelatex=false]{nips2/bank_heist.svg}
    }
    \subfigure[battle\_zone]{
    \includesvg[width=0.22\columnwidth,inkscapelatex=false]{nips2/battle_zone.svg}
    }
\end{figure}

\vspace{-0.0cm}
\begin{figure}
    \subfigure[beam\_rider]{
    \includesvg[width=0.22\columnwidth,inkscapelatex=false]{nips2/beam_rider.svg}
    }
    \subfigure[berzerk]{
    \includesvg[width=0.22\columnwidth,inkscapelatex=false]{nips2/berzerk.svg}
    }
    \subfigure[bowling]{
    \includesvg[width=0.22\columnwidth,inkscapelatex=false]{nips2/bowling.svg}
    }
    \subfigure[boxing]{
    \includesvg[width=0.22\columnwidth,inkscapelatex=false]{nips2/boxing.svg}
    }
\end{figure}

\vspace{-0.0cm}
\begin{figure}
    \subfigure[breakout]{
    \includesvg[width=0.22\columnwidth,inkscapelatex=false]{nips2/breakout.svg}
    }
    \subfigure[centipede]{
    \includesvg[width=0.22\columnwidth,inkscapelatex=false]{nips2/centipede.svg}
    }
    \subfigure[chopper\_command]{
    \includesvg[width=0.22\columnwidth,inkscapelatex=false]{nips2/chopper_command.svg}
    }
    \subfigure[crazy\_climber]{
    \includesvg[width=0.22\columnwidth,inkscapelatex=false]{nips2/crazy_climber.svg}
    }
\end{figure}

\vspace{-0.0cm}
\begin{figure}
    \subfigure[defender]{
    \includesvg[width=0.22\columnwidth,inkscapelatex=false]{nips2/defender.svg}
    }
    \subfigure[demon\_attack]{
    \includesvg[width=0.22\columnwidth,inkscapelatex=false]{nips2/demon_attack.svg}
    }
    \subfigure[double\_dunk]{
    \includesvg[width=0.22\columnwidth,inkscapelatex=false]{nips2/double_dunk.svg}
    }
    \subfigure[enduro]{
    \includesvg[width=0.22\columnwidth,inkscapelatex=false]{nips2/enduro.svg}
    }
\end{figure}
\clearpage
\vspace{-0.0cm}
\begin{figure}
    \subfigure[fishing\_derby]{
    \includesvg[width=0.22\columnwidth,inkscapelatex=false]{nips2/fishing_derby.svg}
    }
    \subfigure[freeway]{
    \includesvg[width=0.22\columnwidth,inkscapelatex=false]{nips2/freeway.svg}
    }
    \subfigure[frostbite]{
    \includesvg[width=0.22\columnwidth,inkscapelatex=false]{nips2/frostbite.svg}
    }
    \subfigure[gopher]{
    \includesvg[width=0.22\columnwidth,inkscapelatex=false]{nips2/gopher.svg}
    }
\end{figure}

\vspace{-0.0cm}
\begin{figure}
    \subfigure[gravitar]{
    \includesvg[width=0.22\columnwidth,inkscapelatex=false]{nips2/gravitar.svg}
    }
    \subfigure[hero]{
    \includesvg[width=0.22\columnwidth,inkscapelatex=false]{nips2/hero.svg}
    }
    \subfigure[ice\_hockey]{
    \includesvg[width=0.22\columnwidth,inkscapelatex=false]{nips2/ice_hockey.svg}
    }
    \subfigure[jamesbond]{
    \includesvg[width=0.22\columnwidth,inkscapelatex=false]{nips2/jamesbond.svg}
    }
\end{figure}

\vspace{-0.0cm}
\begin{figure}
    \subfigure[kangaroo]{
    \includesvg[width=0.22\columnwidth,inkscapelatex=false]{nips2/kangaroo.svg}
    }
    \subfigure[krull]{
    \includesvg[width=0.22\columnwidth,inkscapelatex=false]{nips2/krull.svg}
    }
    \subfigure[kung\_fu\_master]{
    \includesvg[width=0.22\columnwidth,inkscapelatex=false]{nips2/kung_fu_master.svg}
    }
    \subfigure[montezuma\_revenge]{
    \includesvg[width=0.22\columnwidth,inkscapelatex=false]{nips2/montezuma_revenge.svg}
    }
\end{figure}

\vspace{-0.0cm}
\begin{figure}
    \subfigure[ms\_pacman]{
    \includesvg[width=0.22\columnwidth,inkscapelatex=false]{nips2/ms_pacman.svg}
    }
    \subfigure[name\_this\_game]{
    \includesvg[width=0.22\columnwidth,inkscapelatex=false]{nips2/name_this_game.svg}
    }
    \subfigure[phoenix]{
    \includesvg[width=0.22\columnwidth,inkscapelatex=false]{nips2/phoenix.svg}
    }
    \subfigure[pitfall]{
    \includesvg[width=0.22\columnwidth,inkscapelatex=false]{nips2/pitfall.svg}
    }
\end{figure}

\vspace{-0.0cm}
\begin{figure}
    \subfigure[pong]{
    \includesvg[width=0.22\columnwidth,inkscapelatex=false]{nips2/pong.svg}
    }
    \subfigure[private\_eye]{
    \includesvg[width=0.22\columnwidth,inkscapelatex=false]{nips2/private_eye.svg}
    }
    \subfigure[qbert]{
    \includesvg[width=0.22\columnwidth,inkscapelatex=false]{nips2/qbert.svg}
    }
    \subfigure[riverraid]{
    \includesvg[width=0.22\columnwidth,inkscapelatex=false]{nips2/riverraid.svg}
    }
\end{figure}
\clearpage
\vspace{-0.0cm}
\begin{figure}
    \subfigure[road\_runner]{
    \includesvg[width=0.22\columnwidth,inkscapelatex=false]{nips2/road_runner.svg}
    }
    \subfigure[robotank]{
    \includesvg[width=0.22\columnwidth,inkscapelatex=false]{nips2/robotank.svg}
    }
    \subfigure[seaquest]{
    \includesvg[width=0.22\columnwidth,inkscapelatex=false]{nips2/seaquest.svg}
    }
    \subfigure[skiing]{
    \includesvg[width=0.22\columnwidth,inkscapelatex=false]{nips2/skiing.svg}
    }
\end{figure}

\vspace{-0.0cm}
\begin{figure}
    \subfigure[solaris]{
    \includesvg[width=0.22\columnwidth,inkscapelatex=false]{nips2/solaris.svg}
    }
    \subfigure[space\_invader]{
    \includesvg[width=0.22\columnwidth,inkscapelatex=false]{nips2/space_invader.svg}
    }
    \subfigure[star\_gunner]{
    \includesvg[width=0.22\columnwidth,inkscapelatex=false]{nips2/star_gunner.svg}
    }
    \subfigure[surround]{
    \includesvg[width=0.22\columnwidth,inkscapelatex=false]{nips2/surround.svg}
    }
\end{figure}

\vspace{-0.0cm}
\begin{figure}
    \subfigure[tennis]{
    \includesvg[width=0.22\columnwidth,inkscapelatex=false]{nips2/tennis.svg}
    }
    \subfigure[time\_pilot]{
    \includesvg[width=0.22\columnwidth,inkscapelatex=false]{nips2/time_pilot.svg}
    }
    \subfigure[tutankham]{
    \includesvg[width=0.22\columnwidth,inkscapelatex=false]{nips2/tutankham.svg}
    }
    \subfigure[up\_n\_down]{
    \includesvg[width=0.22\columnwidth,inkscapelatex=false]{nips2/up_n_down.svg}
    }
\end{figure}

\vspace{-0.0cm}
\begin{figure}
    \subfigure[venture]{
    \includesvg[width=0.22\columnwidth,inkscapelatex=false]{nips2/venture.svg}
    }
    \subfigure[video\_pinball]{
    \includesvg[width=0.22\columnwidth,inkscapelatex=false]{nips2/video_pinball.svg}
    }
    \subfigure[wizard\_of\_wor]{
    \includesvg[width=0.22\columnwidth,inkscapelatex=false]{nips2/wizard_of_wor.svg}
    }
    \subfigure[yars\_revenge]{
    \includesvg[width=0.22\columnwidth,inkscapelatex=false]{nips2/yars_revenge.svg}
    }
\end{figure}

\vspace{-0.0cm}
\begin{figure}
    \subfigure[zaxxon]{
    \includesvg[width=0.22\columnwidth,inkscapelatex=false]{nips2/zaxxon.svg}
    } 
\end{figure}
\clearpage

\end{appendices}
\end{document}


\appendix

\section{CASA Algorithm}
\label{app:casa}

Our training algorithm is shown in \textbf{Algorithm} \ref{alg:casa}, where we describe how learners and actors work asynchronously.

\begin{figure}[ht]
  \centering
  \begin{minipage}{.7\linewidth}
    \begin{algorithm}[H]
      \caption{CASA Algorithm.}  
          \begin{algorithmic}
            \STATE Initialize Parameter Server (PS) and Data Collector (DC).
            \STATE
            \STATE // LEARNER
            \STATE Initialize $d_{push}$.
            \STATE Initialize $\theta$, i.e. initialize $A_\theta$, $V_\theta$, and calculate $Q_\theta$ and $\pi_\theta$ as in \eqref{eq:casa}.
            \STATE Initialize $count = 0$.
            \WHILE{$True$}
                \STATE Load data from DC.
                \STATE Estimate $vs$ by \eqref{eq:dr-v} and estimate $qs$ by \eqref{eq:dr-q}.
                \STATE Update $\theta$ by \eqref{eq:grad_all}.
                \IF{$count$ mod $d_{push}$ = 0}
                    \STATE Push $\theta$ to PS.
                \ENDIF
                \STATE $count \leftarrow count + 1$.
            \ENDWHILE
            \STATE
            \STATE // ACTOR
            \STATE Initialize $d_{pull}$, $M$.
            \STATE Initialize $\theta$, i.e. initialize $A_\theta$, $V_\theta$, and calculate $Q_\theta$ and $\pi_\theta$ as in \eqref{eq:casa}.
            \STATE Initialize $\{\mathcal{B}_m\}_{m=1,...,M}$ and sample $\tau$ as in \textbf{Algorithm} \ref{alg:bva}.
            \STATE Initialize $count = 0$, $G = 0$.
            \WHILE{$True$}
                \STATE Calculate $\pi(\tau)(\cdot | s)$.
                \STATE Sample $a \sim \pi(\tau)(\cdot | s)$.
                \STATE $s, r, done \sim p(\cdot | s, a)$.
                \STATE $G \leftarrow G + r$.
                \IF{$done$}
                    \STATE Update $\{\mathcal{B}_m\}_{m=1,...,M}$ with $(\tau, G)$ as in \textbf{Algorithm} \ref{alg:bva}.
                    \STATE Send data to DC and reset the environment. \footnotemark
                    \STATE $G \leftarrow 0$.
                    \STATE Sample $\tau$ as in \textbf{Algorithm} \ref{alg:bva}
                \ENDIF
                \IF{$count \mod d_{pull}$ = 0}
                    \STATE Pull $\theta$ from PS and update $\theta$.
                \ENDIF
                \STATE $count \leftarrow count + 1$.
            \ENDWHILE
          \end{algorithmic}
        \label{alg:casa}
    \end{algorithm}
    \footnotetext{In practice, to save memory usage, we pre-send data back as long as the trajectory length surpasses recurrent sequence length.}
  \end{minipage}
\end{figure}


\section{Bandits Vote Algorithm}
\label{app:bva}

In practice, we use a Bandits Vote Algorithm (BVA) to adaptively control the entropy of the \textit{average} policy $\textbf{H}[\pi (\Omega)]$.

The algorithm is shown in \textbf{Algorithm} \ref{alg:bva}.

\begin{figure}[ht]
  \centering
  \begin{minipage}{.7\linewidth}
    \begin{algorithm}[H]
      \caption{Bandits Vote Algorithm.}  
          \begin{algorithmic}
            \FOR{$m=1,...,M$}
                \STATE Sample $mode \sim \{argmax, random\}$, sample $lr$, sample $width$.
                \STATE Initialize $B_m = Bandit(mode, l, r, acc, width, lr, \textbf{w}, \textbf{N}, d)$.
            \ENDFOR
            \WHILE{$True$}  
                \FOR{$m=1,...,M$}
                    \STATE Evaluate $\mathcal{B}_m$ by \eqref{eq:bandit_eval}.
                    \STATE Sample candidates $c_{m, 1}, ..., c_{m, d}$ by $mode$ and \eqref{eq:bandit_score} from $\mathcal{B}_m$.
                \ENDFOR
                \STATE Sample $x$ from $\{c_{m, i}\}_{m=1,...,M; i=1,...,d}$.
                \STATE Execute $x$ and receive $g$.
                \FOR{$m=1,...,M$}
                    \STATE Update $\mathcal{B}_m$ with $(x, g)$ by \eqref{eq:bandit_update}.
                \ENDFOR
            \ENDWHILE
          \end{algorithmic}  
        \label{alg:bva}
    \end{algorithm}
  \end{minipage}
\end{figure}

Let's firstly define a bandit as $\mathcal{B} = Bandit(mode, l, r, acc, width, lr, \textbf{w}, \textbf{N}, d)$.
\begin{itemize}
    \item $mode$ is the mode of sampling, with two choices, $argmax$ and $random$.
    \item $l$ is the left boundary of $\mathcal{B}$, and each $x$ is clipped to $x = \max \{x, l\}$.
    \item $r$ is the right boundary of $\mathcal{B}$, and each $x$ is clipped to $x = \min \{x, r\}$.
    \item $acc$ is the accuracy, where each $x$ is located in the $\lfloor (\min\{\max\{x, l\}, r\} - l) / acc \rfloor$th tile.
    \item $width$ is the tile coding width, where the value of the $i$th tile is estimated by the average of $\{\textbf{w}_{i-width},...,\textbf{w}_{i+width}\}$.
    \item $lr$ is the learning rate.
    \item $\textbf{w}$ is a vector in $\mathbf{R}^{\lfloor (r-l) / acc \rfloor}$, which represents the weight of each tile.
    \item $\textbf{N}$ is a vector in $\mathbf{R}^{\lfloor (r-l) / acc \rfloor}$, which counts the number of sampling of each tile.
    \item $d$ is an integer, which represents how many candidates is provided by $\mathcal{B}$ when sampling.
\end{itemize}

During the evaluation process, we evaluate the value of the $i$th tile by
\begin{equation}
\label{eq:bandit_eval}
V_i = \frac{1}{2 * width + 1}\sum_{k=i-width}^{i+width} \textbf{w}_k.\footnotemark
\end{equation}
\footnotetext{For the boundary case, we change the summation indexes and the denominator accordingly.}

During the training process, for each sample $(x, g)$, where $g$ is the target value. Since $x$ locates in the $i$th tile, we update $\mathcal{B}$ by
\begin{equation}
\label{eq:bandit_update}
\left\{
\begin{aligned}
&i = \lfloor (\min\{\max\{x, l\}, r\} - l) / acc \rfloor, \\
&\textbf{w}_j 
\leftarrow \textbf{w}_j + lr * \left(g - V_i\right), j = i - width, ..., i + width, \\
& \textbf{N}_i \leftarrow \textbf{N}_i + 1.
\end{aligned}
\right.
\end{equation}

During the sampling process, we firstly evaluate $\mathcal{B}$ by \eqref{eq:bandit_eval} and get $(V_1, ..., V_{\lfloor (r-l) / acc \rfloor})$.
We calculate the score of $i$th tile by
\begin{equation}
\label{eq:bandit_score}
score_i = \frac{V_i - \mu(\{V_j\}_{j=1,...,\lfloor(r-l)/acc\rfloor})}{\sigma(\{V_j\}_{j=1,...,\lfloor(r-l)/acc\rfloor})} + c \cdot \sqrt{\frac{\log (1 + \sum_j \textbf{N}_j)}{1 + \textbf{N}_i}}.
\end{equation}
For different $mode$s, we sample the candidates by the following mechanism,
\begin{itemize}
    \item if $mode$ = $argmax$, find tiles with top-$d$ $score$s, then sample $d$ candidates from these tiles, one uniformly from a tile;
    \item if $mode$ = $random$, sample $d$ tiles with $score$s as the logits without replacement, then sample $d$ candidates from these tiles, one uniformly from a tile;
\end{itemize}

In practice, we define a set of bandits $\{\mathcal{B}_m\}_{m=1,...,M}$.
At each step, we sample $d$ candidates $\{c_{m, i}\}_{i=1,...,d}$ from each $B_m$, so we have a set of $m \times d$ candidates $\{c_{m, i}\}_{m=1,...,M; i=1,...,d}$.
Then we sample uniformly from these $m \times d$ candidates to get $x$. 
At last, we transform the selected $x$ to $\tau$ by $\tau = \frac{1}{\exp (x) - 1}$.
When we receive $(\tau, g)$, we transform $\tau$ to $x$ by $x = \log (1 + 1 / \tau)$. 
Then we update each $B_m$ by \eqref{eq:bandit_update}.

\section{Proofs}
\label{app:proof}
\newtheorem{Lemma_app}{Lemma}
\newtheorem{Theorem_app}{Theorem}
\newtheorem*{Remark_app}{Remark}

\begin{Lemma_app}
Let 
$g \in \textbf{C}^{1}(\mathbf{R}^{n}): \mathbf{R}^{n} \to \mathbf{R}^{n}, \ f \in \textbf{C}^{1}(\mathbf{R}^{n+k}): \mathbf{R}^{n+k} \to \mathbf{R}^{n}.
$\\
If
$
\nabla_x g(x) = \nabla_x f(x, y)$, for $\forall x\in \mathbf{R}^{n}, y\in \mathbf{R}^k,
$
then $\exists$ $c \in \textbf{C}^{1}(\mathbf{R}^{k}): \mathbf{R}^{k} \to \mathbf{R}^{n}$, s.t. $f(x, y) = g(x) + c(y)$.
\label{lemma_app:func_sep}
\end{Lemma_app}
\begin{proof}

Let $\Tilde{f}(x, y) = f(x, y) - g(x)$.

Since $\nabla_x g(x) = \nabla_x f(x, y)$, we have 
$$
\nabla_x \Tilde{f} = 0, \ for \ \forall x\in \mathbf{R}^{n}, y\in \mathbf{R}^k.
$$

So $\Tilde{f}$ is a constant function w.r.t $x$, which can be denoted as $c(y) = \Tilde{f}(x, y)$.

Hence, $f(x, y) = g(x) + c(y)$.
\end{proof}

\begin{Lemma_app}
Define $\pi = softmax(A / \tau)$, then $\nabla \log \pi = (\textbf{1} - \pi) \frac{\nabla A}{\tau}$. 
Denote $sg$ to be stop gradient and define $\Bar{A} = A - \textbf{E}_\pi [A]$, $Q = \Bar{A} + sg(V)$, then $\nabla Q = (\textbf{1} - \pi) \nabla A$.
\label{lemma_app:vannila_grad}
\end{Lemma_app}
\begin{proof}

As $Q = \Bar{A} + sg(V) = A - sg(\pi)\cdot A + sg(V)$, it's obvious that $\nabla Q = (\textbf{1} - \pi) \nabla A$.

For $\log \pi$, it's a typical derivative of cross entropy, so we have $\nabla \log \pi = (\textbf{1} - \pi) \nabla (A / \tau) = (\textbf{1} - \pi) \frac{\nabla A}{\tau}$.
\end{proof}

\begin{Lemma_app}
Define $\Bar{A}= A - \textbf{E}_\pi[A]$, $Q = \Bar{A} + sg(V), \pi = softmax(A / \tau)$, then 
$$
\textbf{E}_\pi \left[ (Q - V) \nabla \log \pi \right]
= - \tau \nabla \textbf{H}[\pi].
$$
\label{lemma_app:eqiv_pg_ent}
\end{Lemma_app}
\begin{proof}
Since 
$$
\pi = \exp(A / \tau) / Z,\ Z = \int_\mathcal{A} \exp(A / \tau),
$$
we have 
$$
A = \tau \log \pi + \tau \log Z.
$$
Based on the observation that $\textbf{E}_\pi \left[ f(s) \nabla \log \pi (\cdot | s) \right] = 0$, 
we have 
$$\textbf{E}_\pi \left[ \textbf{E}_\pi[A] \cdot \nabla \log \pi \right] = 0,$$ 
$$\textbf{E}_\pi \left[ \log Z \cdot \nabla \log \pi \right] = 0.$$

On the one hand,
$$
\begin{aligned}
    \textbf{E}_\pi \left[ (Q - V) \nabla \log \pi \right]
    &= \textbf{E}_\pi \left[ A \nabla \log \pi \right] 
    - \textbf{E}_\pi \left[ \textbf{E}_\pi[A] \cdot \nabla \log \pi \right] \\
    &= \tau \textbf{E}_\pi \left[ \log \pi \nabla \log \pi \right]
    + \tau \textbf{E}_\pi \left[ \log Z \cdot \nabla \log \pi \right] \\
    &= \tau \textbf{E}_\pi \left[ \log \pi \nabla \log \pi \right].
\end{aligned}
$$

On the other hand, 
$$
\begin{aligned}
    \nabla \textbf{H} [\pi] 
    &= - \nabla \int_\mathcal{A} \pi_i \log \pi_i \\
    &= - \int_\mathcal{A}  \nabla \pi_i \cdot \log \pi_i - \int_\mathcal{A} \pi_i \nabla \log \pi_i  \\
    &= - \int_\mathcal{A}  \pi_i \nabla \log \pi_i \cdot \log \pi_i - \int_\mathcal{A}  \pi_i \frac{\nabla \pi_i}{\pi_i} \\
    &= - \textbf{E}_\pi \left[ \log \pi \nabla \log \pi \right].
\end{aligned}
$$
Hence, $
\textbf{E}_\pi \left[ (Q - V) \nabla \log \pi \right]
= - \tau \nabla \textbf{H}[\pi]
$.
\end{proof}

\begin{Lemma_app}
Let $v \in \mathbf{R}^{|\mathcal{A}|}$ to be a vector. 
Define 
$
    \pi (\tau) = \exp (v / \tau) / Z,\  Z = \int_\mathcal{A}  \exp(v / \tau).
$
Let $\Omega$ to be a probability measure supported on $[K, +\infty]$,
then $f(\Omega) = \textbf{E}_{\tau \sim \Omega} [\textbf{E}_{\pi(\tau)} [v]]$ satisfies Lipschitz-1 condition with Wasserstein-1 metric.
\label{lemma_app:lips}
\end{Lemma_app}
\begin{proof}
Without loss of generality, we assume $v_1 \geq v_2 \geq ... \geq v_{|\mathcal{A}|}$.

For any $\tau \in [0, +\infty)$, since
$$
\begin{aligned}
    \pi (\tau) = \exp (v / \tau) / Z,\  Z = \int_\mathcal{A}  \exp(v / \tau),
\end{aligned}
$$
we have
$$
    v = \tau \log \pi (\tau) + \tau \log Z.
$$

Denote $\Tilde{v}_j = v_j / \tau$. \\
Since 
$$
\frac{\partial \log \pi_i}{\partial \Tilde{v}_j} = 1_{i=j} - \pi_j,
$$
we have
$$
\begin{aligned}
    \frac{\partial \log \pi_i}{\partial \tau} 
    &= \sum_j \frac{\partial \log \pi_i}{\partial \Tilde{v}_j} \cdot \frac{\partial \Tilde{v}_j}{\partial \tau} \\
    &= - \sum_j (1_{i=j} - \pi_j) \frac{v_j}{\tau^2} \\
    &= - \frac{1}{\tau^2} ( v_i - \sum_j \pi_j v_j ) \\ 
    &= - \frac{1}{\tau^2} \left( v_i - \textbf{E}_\pi [v] \right). 
\end{aligned}
$$
Therefore, we have
$$
\begin{aligned}
    \frac{\partial \pi_i}{\partial \tau} 
    &= \pi_i \frac{\partial \log \pi_i}{\partial \tau} \\
    &= - \frac{\pi_i}{\tau^2} \left( v_i - \textbf{E}_\pi [v] \right).
\end{aligned}
$$
Let $f(\tau) = v \cdot \pi(\tau)$, then
$$
\frac{\partial f}{\partial \tau} = - \frac{1}{\tau^2} \sum_i v_i \pi_i \left( v_i - \textbf{E}_\pi [v] \right).
$$
Since $\sum_i \textbf{E}_\pi [v] \pi_i \left( v_i - \textbf{E}_\pi [v] \right) = 0$,
we know
$$
\begin{aligned}
    \frac{\partial f}{\partial \tau} 
    &= - \frac{1}{\tau^2} \sum_i \left( v_i - \textbf{E}_\pi [v] \right) \pi_i \left( v_i - \textbf{E}_\pi [v] \right) \\
    &= - \frac{1}{\tau^2} \sum_i \pi_i \left( v_i - \textbf{E}_\pi [v] \right)^2 \\
    &= - \frac{1}{\tau^2} \textbf{Var}_\pi [v].
\end{aligned}
$$
It's obvious that
$$
\left| \frac{\partial f}{\partial \tau} \right| \leq \frac{1}{K^2} |v_1 - v_{|\mathcal{A}|}|^2.
$$
Hence, for any $\tau_1, \tau_2 \in [K, +\infty]$,
$$
|v \cdot \pi(\tau_1) - v \cdot \pi(\tau_2)| \leq C | \tau_1 - \tau_2|.
$$

Finally, for any $\gamma \in \Gamma(\Omega_1, \Omega_2)$ \footnote{$\Gamma(\Omega_1, \Omega_2)$ is the collection of all measures on $[K, +\infty] \times [K, +\infty]$ with marginals $(\Omega_1, \Omega_2)$.}, we have
$$
\begin{aligned}
\left| \textbf{E}_{\tau_1 \sim \Omega_1} [v \cdot \pi (\tau_1)] - \textbf{E}_{\tau_2 \sim \Omega_2} [v \cdot \pi (\tau_2)] \right|
&= \left| \int_{[K, +\infty] \times [K, +\infty]} (v \cdot \pi (\tau_1) - v \cdot \pi (\tau_2)) d \gamma (\tau_1, \tau_2) \right| \\
&\leq \int_{[K, +\infty] \times [K, +\infty]} |v \cdot \pi (\tau_1) - v \cdot \pi (\tau_2)| d \gamma (\tau_1, \tau_2) \\
&\leq C \int_{[K, +\infty] \times [K, +\infty]} |\tau_1 - \tau_2| d \gamma (\tau_1, \tau_2).
\end{aligned}
$$

Taking infimum over $\Gamma(\Omega_1, \Omega_2)$, we have
$$
\left| \textbf{E}_{\tau_1 \sim \Omega_1} [v \cdot \pi (\tau_1)] - \textbf{E}_{\tau_2 \sim \Omega_2} [v \cdot \pi (\tau_2)] \right|
\leq C W_1 (\Omega_1, \Omega_2),
$$

which proves that $f(\Omega) = \textbf{E}_{\tau \sim \Omega} [\textbf{E}_{\pi (\tau)} [v]]$ satisfies Lipschitz-1 condition with Wasserstein-1 metric.

\end{proof}

\begin{Lemma_app}
Define $\Bar{A}= A - \textbf{E}_\pi[A]$, $Q = \Bar{A} + sg(V)$,
then operator 
$$
    \mathscr{T}(Q) \overset{def}{=} \textbf{E}_{\mu, p}   [
        Q(s_t, a_t) + \sum_{k \geq 0}  \gamma^k
        c_{[t+1:t+k-1]} \Tilde{\rho}_{t, k}
        \delta^{DR}_{t+k} V
        ]
$$
is a contraction mapping w.r.t. $Q$.
\label{lemma_app:dr_q}
\end{Lemma_app}
\begin{Remark_app}
Note that $\mathscr{T}(Q)$ is exactly \eqref{eq:dr-q}. 

Since $Q = A + sg(V)$, the gradient of $V$ is stopped when estimating $Q$, updating $Q$ won't change $V$, which is equivalent to updating $A$.
Without loss of generality, we assume $V$ is fixed as $V^*$ in the proof.
\end{Remark_app}
\begin{proof}

$\Bar{A} = A - \textbf{E}_\pi[A]$ shows $\textbf{E}_\pi[\Bar{A}] = 0$, which guarantees that no matter how we update $A$, we always have $\textbf{E}_\pi[Q] = V^*$.

Based on above observations, define 
$$
    \widetilde{\mathscr{T}}(Q) \overset{def}{=} - \textbf{E}_\pi [Q] + \mathscr{T}(Q).
$$

It's obvious that we only need to prove $\widetilde{\mathscr{T}}(Q)$ is a contraction mapping.

For brevity, we denote $$Q_t = Q(s_t, a_t), A_t = A(s_t, a_t), V^*_t = V^*(s_t).$$

Notice that $\Tilde{\rho}_{t, 0} = 1$, similar to \citep{retrace}, we can rewrite $\widetilde{\mathscr{T}}$ as 
\begin{equation}
\label{eq:dr_a_2}
\begin{aligned}
    \widetilde{\mathscr{T}}(Q)
    &= \textbf{E}_{\mu, p}   [
        A_t + \sum_{k \geq 0}  \gamma^k
        c_{[t+1:t+k-1]} \Tilde{\rho}_{t, k}
        \delta^{DR}_{t+k} V
        ] \\
    &= \textbf{E}_{\mu, p}   [
        -V^*_t + \sum_{k \geq 0}  \gamma^k
        c_{[t+1:t+k-1]} \Tilde{\rho}_{t, k}
        r_{t+k}
        + 
        \sum_{k \geq 0}  \gamma^{k+1}
        c_{[t+1:t+k-1]} \Delta_k ],
        \\
\end{aligned}
\end{equation}
where 
\begin{equation}
\label{eq:dr_delta}
    \Delta_k = \textbf{E}_{\mu, p}\left[\Tilde{\rho}_{t, k} V^*_{t+k+1} - c_{t+k} \Tilde{\rho}_{t, k+1} Q_{t+k+1} | \mathscr{F}_{t+k}\right]. \footnotemark
\footnotetext{$\mathscr{F}$ represents filtration.}
\end{equation}
By definition of $Q$,
$$
    \textbf{E}_{\mu, p}[V_{t+k+1}^*|\mathscr{F}_{t+k}] 
    = \textbf{E}_{\mu, p}[
    \textbf{E}_\pi[Q_{t+k+1}|\mathscr{F}_{t+k+1}]
    |\mathscr{F}_{t+k}], \\
$$
we can rewrite \eqref{eq:dr_delta} as
\begin{equation}
\label{eq:dr_q_delta}
\Delta_k = \textbf{E}_{\mu, p}[
(
\Tilde{\rho}_{t, k} \frac{\pi_{t+k+1}}{\mu_{t+k+1}}- c_{t+k} \Tilde{\rho}_{t, k+1} 
) Q_{t+k+1} | \mathscr{F}_{t+k}
].
\end{equation}
For any $Q^1 = A^1 + sg(V^*)$, $Q^2 = A^2 + sg(V^*)$, since
$$
\textbf{E}_{\mu, p}[
(
\Tilde{\rho}_{t, k} \frac{\pi_{t+k+1}}{\mu_{t+k+1}}- c_{t+k} \Tilde{\rho}_{t, k+1} 
) | \mathscr{F}_{t+k}
] \geq 0,
$$
by \eqref{eq:dr_a_2} \eqref{eq:dr_q_delta}, we have 
$$
        || \widetilde{\mathscr{T}}(Q^1) - \widetilde{\mathscr{T}}(Q^2) || 
        \leq \mathcal{C} || Q^1 - Q^2 ||,
$$
where 
$$
    \begin{aligned}
        \mathcal{C} 
        &= \textbf{E}_{\mu, p} [ \sum_{k \geq 0}  \gamma^{k+1} c_{[t+1:t+k-1]} 
        (
        \Tilde{\rho}_{t, k} \frac{\pi_{t+k+1}}{\mu_{t+k+1}}- c_{t+k} \Tilde{\rho}_{t, k+1} 
        ) ]
        \\
        &= \textbf{E}_{\mu, p} [1 -1 + \sum_{k \geq 0}  \gamma^{k+1} c_{[t+1:t+k-1]} 
        \left(
        \Tilde{\rho}_{t, k} - c_{t+k} \Tilde{\rho}_{t, k+1} 
        \right) ] 
        \\
        &= 1 - (1 - \gamma)  \textbf{E}_{\mu, p} [\sum_{k \geq 0} \gamma^{k}c_{[t+1:t+k-1]} \Tilde{\rho}_{t, k}  ] \\
        &\leq 1 - (1 - \gamma) < 1.
    \end{aligned}
$$
Hence, $\widetilde{\mathscr{T}}(Q)$ is a contraction mapping and converges to some fixed function, which we denote as $A^*$. So $\mathscr{T}(Q)$ is also a contraction mapping and converges to $A^*+V^*$.
\end{proof}

\begin{Lemma_app}
Define $Q = A + sg(V)$ with $\textbf{E}_\pi [A] = 0$,
then operator 
$$
    \mathscr{S}(V) \overset{def}{=} \textbf{E}_{\mu, p}  [
        V(s_t) + \sum_{k \geq 0}  \gamma^k
        c_{[t:t+k-1]} \rho_{t, k}
        \delta^{DR}_{t+k} V
        ]
$$
is a contraction mapping w.r.t. $V$.
\label{lemma_app:dr_v}
\end{Lemma_app}
\begin{Remark_app}
Note that $\mathscr{S}(V)$ is exactly \eqref{eq:dr-v}. 
\end{Remark_app}
\begin{proof}

Same as Lemma \ref{lemma_app:dr_q}, we can get
$$
    \Delta_k = \textbf{E}_{\mu, p}\left[
    \left( \rho_{t+k} - c_{t+k} \rho_{t+k+1}\right) V_{t+k+1} 
     -  c_{t+k} \rho_{t+k+1} A^*_{t+k+1} | \mathscr{F}_{t+k}\right],
$$
so we have 
$$
    \Delta^1_k - \Delta^2_k = \textbf{E}_{\mu, p}\left[ 
    \left( \rho_{t+k} - c_{t+k} \rho_{t+k+1}\right) \cdot  
   (V^1_{t+k+1} -  V^2_{t+k+1})
     | \mathscr{F}_{t+k}\right].
$$
The proof attributes to \citep{impala}.
\end{proof}

\begin{Theorem_app}
    Define $\Bar{A} = A - \textbf{E}_\pi[A]$, $Q = \Bar{A} + sg(V)$.
    Define $$
    \begin{aligned}
    &\mathscr{T}(Q) \overset{def}{=} \textbf{E}_{\mu, p}   [
        Q(s_t, a_t) + \sum_{k \geq 0}  \gamma^k
        c_{[t+1:t+k-1]} \Tilde{\rho}_{t, k}
        \delta^{DR}_{t+k} V
        ], \\
    &\mathscr{S}(V) \overset{def}{=} \textbf{E}_{\mu, p}   [
        V(s_t) + \sum_{k \geq 0}  \gamma^k
        c_{[t:t+k-1]} \rho_{t, k}
        \delta^{DR}_{t+k} V
        ], \\
    &\mathscr{U}(Q, V) = (\mathscr{T}(Q) - \textbf{E}_\pi[Q] + \mathscr{S}(V), \mathscr{S}(V)), \\
    &\mathscr{U}^{(n)}(Q, V) = \mathscr{U}(\mathscr{U}^{(n-1)}(Q, V)),
    \end{aligned}
    $$
    then $\mathscr{U}^{(n)}(Q, V) \rightarrow (Q^{\Tilde{\pi}}, V^{\Tilde{\pi}})$ that corresponds to 
    $$
        \Tilde{\pi}(a|s) = \frac
        {\min \left\{\Bar{\rho} \mu (a|s), \pi(a|s)\right\}}
        {\sum_{b \in \mathcal{A}}\min \left\{\Bar{\rho} \mu (b|s), \pi(b|s)\right\}}.
    $$ as $n \rightarrow +\infty$.
\label{thm_app:dr}
\end{Theorem_app}
\begin{Remark_app}
$\mathscr{T}(Q) - \textbf{E}_\pi[Q] + \mathscr{S}(V)$ is \textbf{exactly} how $Q$ is updated at training time. 
Since $Q = A + sg(V)$, if we apply gradient ascent on $Q$ and $V$ in directions \eqref{eq:grad_qv} respectively, change of $Q$ comes from two aspects. One comes from $\nabla \mathcal{Q}(\theta)$, which changes $A$, the other comes from $\nabla \mathcal{V}(\theta)$, which changes $V$. Because the gradient of $V$ is stopped when estimating $Q$, the latter is captured by "minus old baseline, add new baseline", which is $- \textbf{E}_\pi[Q] + \mathscr{S}(V)$ in Theorem \ref{thm_app:dr}.
\end{Remark_app}
\begin{proof}
 Define
 $$
 \begin{aligned}
        \widetilde{\mathscr{T}}(Q) &= - \textbf{E}_\pi[Q] + \mathscr{T}(Q), \\
        \widetilde{\mathscr{U}}(Q, V) &= (\widetilde{\mathscr{T}}(Q), \mathscr{S}(V)), \\
        \widetilde{\mathscr{U}}^{(n)}(Q, V) &=   \widetilde{\mathscr{U}}(\widetilde{\mathscr{U}}^{(n-1)}(Q, V)).
 \end{aligned}
 $$
By Lemma \ref{lemma_app:dr_q}, $\widetilde{\mathscr{T}}^{(n)}(Q)$ converges to some $A^*$ as $n \rightarrow \infty$. This process won't influence the estimation of $V$ as the gradient of $V$ is stopped when estimating $Q$. According to the proof, $A^*$ doesn't depend on $V$. \\
By Lemma \ref{lemma_app:dr_v}, $\mathscr{S}^{(n)}(V)$ converges to some $V^*$ as $n \rightarrow \infty$. \\
Hence, we have
$$
\widetilde{\mathscr{U}}^{(n)}(Q, V) \rightarrow (A^*, V^*)\ \ as\ \ n \rightarrow +\infty. 
$$
By definition, 
$$
\mathscr{U}(Q, V) = (\widetilde{\mathscr{T}}(Q) + \mathscr{S}(V), \mathscr{S}(V)),
$$
we can regard $\widetilde{\mathscr{T}}(Q) + \mathscr{S}(V)$ as $Q$ and regard $\mathscr{S}(V)$ as $V$, then
$$
\begin{aligned}
    \mathscr{U}^{(2)}(Q, V) 
    &= \mathscr{U}(\widetilde{\mathscr{T}}(Q) + \mathscr{S}(V), \mathscr{S}(V)) \\
    &= (\mathscr{T}(\widetilde{\mathscr{T}}(Q) + \mathscr{S}(V)) -\mathscr{S}(V) + \mathscr{S}^{(2)}(V), \mathscr{S}^{(2)}(V)) \\
    &= (\widetilde{\mathscr{T}}^{(2)}(Q) + \mathscr{S}^{(2)}(V), \mathscr{S}^{(2)}(V)).
\end{aligned}
$$
By induction, 
$$
\begin{aligned}
    \mathscr{U}^{(n)}(Q, V) &= (\widetilde{\mathscr{T}}^{(n)}(Q) + \mathscr{S}^{(n)}(V), \mathscr{S}^{(n)}(V)) \\
    &\rightarrow (A^*+V^*, V^*)\ \ as\ \ n\rightarrow + \infty.
\end{aligned}
$$
Same as \citep{impala}, 
$$
    \Tilde{\pi}(a|s) = \frac
    {\min \left\{\Bar{\rho} \mu (a|s), \pi(a|s)\right\}}
    {\sum_{b \in \mathcal{A}}\min \left\{\Bar{\rho} \mu (b|s), \pi(b|s)\right\}}.
$$ 
is the policy s.t. the Bellman equation holds, which is 
$$\textbf{E}_\mu[\rho_t (r_t + \gamma V_{t+1} - V_t) | \mathscr{F}_t] = 0,$$ and $\mathscr{U}(Q^{\Tilde{\pi}}, V^{\Tilde{\pi}}) = (Q^{\Tilde{\pi}}, V^{\Tilde{\pi}})$. \\
So we have
$(A^*+V^*, V^*) = (Q^{\Tilde{\pi}}, V^{\Tilde{\pi}}).$
\end{proof}

\section{Hyperparameters}
\label{app:hyperparameters}
\begin{table}[H]
\begin{center}
\begin{tabular}{l@{\hspace{.43cm}}l@{\hspace{.22cm}}}
\toprule
\textbf{Parameter} & \textbf{Value}  \\
\midrule
Image Size & (84, 84) \\
Grayscale & Yes \\
Num. Action Repeats & 4 \\
Num. Frame Stacks & 4 \\
Action Space & Full \\
End of Episode When Life Lost & No \\
Num. States & 200M \\
Sample Reuse & 2 \\
Num. Environments & 160 \\
Reward Shape & $\log (abs (r) + 1.0) \cdot (2 \cdot 1_{\{r \geq 0\}} - 1_{\{r < 0\}})$ \\
Reward Clip & No \\
Intrinsic Reward & No \\
Random No-ops & 30 \\
Burn-in & 40 \\
Seq-length & 80 \\
Burn-in Stored Recurrent State & Yes \\
Bootstrap & Yes \\
Batch size & 64 \\
Discount ($\gamma$) & 0.997 \\
$V$-loss Scaling ($\xi$) & 1.0 \\
$Q$-loss Scaling ($\alpha$) & 10.0 \\
$\pi$-loss Scaling ($\beta$) & 10.0 \\
Entropy Regularization & No \\
DR-Trace Importance Sampling Clip $\Bar{c}$ & 1.05 \\
DR-Trace Importance Sampling Clip $\Bar{\rho}$ & 1.05 \\
Backbone & IMPALA,deep \\
LSTM Units & 256 \\
Optimizer & Adam Weight Decay \\
Weight Decay Rate & 0.01 \\
Weight Decay Schedule & Anneal linearly to 0 \\
Learning Rate & 5e-4 \\
Warmup Steps & 4000 \\
Learning Rate Schedule & Anneal linearly to 0 \\
AdamW $\beta_1$ & 0.9 \\
AdamW $\beta_2$ & 0.98 \\
AdamW $\epsilon$ & 1e-6 \\
AdamW Clip Norm & 50.0 \\
Auxiliary Forward Dynamic Task & Yes \\
Auxiliary Inverse Dynamic Task & Yes \\
Learner Push Model Every $n$ Steps & 25 \\
Actor Pull Model Every $n$ Steps & 64 \\
Num. Bandits & 7 \\
Bandit Learning Rate & Uniform([0.05, 0.1, 0.2]) \\
Bandit Tiling Width & Uniform([1, 2, 3]) \\
Num. Bandit Candidates & 7 \\
Bandit Value Normalization & Yes \\
Bandit UCB Scaling & 1.0 \\
Bandit Search Range for $1 / \tau$ & [0.0, 50.0] \\
\bottomrule
\end{tabular}
\caption{Hyperparameters for Atari Experiments.}
\end{center}
\label{tab:fixed_model_hyperparameters_atari}
\end{table}

\section{Ablation Study}
\label{app:ablation}

\begin{figure*}[h]
\subfigure[FIGTOPCAP][Breakout]{
\includesvg[width=0.33\textwidth]{ablation/breakout.svg}
}
\subfigure[FIGTOPCAP][Hero]{
\includesvg[width=0.33\textwidth]{ablation/hero_ok.svg}
}
\subfigure[FIGTOPCAP][ChopperCommand]{
\includesvg[width=0.33\textwidth]{ablation/chopper_ok.svg}
}
\\
\subfigure[FIGTOPCAP][Gravitar]{
\includesvg[width=0.33\textwidth]{ablation/gravitar.svg}
}
\subfigure[FIGTOPCAP][Qbert]{
\includesvg[width=0.33\textwidth]{ablation/qbert_ok.svg}
}
\subfigure{
\includegraphics[trim=0 0.25cm 1cm 1cm,scale=1.0]{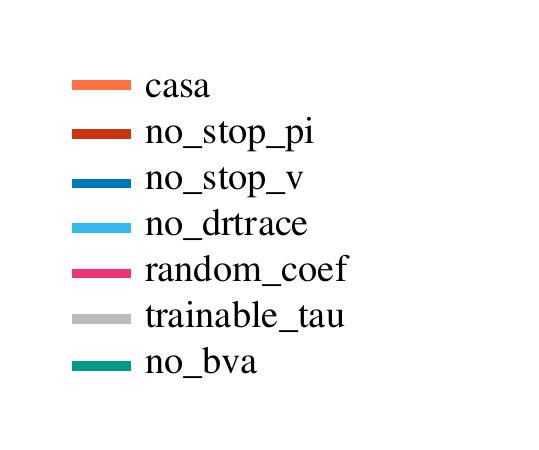}
}
\caption{Evaluation Return Curves.
The labels are identical to the ablation changes in Table \ref{tab:ablation_parameters}.
For ease of comparison, all curves are smoothed with rate 0.99.
ChopperCommand's y-axes is in a \textit{log} scale, others' are in a \textit{linear} scale.}
\label{fig:ablation}
\end{figure*}

The details of ablation changes are listed in Table \ref{tab:ablation_parameters}.
The evaluation curves are attached in Figure \ref{fig:ablation}.

\begin{table}[H]
\begin{center}
\begin{tabular}{l@{\hspace{.43cm}}l@{\hspace{.22cm}}}
\toprule
\textbf{Ablation} & \textbf{Change}  \\
\midrule
CASA 
& N/A \\
\textit{no\_stop\_$\pi$} 
    & $\Bar{A} = A - \textbf{E}_\pi[A] = A - sg(\pi) \cdot A \Rightarrow \Bar{A} = A - \pi \cdot A$ \\
\textit{no\_stop\_v} 
    & $Q = \Bar{A} + sg(V) \Rightarrow Q = \Bar{A} + V$ \\
\textit{no\_drtrace} 
    & $vs = V_t + \sum_{k \geq 0} \gamma^k c_{[t:t+k-1]} \rho_{t+k}  (r_{t+k} + \gamma V_{t+k+1} - Q_{t+k})$ \\
    & $\Rightarrow vs = V(s_t) + \sum_{k \geq 0} \gamma^k c_{[t:t+k-1]} \rho_{t+k}  (r_{t+k} + \gamma V_{t+k+1} - V_{t+k})$, \\
    & $qs = Q(s_t, a_t) + \sum_{k \geq 0}  \gamma^k
        c_{[t+1:t+k-1]} \Tilde{\rho}_{t, k}
        (r_{t+k} + \gamma V_{t+k+1} - Q_{t+k})$ \\
    & $\Rightarrow qs = Q(s_t, a_t) + \sum_{k \geq 0}  \gamma^k
        c_{[t+1:t+k-1]} \Tilde{\rho}_{t, k}
        (r_{t+k} + \gamma Q_{t+k+1} - Q_{t+k})$ \\
\textit{random\_scaling}
    & $Q$-loss Scaling $\alpha = 10.0 \Rightarrow \alpha \sim Uniform([0.0, 20.0])$ for every sample of every batch, \\
    & $\pi$-loss Scaling $\beta = 10.0 \Rightarrow \beta \sim Uniform([0.0, 20.0])$ for every sample of every batch \\
\textit{trainable\_tau}
    & $\tau$ is treated as a trainable variable rather than an input in $\pi = P(A/\tau)$, BVA is also disabled \\
\textit{no\_bva}
    & BVA is not updated, so $\tau$ is sampled from a fixed distribution $\Omega$ \\
\bottomrule
\end{tabular}
\caption{Ablation Changes.
The baseline of these ablation changes is \textbf{original CASA}.
Except for the changes listed in the table, there is no other change of CASA in each ablation case.}
\end{center}
\label{tab:ablation_parameters}
\end{table}

\begin{figure*}[h]
\subfigure[Entropy of Breakout]{
\includesvg[width=0.32\textwidth]{entropy/entropy_breakout.svg}
}
\subfigure[Entropy of Hero]{
\includesvg[width=0.34\textwidth]{entropy/entropy_hero.svg}
}
\subfigure[Entropy of ChopperCommand]{
\includesvg[width=0.33\textwidth]{entropy/entropy_chopper_command.svg}
}
\\
\subfigure[FIGTOPCAP][Entropy of Gravitar]{
\includesvg[width=0.33\textwidth]{entropy/entropy_gravitar.svg}
}
\subfigure[FIGTOPCAP][Entropy of Qbert]{
\includesvg[width=0.33\textwidth]{entropy/entropy_qbert.svg}
}
\subfigure{
\includegraphics[trim=0 0.25cm 1cm 1cm,scale=1.0]{ablation/tuli.pdf}
}
\caption{Entropy of Ablation Study.
The labels are identical to the ablation changes Table \ref{tab:ablation_parameters}.
We only draw entropy curves of CASA, \textit{trainable\_tau} and \textit{no\_bva}, because theirs entropy curves show very same trend in different games.
In general, the entropy curve of \textit{no\_bva} is above CASA along the whole training process.
The entropy curve of \textit{trainable\_tau} drops slower than CASA's.
It's above CASA's at the beginning of training for a period,
but it finally drops below CASA's at the end of training.
}
\label{fig:ablation_entropy}
\end{figure*}

\clearpage

\section{Atari Results}
\label{app:atari_results}

Random scores and average human's scores are from \citep{agent57}.
Rainbow's scores are from \citep{rainbow}.
IMPALA's scores are from \citep{impala}.
LASER's scores are from \citep{laser}, no sweep at 200M.
As there are many versions of R2D2 and NGU, we use original papers'.
R2D2's scores are from \citep{r2d2}.
NGU's scores are from \citep{ngu}.
Agent57's scores are from \citep{agent57}.

According to the videos, we observe that there exist 19 games whose results achieve \textit{Full Score} by our method.
We underline the results of these games in the table below.

\scriptsize
\begin{center}
\begin{tabular}{|c|c|c|c c|c c|c c|c c|}
\toprule
Games & RND & HUMAN & RAINBOW & HNS(\%) & IMPALA & HNS(\%) & LASER & HNS(\%) & CASA & HNS(\%) \\
\midrule
Scale  &     &       & 200M   &       &  200M    &        & 200M   &
       &  200M   &  \\
\midrule
 alien  & 227.8 & 7127.8 & 9491.7 & 134.26 & 15962.1  & 228.03 & \textbf{35565.9} & \textbf{512.15} & 10641 & 150.92 \\
 amidar & 5.8   & 1719.5 & \textbf{5131.2} & \textbf{299.08} & 1554.79  & 90.39  & 1829.2  & 106.4  & 653.9  & 37.82 \\
 assault & 222.4 & 742   & 14198.5 & 2689.78 & 19148.47 & 3642.43  & 21560.4 & 4106.62 & \textbf{36251} & \textbf{6933.91}  \\
 asterix & 210   & 8503.3 & 428200 & 5160.67 & 300732   & 3623.67  & 240090  & 2892.46 & \underline{\textbf{851210}} & \textbf{10261.30} \\
 asteroids & 719 & 47388.7 & 2712.8 & 4.27   & 108590.05 & 231.14  & 213025  &  454.91 & \underline{\textbf{759170}} & \textbf{1625.15}\\
 atlantis & 12850 & 29028.1 & 826660 & 5030.32 & 849967.5 & 5174.39 & 841200 & 5120.19 & \underline{\textbf{3670700}} & \textbf{22609.89} \\
 bank heist & 14.2 & 753.1  & 1358   & 181.86  & 1223.15  & 163.61  & 569.4  & 75.14   & \textbf{1381}   & \textbf{184.98} \\
 battle zone & 236 & 37187.5 & 62010 & 167.18  & 20885    & 55.88  & 64953.3 & 175.14  & \textbf{130410} & \textbf{352.28} \\
 beam rider & 363.9 & 16926.5 & 16850.2 & 99.54 & 32463.47 & 193.81 & 90881.6 & 546.52 & \textbf{104030}  & \textbf{625.90} \\
 berzerk & 123.7 & 2630.4  & 2545.6   & 96.62  & 1852.7   & 68.98  & \textbf{25579.5} & \textbf{1015.51} & 1222   & 43.81 \\
 bowling & 23.1 & 160.7   & 30   & 5.01        & 59.92    & 26.76  & 48.3    & 18.31   & \textbf{176.4}     & \textbf{111.41} \\
 boxing  & 0.1  & 12.1    & 99.6 & 829.17      & 99.96    & 832.17 & \textbf{100}   & \textbf{832.5}     & \underline{99.9}   & 831.67 \\
 breakout & 1.7 & 30.5    & 417.5 & 1443.75    & \textbf{787.34}   & \textbf{2727.92} & 747.9 & 2590.97  & \underline{696}    & 2410.76 \\
 centipede & 2090.9 & 12017 & 8167.3 & 61.22   & 11049.75 & 90.26   & \textbf{292792} & \textbf{2928.65} & 38938  & 371.21 \\
 chopper command & 811 & 7387.8 & 16654 & 240.89 & 28255  & 417.29  & \textbf{761699} & \textbf{11569.27} & 41495 & 618.60 \\
 crazy climber & 10780.5 & 36829.4 & \textbf{168788.5} & \textbf{630.80} & 136950 & 503.69 & 167820  & 626.93 & 157250 & 584.73 \\
 defender & 2874.5 & 18688.9 & 55105 & 330.27 & 185203 & 1152.93 & 336953  & 2112.50   & \underline{\textbf{837750}} & \textbf{5279.21} \\
 demon attack & 152.1 & 1971 & 111185 & 6104.40 & 132826.98 & 7294.24 & 133530 & 7332.89 & \underline{\textbf{549450}} & \textbf{30199.46} \\
 double dunk & -18.6 & -16.4 & -0.3   & 831.82  & -0.33     & 830.45  & 14     & 1481.82 & \underline{\textbf{23}}   & \textbf{1890.91} \\
 enduro      & 0   & 860.5 & 2125.9 & 247.05  & 0       & 0.00     & 0    & 0.00       & \underline{\textbf{14317}}  & \textbf{1663.80} \\
 fishing derby & -91.7 & -38.8 & 31.3 & 232.51  & 44.85   & 258.13    & 45.2   & 258.79  & \textbf{48.8} & \textbf{265.60} \\
 freeway       & 0     & 29.6  & \textbf{34} & \textbf{114.86}  & 0     & 0.00       & 0    & 0.00       & \underline{33.7}   & 113.85 \\
 frostbite     & 65.2  & 4334.7 & \textbf{9590.5} & \textbf{223.10} & 317.75 & 5.92     & 5083.5 & 117.54  & 8102 & 188.24 \\
 gopher  & 257.6 & 2412.5 & 70354.6 & 3252.91    & 66782.3 & 3087.14 & 114820.7 & 5316.40 & \underline{\textbf{454150}} & \textbf{21063.27} \\
 gravitar & 173 & 3351.4  & 1419.3  & 39.21   & 359.5      & 5.87    & 1106.2   & 29.36   & \textbf{6150}  & \textbf{188.05} \\
 hero   & 1027 & 30826.4 & \textbf{55887.4} & \textbf{184.10}   & 33730.55  & 109.75   & 31628.7 & 102.69   & 17655 & 55.80 \\
 ice hockey & -11.2 & 0.9 & 1.1    & 101.65   & 3.48      & 121.32   & \textbf{17.4}    & \textbf{236.36}   & -8.1  & 25.62 \\
 jamesbond  & 29    & 302.8 & 19809 & 72.24   & 601.5     & 209.09   & 37999.8 & 13868.08 & \underline{\textbf{567020}} & \textbf{207082.18} \\
 kangaroo   & 52    & 3035 & \textbf{14637.5} & \textbf{488.05} & 1632    & 52.97    & 14308   & 477.91     & 14286 & 477.17 \\
 krull     & 1598   & 2665.5 & 8741.5  & 669.18 & 8147.4  & 613.53   & 9387.5  &  729.70  & \textbf{11104} & \textbf{890.49} \\
 kung fu master & 258.5 & 22736.3 & 52181 & 230.99 & 43375.5 & 191.82 & 607443 & 2701.26  & \underline{\textbf{1270800}} & \textbf{5652.43} \\
 montezuma revenge & 0  & \textbf{4753.3}  & 384   & 8.08   & 0       & 0.00   & 0.3    & 0.01     & 2528  & 53.18 \\
 ms pacman  & 307.3 & 6951.6   & 5380.4  & 76.35   & \textbf{7342.32} & \textbf{105.88} & 6565.5 & 94.19    & 4296  & 60.03 \\
 name this game & 2292.3 & 8049 & 13136 & 188.37   & 21537.2 & 334.30 & 26219.5 & 415.64  & \textbf{30037} & \textbf{481.95} \\
 phoenix & 761.5 & 7242.6  & 108529 & 1662.80   & 210996.45  & 3243.82 & 519304 & 8000.84 & \underline{\textbf{597580}} & \textbf{9208.60} \\
 pitfall & -229.4 & \textbf{6463.7} & 0      & 3.43      & -1.66      & 3.40    & -0.6   & 3.42    & -21.8 & 3.10 \\
 pong    & -20.7  & 14.6   & 20.9   & 117.85    & 20.98      & 118.07  & \textbf{21}     &  \textbf{118.13} & \underline{\textbf{21}}    & \textbf{118.13} \\
 private eye & 24.9 & \textbf{69571.3} & 4234 & 6.05     & 98.5       & 0.11    & 96.3   & 0.10    & 15095  & 21.67 \\
 qbert  & 163.9 & 13455.0 & 33817.5  & 253.20   & \textbf{351200.12}  & \textbf{2641.14} & 21449.6 & 160.15 & 19091 & 142.40 \\
 riverraid & 1338.5 & 17118.0 & 22920.8 & 136.77 & 29608.05  & 179.15  & \textbf{40362.7} & \textbf{247.31} & 17081 & 99.77 \\
 road runner & 11.5 & 7845    & \textbf{62041}   & \textbf{791.85} & 57121     & 729.04  & 45289   & 578.00 & 57102 & 728.80 \\
 robotank   & 2.2   & 11.9  & 61.4   & 610.31    & 12.96     & 110.93  & 62.1    & 617.53 & \textbf{69.7}  & \textbf{695.88} \\
 seaquest  & 68.4 & \textbf{42054.7} & 15898.9 & 37.70    & 1753.2    & 4.01    & 2890.3  & 6.72   & 2728  & 6.33 \\
 skiing & -17098  & \textbf{-4336.9} & -12957.8 & 32.44  & -10180.38 & 54.21   & -29968.4 & -100.86 & -9327 & 60.90 \\
 solaris & 1236.3 & \textbf{12326.7} & 3560.3  & 20.96  & 2365      & 10.18   & 2273.5   & 9.35    & 3653  & 21.79 \\
 space invaders & 148 & 1668.7 & 18789 & 1225.82 & 43595.78 & 2857.09 & 51037.4 & 3346.45 & \underline{\textbf{105810}} & \textbf{6948.25} \\
 star gunner & 664 & 10250 & 127029    & 1318.22 & 200625   & 2085.97 & 321528  & 3347.21 & \textbf{358650} & \textbf{3734.47} \\
 surround    & -10 & 6.5   & \textbf{9.7}       & \textbf{119.39}  & 7.56     & 106.42  & 8.4     & 111.52  & -9.8  & 1.21 \\
 tennis  & -23.8   & -8.3 & 0        & 153.55    & 0.55     & 157.10  & 12.2    & 232.26  & \underline{\textbf{23.7}}  & \textbf{306.45} \\
 time pilot & 3568 & 5229.2 & 12926 & 563.36     & 48481.5  & 2703.84 & 105316  & 6125.34 & \textbf{150930} & \textbf{8871.35} \\
 tutankham  & 11.4 & 167.6  & 241   & 146.99     & 292.11   & 179.71  & 278.9   & 171.25  & \textbf{380.3} & \textbf{236.17} \\
 up n down  & 533.4 & 11693.2 & 125755 & 1122.08 & 332546.75 & 2975.08 & 345727 & 3093.19 & \underline{\textbf{907170}} & \textbf{8124.13} \\
 venture    & 0     & 1187.5  & 5.5    & 0.46    & 0         & 0.00    & 0      & 0.00    & \textbf{1969}  & \textbf{165.81} \\
 video pinball & 0 & 17667.9  & 533936.5 & 3022.07 & 572898.27 & 3242.59 & 511835 & 2896.98 & \underline{\textbf{673840}} & \textbf{3813.92} \\
 wizard of wor & 563.5 & 4756.5 & 17862.5 & 412.57 & 9157.5    & 204.96  & \textbf{29059.3} & \textbf{679.60} & 21325  & 495.15 \\
 yars revenge & 3092.9 & 54576.9 & 102557 & 193.19 & 84231.14  & 157.60 & \textbf{166292.3} & \textbf{316.99} & 84684  & 158.48 \\
 zaxxon       & 32.5   & 9173.3 & 22209.5 & 242.62 & 32935.5   & 359.96 & 41118    & 449.47 & \textbf{62133}  & \textbf{679.38} \\
\hline
MEAN HNS(\%) &     0.00 & 100.00   &         & 873.97 &         & 957.34  &        & 1741.36 &      & \textbf{6456.63} \\
\hline
MEDIAN HNS(\%) & 0.00   & 100.00   &         & 230.99 &         & 191.82  &        & 454.91  &      & \textbf{477.17} \\
\bottomrule
\end{tabular}
\end{center}
\normalsize
\clearpage

\small
\begin{center}
\begin{tabular}{|c|c c|c c|c c|c c|}
\toprule
 Games & R2D2 & HNS(\%) & NGU & HNS(\%) & AGENT57 & HNS(\%) & CASA & HNS(\%) \\
\midrule
Scale  & 10B   &        & 35B &         & 100B     &        & 200M & \\
\midrule
 alien  & 109038.4 & 1576.97 & 248100 & 3592.35 & \textbf{297638.17} & \textbf{4310.30} & 10641 & 150.92 \\
 amidar & 27751.24 & 1619.04 & 17800  & 1038.35 & \textbf{29660.08}  & \textbf{1730.42} & 653.9  & 37.82 \\
 assault & \textbf{90526.44} & \textbf{17379.53} & 34800 & 6654.66 & 67212.67 & 12892.66 & 36251 & 6933.91  \\
 asterix & \textbf{999080}   & \textbf{12044.30} & 950700 & 11460.94 & 991384.42 & 11951.51 & \underline{851210} & 10261.30 \\
 asteroids & 265861.2 & 568.12 & 230500 & 492.36   & 150854.61 & 321.70   & \underline{\textbf{759170}} & \textbf{1625.15}\\
 atlantis & 1576068   & 9662.56 & 1653600 & 10141.80 & 1528841.76 & 9370.64 & \underline{\textbf{3670700}} & \textbf{22609.89} \\
 bank heist & \textbf{46285.6} & \textbf{6262.20} & 17400   & 2352.93  & 23071.5    & 3120.49 & 1381   & 184.98 \\
 battle zone & 513360 & 1388.64 & 691700  & 1871.27  & \textbf{934134.88}  & \textbf{2527.36} & 130410 & 352.28 \\
 beam rider & 128236.08 & 772.05 & 63600  & 381.80   & \textbf{300509.8}   & \textbf{1812.19} & 104030  & 625.90 \\
 berzerk & 34134.8      & 1356.81 & 36200 & 1439.19  & \textbf{61507.83}   & \textbf{2448.80} & 1222   & 43.81 \\
 bowling & 196.36       & 125.92  & 211.9 & 137.21   & \textbf{251.18}     & \textbf{165.76}  & 176.4     & 111.41 \\
 boxing  & 99.16        & 825.50  & 99.7  & 830.00   & \textbf{100}        & \textbf{832.50}  & \underline{\textbf{99.9}}   & 831.67 \\
 breakout & \textbf{795.36}      & \textbf{2755.76} & 559.2 & 1935.76  & 790.4      & 2738.54 & \underline{696}    & 2410.76 \\
 centipede & 532921.84  & 5347.83 & \textbf{577800} & \textbf{5799.95} & 412847.86  & 4138.15 & 38938  & 371.21 \\
 chopper command & 960648 & 14594.29 & \textbf{999900} & \textbf{15191.11} & \textbf{999900} & \textbf{15191.11} & 41495 & 618.60 \\
 crazy climber & 312768   & 1205.59  & 313400 & 1208.11  & \textbf{565909.85} & \textbf{2216.18} & 157250 & 584.73 \\
 defender & 562106        & 3536.22  & 664100 & 4181.16  & 677642.78 & 4266.80 & \underline{\textbf{837750}} & \textbf{5279.21} \\
 demon attack & 143664.6  & 7890.07  & 143500 & 7881.02  & 143161.44 & 7862.41 & \underline{\textbf{549450}} & \textbf{30199.46} \\
 double dunk & 23.12      & 1896.36  & -14.1  & 204.55   & \textbf{23.93}     & \textbf{1933.18} & \underline{23}   & 1890.91 \\
 enduro      & 2376.68    & 276.20   & 2000   & 232.42   & 2367.71   & 275.16  & \underline{\textbf{14317}}  & \textbf{1663.80} \\
 fishing derby & 81.96    & 328.28   & 32     & 233.84   & \textbf{86.97}     & \textbf{337.75}  & 48.8 & 265.60 \\
 freeway       & \textbf{34}       & \textbf{114.86}   & 28.5   & 96.28    & 32.59     & 110.10  & \underline{33.7}   & 113.85 \\
 frostbite    & 11238.4  & 261.70   & 206400 & 4832.76  & \textbf{541280.88} & \textbf{12676.32} & 8102 & 188.24 \\
 gopher  & 122196        & 5658.66  & 113400 & 5250.47  & 117777.08 & 5453.59  & \underline{\textbf{454150}} & \textbf{21063.27} \\
 gravitar & 6750         & 206.93   & 14200  & 441/32   & \textbf{19213.96}  & \textbf{599.07}   & 6150  & 188.05 \\
 hero   & 37030.4        & 120.82   & 69400  & 229.44   & \textbf{114736.26} & \textbf{381.58}    & 17655 & 55.80 \\
 ice hockey & \textbf{71.56}      & \textbf{683.97}   & -4.1   & 58.68    & 63.64     & 618.51   & -8.1  & 25.62 \\
 jamesbond  & 23266      & 8486.85  & 26600  & 9704.53  & 135784.96 & 49582.16 & \underline{\textbf{567020}} & \textbf{207082.18} \\
 kangaroo   & 14112      & 471.34   & \textbf{35100}  & \textbf{1174.92}  & 24034.16  & 803.96   & 14286 & 477.17 \\
 krull     & 145284.8    & 13460.12 & 127400 & 11784.73 & \textbf{251997.31} & \textbf{23456.61} & 11104 & 890.49 \\
 kung fu master & 200176 & 889.40   & 212100 & 942.45   & 206845.82 & 919.07   & \underline{\textbf{1270800}} & \textbf{5652.43} \\
 montezuma revenge & 2504 & 52.68   & \textbf{10400}  & \textbf{218.80}   & 9352.01   & 196.75   & 2528  & 53.18 \\
 ms pacman  & 29928.2     & 445.81  & 40800  & 609.44   & \textbf{63994.44}  & \textbf{958.52}   & 4296  & 60.03 \\
 name this game & 45214.8 & 745.61  & 23900  & 375.35   & \textbf{54386.77}  & \textbf{904.94}   & 30037 & 481.95 \\
 phoenix & 811621.6       & 125.11  & \textbf{959100} & \textbf{14786.66} & 908264.15 & 14002.29 & \underline{597580} & 9208.60 \\
 pitfall & 0              & 3.43    & 7800   & 119.97   & \textbf{18756.01}  & \textbf{283.66}   & -21.8 & 3.10 \\
 pong    & \textbf{21}             & \textbf{118.13}  & 19.6   & 114.16   & 20.67     & 117.20   & \underline{\textbf{21}}    & \textbf{118.13} \\
 private eye & 300        & 0.40    & \textbf{100000} & \textbf{143.75}   & 79716.46  & 114.59   & 15095  & 21.67 \\
 qbert  & 161000          & 1210.10 & 451900 & 3398.79  & \textbf{580328.14} & \textbf{4365.06}  & 19091 & 142.40 \\
 riverraid & 34076.4      & 207.47  & 36700  & 224.10   & \textbf{63318.67}  & \textbf{392.79}   & 17081 & 99.77 \\
 road runner & \textbf{498660}     & \textbf{6365.59} & 128600 & 1641.52  & 243025.8  & 3102.24  & 57102 & 728.80 \\
 robotank   & \textbf{132.4}       & \textbf{1342.27} & 9.1    & 71.13    & 127.32    & 1289.90  & 69.7  & 695.88 \\
 seaquest  & 999991.84    & 2381.55 & \textbf{1000000} & \textbf{2381.57} & 999997.63 & 2381.56  & 2728  & 6.33 \\
 skiing & -29970.32       & -100.87 & -22977.9 & -46.08 & \textbf{-4202.6}   & \textbf{101.05}   & -9327 & 60.90 \\
 solaris & 4198.4         & 26.71   & 4700     & 31.23  & \textbf{44199.93}  & \textbf{387.39}   & 3653  & 21.79 \\
 space invaders & 55889   & 3665.48 & 43400    & 2844.22 & 48680.86 & 3191.48  & \underline{\textbf{105810}} & \textbf{6948.25}\\
 star gunner & 521728     & 5435.68 & 414600   & 4318.13 & \textbf{839573.53} & \textbf{8751.40} & 358650 & 3734.47\\
 surround    & \textbf{9.96}       & \textbf{120.97}  & -9.6     & 2.42    & 9.5       & 118.18  & -9.8  & 1.21 \\
 tennis  & \textbf{24}             & \textbf{308.39}  & 10.2     & 219.35  & 23.84     & 307.35  & \underline{23.7}  & 306.45 \\
 time pilot & 348932    & 20791.28 & 344700  & 20536.51 & \textbf{405425.31} & \textbf{24192.24} & 150930 & 8871.35 \\
 tutankham  & 393.64    & 244.71   & 191.1   & 115.04   & \textbf{2354.91}   & \textbf{1500.33}  & 380.3 & 236.17 \\
 up n down  & 542918.8  & 4860.17  & 620100  & 5551.77  & 623805.73 & 5584.98  & \underline{\textbf{907170}} & \textbf{8124.13} \\
 venture    & 1992      & 167.75   & 1700    & 143.16   & \textbf{2623.71}   & \textbf{220.94}   & 1969  & 165.81 \\
 video pinball & 483569.72 & 2737.00 & 965300 & 5463.58 & \textbf{992340.74} & \textbf{5616.63}  & \underline{673840} & 3813.92 \\
 wizard of wor & 133264 & 3164.81  & 106200  & 2519.35  & \textbf{157306.41} & \textbf{3738.20}  & 21325  & 495.15 \\
 yars revenge & 918854.32 & 1778.73 & 986000 & 1909.15  & \textbf{998532.37} & \textbf{1933.49}  & 84684  & 158.48 \\
 zaxxon & 181372        & 1983.85  & 111100  & 1215.07  & \textbf{249808.9}  & \textbf{2732.54}  & 62133  & 679.38 \\
\hline
MEAN HNS(\%) &               & 3374.31  &         &  3169.90 &           & 4763.69  &     & \textbf{6456.63} \\
\hline
MEDIAN HNS(\%) &            & 1342.27   &         & 1208.11  &           & \textbf{1933.49}  &     & 477.17 \\
\bottomrule
\end{tabular}
\end{center}

\normalsize
\clearpage
\tiny
\renewcommand{\thesubfigure}{\arabic{subfigure}.}
\begin{figure} 
    \subfigure[alien]{
    \includesvg[width=0.25\textwidth,inkscapelatex=false]{bfigure/alien.svg}
    }
    \subfigure[amidar]{
    \includesvg[width=0.25\textwidth,inkscapelatex=false]{bfigure/amidar.svg}
    }
    \subfigure[assault]{
    \includesvg[width=0.25\textwidth,inkscapelatex=false]{bfigure/assault.svg}
    }
    \subfigure[asterix]{
    \includesvg[width=0.25\textwidth,inkscapelatex=false]{bfigure/asterix.svg}
    }
\end{figure}

\vspace{-0.0cm}
\begin{figure}
    \subfigure[asteroids]{
    \includesvg[width=0.25\textwidth,inkscapelatex=false]{bfigure/asteroids.svg}
    }
    \subfigure[atlantis]{
    \includesvg[width=0.25\textwidth,inkscapelatex=false]{bfigure/atlantis.svg}
    }
    \subfigure[bank\_heist]{
    \includesvg[width=0.25\textwidth,inkscapelatex=false]{bfigure/bank_heist.svg}
    }
    \subfigure[battle\_zone]{
    \includesvg[width=0.25\textwidth,inkscapelatex=false]{bfigure/battle_zone.svg}
    }
\end{figure}

\vspace{-0.0cm}
\begin{figure}
    \subfigure[beam\_rider]{
    \includesvg[width=0.25\textwidth,inkscapelatex=false]{bfigure/beam_rider.svg}
    }
    \subfigure[berzerk]{
    \includesvg[width=0.25\textwidth,inkscapelatex=false]{bfigure/berzerk.svg}
    }
    \subfigure[bowling]{
    \includesvg[width=0.25\textwidth,inkscapelatex=false]{bfigure/bowling.svg}
    }
    \subfigure[boxing]{
    \includesvg[width=0.25\textwidth,inkscapelatex=false]{bfigure/boxing.svg}
    }
\end{figure}

\vspace{-0.0cm}
\begin{figure}
    \subfigure[breakout]{
    \includesvg[width=0.25\textwidth,inkscapelatex=false]{bfigure/breakout.svg}
    }
    \subfigure[centipede]{
    \includesvg[width=0.25\textwidth,inkscapelatex=false]{bfigure/centipede.svg}
    }
    \subfigure[chopper\_command]{
    \includesvg[width=0.25\textwidth,inkscapelatex=false]{bfigure/chopper_command.svg}
    }
    \subfigure[crazy\_climber]{
    \includesvg[width=0.25\textwidth,inkscapelatex=false]{bfigure/crazy_climber.svg}
    }
\end{figure}

\vspace{-0.0cm}
\begin{figure}
    \subfigure[defender]{
    \includesvg[width=0.25\textwidth,inkscapelatex=false]{bfigure/defender.svg}
    }
    \subfigure[demon\_attack]{
    \includesvg[width=0.25\textwidth,inkscapelatex=false]{bfigure/demon_attack.svg}
    }
    \subfigure[double\_dunk]{
    \includesvg[width=0.25\textwidth,inkscapelatex=false]{bfigure/double_dunk.svg}
    }
    \subfigure[enduro]{
    \includesvg[width=0.25\textwidth,inkscapelatex=false]{bfigure/enduro.svg}
    }
\end{figure}
\clearpage
\vspace{-0.0cm}
\begin{figure}
    \subfigure[fishing\_derby]{
    \includesvg[width=0.25\textwidth,inkscapelatex=false]{bfigure/fishing_derby.svg}
    }
    \subfigure[freeway]{
    \includesvg[width=0.25\textwidth,inkscapelatex=false]{bfigure/freeway.svg}
    }
    \subfigure[frostbite]{
    \includesvg[width=0.25\textwidth,inkscapelatex=false]{bfigure/frostbite.svg}
    }
    \subfigure[gopher]{
    \includesvg[width=0.25\textwidth,inkscapelatex=false]{bfigure/gopher.svg}
    }
\end{figure}

\vspace{-0.0cm}
\begin{figure}
    \subfigure[gravitar]{
    \includesvg[width=0.25\textwidth,inkscapelatex=false]{bfigure/gravitar.svg}
    }
    \subfigure[hero]{
    \includesvg[width=0.25\textwidth,inkscapelatex=false]{bfigure/hero.svg}
    }
    \subfigure[ice\_hockey]{
    \includesvg[width=0.25\textwidth,inkscapelatex=false]{bfigure/ice_hockey.svg}
    }
    \subfigure[jamesbond]{
    \includesvg[width=0.25\textwidth,inkscapelatex=false]{bfigure/jamesbond.svg}
    }
\end{figure}

\vspace{-0.0cm}
\begin{figure}
    \subfigure[kangaroo]{
    \includesvg[width=0.25\textwidth,inkscapelatex=false]{bfigure/kangaroo.svg}
    }
    \subfigure[krull]{
    \includesvg[width=0.25\textwidth,inkscapelatex=false]{bfigure/krull.svg}
    }
    \subfigure[kung\_fu\_master]{
    \includesvg[width=0.25\textwidth,inkscapelatex=false]{bfigure/kung_fu_master.svg}
    }
    \subfigure[montezuma\_revenge]{
    \includesvg[width=0.25\textwidth,inkscapelatex=false]{bfigure/montezuma_revenge.svg}
    }
\end{figure}

\vspace{-0.0cm}
\begin{figure}
    \subfigure[ms\_pacman]{
    \includesvg[width=0.25\textwidth,inkscapelatex=false]{bfigure/ms_pacman.svg}
    }
    \subfigure[name\_this\_game]{
    \includesvg[width=0.25\textwidth,inkscapelatex=false]{bfigure/name_this_game.svg}
    }
    \subfigure[phoenix]{
    \includesvg[width=0.25\textwidth,inkscapelatex=false]{bfigure/phoenix.svg}
    }
    \subfigure[pitfall]{
    \includesvg[width=0.25\textwidth,inkscapelatex=false]{bfigure/pitfall.svg}
    }
\end{figure}

\vspace{-0.0cm}
\begin{figure}
    \subfigure[pong]{
    \includesvg[width=0.25\textwidth,inkscapelatex=false]{bfigure/pong.svg}
    }
    \subfigure[private\_eye]{
    \includesvg[width=0.25\textwidth,inkscapelatex=false]{bfigure/private_eye.svg}
    }
    \subfigure[qbert]{
    \includesvg[width=0.25\textwidth,inkscapelatex=false]{bfigure/qbert.svg}
    }
    \subfigure[riverraid]{
    \includesvg[width=0.25\textwidth,inkscapelatex=false]{bfigure/riverraid.svg}
    }
\end{figure}
\clearpage
\vspace{-0.0cm}
\begin{figure}
    \subfigure[road\_runner]{
    \includesvg[width=0.25\textwidth,inkscapelatex=false]{bfigure/road_runner.svg}
    }
    \subfigure[robotank]{
    \includesvg[width=0.25\textwidth,inkscapelatex=false]{bfigure/robotank.svg}
    }
    \subfigure[seaquest]{
    \includesvg[width=0.25\textwidth,inkscapelatex=false]{bfigure/seaquest.svg}
    }
    \subfigure[skiing]{
    \includesvg[width=0.25\textwidth,inkscapelatex=false]{bfigure/skiing.svg}
    }
\end{figure}

\vspace{-0.0cm}
\begin{figure}
    \subfigure[solaris]{
    \includesvg[width=0.25\textwidth,inkscapelatex=false]{bfigure/solaris.svg}
    }
    \subfigure[space\_invader]{
    \includesvg[width=0.25\textwidth,inkscapelatex=false]{bfigure/space_invader.svg}
    }
    \subfigure[star\_gunner]{
    \includesvg[width=0.25\textwidth,inkscapelatex=false]{bfigure/star_gunner.svg}
    }
    \subfigure[surround]{
    \includesvg[width=0.25\textwidth,inkscapelatex=false]{bfigure/surround.svg}
    }
\end{figure}

\vspace{-0.0cm}
\begin{figure}
    \subfigure[tennis]{
    \includesvg[width=0.25\textwidth,inkscapelatex=false]{bfigure/tennis.svg}
    }
    \subfigure[time\_pilot]{
    \includesvg[width=0.25\textwidth,inkscapelatex=false]{bfigure/time_pilot.svg}
    }
    \subfigure[tutankham]{
    \includesvg[width=0.25\textwidth,inkscapelatex=false]{bfigure/tutankham.svg}
    }
    \subfigure[up\_n\_down]{
    \includesvg[width=0.25\textwidth,inkscapelatex=false]{bfigure/up_n_down.svg}
    }
\end{figure}

\vspace{-0.0cm}
\begin{figure}
    \subfigure[venture]{
    \includesvg[width=0.25\textwidth,inkscapelatex=false]{bfigure/venture.svg}
    }
    \subfigure[video\_pinball]{
    \includesvg[width=0.25\textwidth,inkscapelatex=false]{bfigure/video_pinball.svg}
    }
    \subfigure[wizard\_of\_wor]{
    \includesvg[width=0.25\textwidth,inkscapelatex=false]{bfigure/wizard_of_wor.svg}
    }
    \subfigure[yars\_revenge]{
    \includesvg[width=0.25\textwidth,inkscapelatex=false]{bfigure/yars_revenge.svg}
    }
\end{figure}

\vspace{-0.0cm}
\begin{figure}
    \subfigure[zaxxon]{
    \includesvg[width=0.25\textwidth,inkscapelatex=false]{bfigure/zaxxon.svg}
    }
\end{figure}
\clearpage